\documentclass{article}
\usepackage{iclr2027_conference,times}

\usepackage{amsmath,amsfonts,bm}

\newcommand{\figleft}{{\em (Left)}}

\newcommand{\figright}{{\em (Right)}}

\def\eqref#1{equation~\ref{#1}}
\def\1{\bm{1}}

\DeclareMathAlphabet{\mathsfit}{\encodingdefault}{\sfdefault}{m}{sl}
\SetMathAlphabet{\mathsfit}{bold}{\encodingdefault}{\sfdefault}{bx}{n}

\usepackage{hyperref}
\usepackage{url}
\usepackage{amssymb}
\usepackage{graphicx}
\usepackage{microtype}
\usepackage{booktabs}
\usepackage{longtable}
\usepackage{float}

\graphicspath{{./}{figures/}}

\providecommand{\figleft}{{\em (Left)}}
\providecommand{\figright}{{\em (Right)}}

\title{Training on the Future: A Delay-Aware Audit of Test-Time Adaptation for Time-Series Forecasting}

\author{Mohamed Readh Fentazi, Mazene Ameur \& Adlen Ksentini \\
Communication Systems Department\\
EURECOM\\
Sophia Antipolis, France \\
\texttt{\{mohamed-readh.fentazi,mazene.ameur,adlen.ksentini\}@eurecom.fr} \\
}

\iclrfinalcopy

\begin{document}

\maketitle
\lhead{Preprint.}

\begin{abstract}
Test-time adaptation (TTA) methods for time-series forecasting update a deployed model, or a small adapter around it, from incoming ground truth. But the label of an $H$-step forecast exists only $H$ steps later, and real data pipelines add further delay. We build a leakage-free harness in which the label of forecast origin $s$ is released for updates only at step $s+d$ with $d \ge H$, and enforce this rule inside the released code of four recent TTA methods (TAFAS, COSA, PETSA and DynaTTA), run on their own backbones and checkpoints across five benchmarks (ETTm1, ETTh2, Weather, Electricity and Traffic). As references we add two closed-form correctors: a bank of recursive least squares (RLS) filters combined by a per-coordinate median, with no tunable hyperparameters and 56 microseconds per step on the 7-channel streams, and an ELF-style linear corrector. Under causal delayed labels the picture is asymmetric. On ETTm1 every audited method genuinely adapts, yet the RLS bank still beats three of the four at a fraction of their cost; only DynaTTA beats the bank, only at the minimum causal delay, and at roughly 2{,}500 times the per-update cost; the ELF-style corrector beats all four. On the other four datasets, the largest statistically significant improvement any published method achieves over its own frozen checkpoint is half a percent, on all four at least one published method is significantly worse than the frozen model at the minimum causal delay, and on drift-heavy ETTh2 longer label delays make every adapter that separates from the frozen model, ours included, significantly harmful. Leaky next-step updates inflate the apparent gains of simple adapters by up to 110\%, and the backbone training recipe moves frozen online error by up to a factor of 25, more than any adaptation effect we measure. We release the harness, integration patches and all cached runs.
\end{abstract}

\section{Introduction}
\label{sec:intro}

Deployed forecasting models degrade as the data they see drifts away from the data they were trained on, and a growing literature proposes test-time adaptation (TTA) as the remedy: small gradient updates to the deployed model, or to a lightweight adapter around it, driven by the ground truth that arrives during deployment \citep{kim2025tafas,medeiros2025petsa,im2026cosa,grover2025dynatta}. The premise is attractive. The evaluation is not yet trustworthy, for a simple reason: the label of an $H$-step-ahead forecast does not exist until $H$ steps after the forecast is issued. A protocol that lets an adapter consume the full target window of a forecast before that window has been observed is, in effect, training on the future. Online-forecasting setups descended from FSNet \citep{pham2023fsnet} do this at every step, and recent work has begun to correct it \citep{liang2024actnow,lau2025dsof,kim2025tafas}. Even the corrected protocols assume that a label can be used the moment it matures; real pipelines add lag on top, since metering, verification and ETL all delay the point at which an observation becomes usable as a label.

\begin{figure}[t]
\begin{center}
\includegraphics[width=0.85\linewidth]{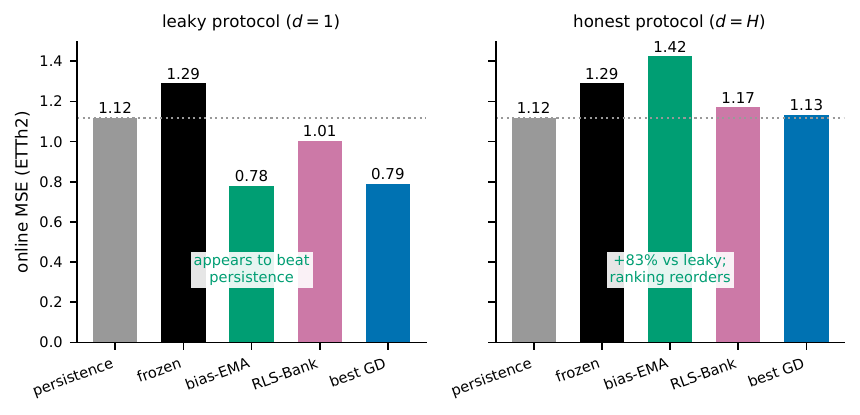}%
\end{center}
\caption{Leaky versus causal evaluation on ETTh2 (DLinear stream, standardized MSE). \figleft{} With next-step updates ($d{=}1$), all three adapters (bias-EMA, the RLS-Bank, and the best gradient adapter) score below the persistence baseline. \figright{} With updates restricted to matured windows ($d{=}H$), none does, and the ranking changes. Best GD: the per-panel oracle over the learning-rate grid. Values from Table~\ref{tab:full_etth2_dl}.}
\label{fig:leaderboards}
\end{figure}

This paper makes the delay explicit and turns it into an experimental axis. In our harness the label of forecast origin $s$ is released for adapter updates only at step $s+d$. The setting $d = H$ is the minimum causal delay, $d > H$ models pipeline latency, and $d = 1$ reproduces the leaky legacy protocol, which we report only as a reference. Figure~\ref{fig:leaderboards} previews why the distinction matters: on ETTh2, three simple adapters appear to beat a persistence baseline under leaky updates and lose to it under causal updates. We do not evaluate published methods on re-implementations: TAFAS, COSA, PETSA and DynaTTA run from their own repositories on their own checkpoints, on five standard benchmarks (ETTm1, ETTh2, Weather, Electricity and Traffic), with the delay enforced by small, fully logged patches inside their update loops and their partial-ground-truth mechanisms retained with causally shortened prefixes.

Against this stack we field two closed-form reference adapters. The first is ours. For every channel and horizon step, a recursive least squares (RLS) filter corrects the frozen model's forecast from four features. RLS is a textbook update rule from adaptive filtering with no learning rate. Its one knob, the forgetting factor, turns out to be regime-dependent, so we run a small bank of RLS filters across forgetting factors together with a passthrough expert and combine their outputs by a per-coordinate median. The median is not a stylistic choice: feedback-driven selection and mixing, the direct analogues of learned combination in the adaptive-filters literature, fail specifically under delay, because the feedback that would dismiss a failing expert arrives up to $d$ steps late (Section~\ref{sec:bank}). The resulting RLS-Bank has no tunable hyperparameters at deployment and costs 56 microseconds per step on one CPU core for a 7-channel, 24-horizon problem. The second reference is not ours: an ELF-style corrector \citep{lee2025adapts}, the closest published relative of our adapter, re-implemented inside the harness so that it, too, updates only on released labels. We report it with the same rigor as everything else, including the cells where it beats our bank.

Our contributions are the following. (i) A leakage-free, delay-parameterized evaluation protocol, including causal semantics for partial-ground-truth mechanisms under label lag, enforced inside all four audited methods' own code. (ii) The RLS-Bank, a closed-form adapter with no tunable hyperparameters at deployment, with an analysis of why feedback-driven expert selection fails under delayed labels and evidence that the median absorbs numerically exploding experts on the 321- and 862-channel streams. (iii) A five-dataset audit of TAFAS, COSA, PETSA and DynaTTA on their own checkpoints, with three checkpoint seeds on the ETT streams and block-bootstrap significance. On ETTm1, the one dataset where the published stack genuinely adapts, the bank beats three of the four at every delay at a small fraction of their cost, and the ELF-style corrector beats all four. Outside ETTm1, the largest significant improvement any published method achieves over its own frozen checkpoint is half a percent, and
on all four of the other datasets at least one published method is significantly worse than the frozen model at the minimum causal delay. (iv) Three findings about evaluation practice: leaky updates inflate the apparent gains of simple adapters by up to 110 percent; the backbone training recipe alone moves frozen online error by up to a factor of 25, more than any adaptation effect we measure; and gradient adapters are fragile in ways closed-form updates are not, with up to eleven of twelve gradient configurations diverging per cell.

We state the scope of the claim. This paper does not argue that gradient-based adaptation cannot help; on ETTm1 all four audited methods help, and DynaTTA beats the bank there. Nor is RLS new. The claim is that under causal, delay-aware evaluation, properly configured closed-form correctors set a bar that the audited stack clears exactly once across five datasets, and that evaluation practice, not only the methods, needs recalibration.

\section{Related Work}
\label{sec:related}

\paragraph{Test-time adaptation for forecasting.}
TAFAS \citep{kim2025tafas} introduced TSF-specific TTA with gated calibration modules trained on partially observed and matured ground truth; PETSA \citep{medeiros2025petsa} makes the adapters parameter-efficient; COSA \citep{im2026cosa} reduces the design to one gated output adapter and evaluates under a leakage-free streaming protocol; DynaTTA \citep{grover2025dynatta} adds shift-aware adaptation across five backbones, compared against the base model and TAFAS only; FAC \citep{fac2026} calibrates in the frequency domain, still by gradient steps (we adopt its term, matured ground truth). None varies the delay, includes a closed-form competitor, or reports naive baselines under a causal protocol.

\paragraph{Frozen models with lightweight correctors.}
ELF \citep{lee2025adapts}, called AdapTS in earlier preprints, is the closest relative of our adapter: a linear forecaster, OLS-equivalent in its uncompressed form, is fit online next to a frozen foundation model and combined with it by an online weighting rule; we re-implement it inside the harness as a baseline on every stream with $C \le 21$ (Section~\ref{sec:experiments}). Concurrent ORCA \citep{orca2026} trains a linear residual adapter from a replay buffer that implements the same maturity rule we use, evidence that the causal protocol is becoming standard, and is a natural future competitor. Neither varies the delay or audits the TTA stack.

\paragraph{Classical adaptive filtering.}
RLS with exponential forgetting is textbook material \citep{haykin2002adaptive}, and delayed RLS updates have a long applied history: Madsen's $k$-step recursive scheme updates coefficients from regressors $k$ steps stale \citep{madsen2007timeseries}, our $d$-lagged setting under another name. The combination-of-adaptive-filters literature \citep{arenas2006meansquare,arenas2016combinations} runs filters with different forgetting factors in parallel, including fast and slow RLS pairs, but combines them with learned mixing weights, the feedback-driven mechanism our ablations show breaks under delayed labels. Related lines: variable-forgetting-factor RLS \citep{fortescue1981vff}, differentiable forgetting \citep{bennett2022differentiable}, universal linear prediction \citep{singer1999universal}, and prediction with delayed feedback \citep{weinberger2002delayed}, which underpins our selection-under-lag findings.

\paragraph{Causal protocols and delayed feedback in online forecasting.}
DSOF \citep{lau2025dsof}, Act-Now \citep{liang2024actnow} and Proceed \citep{zhao2025proceed} remove leakage from online-forecasting evaluation; OneNet's appendix \citep{zhang2023onenet} showed methods collapsing under delayed feedback, and ADAPT-Z \citep{adaptz2025}, ADOWIP \citep{adowip2025} and CEP \citep{zhan2025continuous} study delayed feedback directly, CEP reporting online methods below naive baselines. We extend this line from training-time online learning to test-time adaptation, make $d$ a variable, and enforce it in published code. Like the It's-TIME benchmark \citep{itstime2025}, we report naive baselines in every table; Appendix~\ref{app:integration} notes the nowcasting (ragged-edge) connection.

\section{The Causal Delayed Evaluation Protocol}
\label{sec:protocol}

\paragraph{Setup.}
We use five standard multivariate benchmarks: ETTh2 and ETTm1 \citep{zhou2021informer} with 7 channels, and the Weather (21 channels), Electricity (321 channels) and Traffic (862 channels) datasets from the standard long-term-forecasting benchmark suite \citep{wu2021autoformer}. Each series $y_{1:N} \in \mathbb{R}^{N\times C}$ is split by ratio into pretraining (25\%), validation (5\%) and an online stream (70\%); standardization uses pretraining statistics only. The split is deliberately nonstandard: a 70\% online stream gives delay effects and drift enough support for block-bootstrap power, at the price of a smaller pretraining region, whose cost Section~\ref{sec:grid} shows for the frozen backbones. Every internal comparison replays the same frozen stream, so the split handicaps all update rules equally, and the external audit (Section~\ref{sec:headline}) is insulated because it runs on the authors' own recipes and checkpoints. A frozen backbone maps the lookback window $y_{t-L:t}$ ($L{=}96$) to a forecast $\hat y_{t:t+H}$ ($H{=}24$) at every stream step $t$, which we call the forecast origin (one origin per step, stride 1). Adapters may modify the forecast; the backbone itself is never updated. We refer to a dataset-and-backbone pair as a cell.

\begin{figure}[t]
\begin{center}
\includegraphics[width=0.7\linewidth]{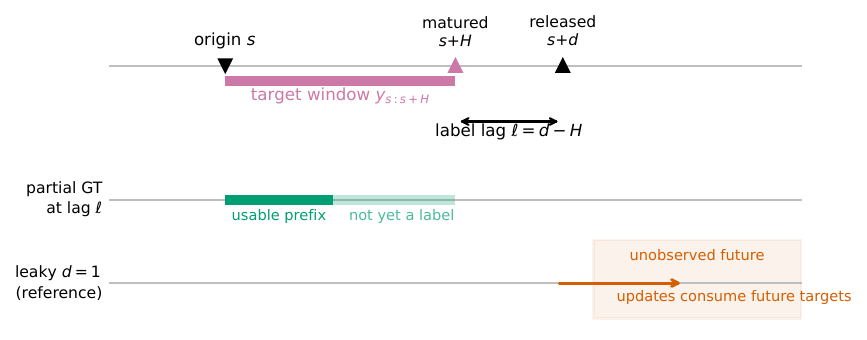}%
\end{center}
\caption{Delay-parameterized label release. The target window of origin $s$ matures at $s+H$ and is released to the adapter at $s+d$. $d{=}H$ is the minimum causal delay; $d{>}H$ adds a label lag $\ell{=}d-H$ that also shortens the usable partial-ground-truth prefix; $d{=}1$ is the leaky reference, in which updates consume targets that have not yet been observed.}
\label{fig:protocol}
\end{figure}

\paragraph{Delay semantics.}
The label window of origin $s$ is $y_{s:s+H}$. It may be consumed for adapter updates only from step $s+d$. We call $d = H$ the minimum causal delay: a window is used as soon as it has been fully observed. For $d > H$ the labels are delayed by $\ell = d - H$ steps; observations reach the adapter as labels $\ell$ steps after they occur, while the forecaster always sees fresh inputs. The setting $d = 1$ lets an adapter use the full window of origin $s$ at step $s+1$, before $H-1$ of its values exist; this is the leaky legacy protocol, and we report it only as a reference. Figure~\ref{fig:protocol} summarizes the three cases. Some methods also update on partial ground truth, the already observed prefix $y_{s:t}$ of an unmatured window; at label delay $\ell$ that prefix shrinks by $\ell$ steps, and all four audited methods keep their partial-ground-truth mechanisms with the prefix reduced accordingly (bit-identical to the released code at $\ell = 0$).

\paragraph{Measurement.}
We report standardized MSE over the full online stream ($T{=}12{,}170$ steps for ETTh2, $48{,}752$ for ETTm1, $36{,}864$ for Weather, $18{,}389$ for Electricity and $12{,}257$ for Traffic). Uncertainty is assessed on paired per-step loss differentials with a circular moving-block bootstrap (block length twice the seasonal period; the seasonal periods are 24, 96, 144, 24 and 24 steps respectively; $B{=}500$; Appendix~\ref{app:bootstrap}). A difference is significant when its 95\% interval excludes zero (asterisks in the appendix tables; n.s.\ otherwise). The internal pipeline is deterministic given the backbone (reproduced across three machines to within $10^{-4}$, the level of BLAS floating-point variation). Internal PatchTST cells use three seeds on ETTh2, ETTm1 and Weather and one on Electricity and Traffic, a compute-budget choice we state up front; external ETT cells use three checkpoint seeds (Section~\ref{sec:headline}).

\paragraph{Audited methods.}
TAFAS \citep{kim2025tafas} attaches gated calibration modules to the input and output of the frozen backbone, trained by gradient steps on partially observed and matured targets and scheduled over roughly period-sized batches of origins (periodicity-aware adaptation scheduling, PAAS). PETSA \citep{medeiros2025petsa} keeps the input-and-output calibration design with low-rank adapters and a modified loss. COSA \citep{im2026cosa} reduces the adapter to a single gated output-space module conditioned on a short context buffer, trained for a fixed number of gradient steps per adaptation event with an adaptive learning-rate schedule. DynaTTA \citep{grover2025dynatta} is the one input-space method; update strength depends on a measured shift on the incoming window. All four are gradient-based, and none varies the label delay.

\paragraph{Running the published methods under delay.}
We execute TAFAS, COSA, PETSA and DynaTTA from the COSA reference repository, which bundles all four, on their PatchTST backbone and training recipe, with our split and scaling. Three classes of logged patches are applied (Appendix~\ref{app:integration}): a \texttt{LABEL\_LAG} setting that shifts every maturity trigger by $\ell$ and shortens the causally observed partial-ground-truth prefix accordingly in each update loop (the identity at $\ell = 0$); a per-origin prediction export; and two compatibility fixes for current library versions. Predictions are aligned on forecast origin before scoring, so all methods, theirs and ours, are evaluated on identical tensors. DynaTTA is not run on the 862-channel Traffic stream, where its input-space update is prohibitively slow (Section~\ref{sec:limitations}).

\section{The RLS-Bank}
\label{sec:bank}

\paragraph{Per-coordinate correction.}
For channel $c$ and horizon step $h$, the corrected forecast is $\tilde y = w_{c,h}^{\top}x$ with features $x = [\hat y_{c,h},\ 1,\ y^{\mathrm{last}}_{c} - \bar y^{\mathrm{day}}_{c},\ y^{\mathrm{sn}}_{c,h} - \bar y^{\mathrm{day}}_{c}]$: the backbone forecast, an intercept, and two deviation features, the last observation and the seasonal-naive value, each centered by the trailing daily mean (the mean over the most recent day of observations, computed causally).
Centering matters: level-tracking raw features are strongly correlated, and under exponential forgetting such regressors let the RLS covariance matrix grow without bound in the directions they do not excite, the covariance wind-up that makes the gain explode. Weights are initialized at passthrough, $w = [1,0,0,0]$, so at the first step the adapter reproduces the frozen forecast.

\paragraph{RLS with anti-windup.}
When a matured sample $(x, y)$ is released, each unit performs the standard exponentially weighted RLS update with forgetting factor $\lambda$,
\begin{equation}
g = \frac{Px}{\lambda + x^{\top}Px},\qquad
w \leftarrow w + g\,(y - w^{\top}x),\qquad
P \leftarrow \frac{1}{\lambda}\left(P - g\,x^{\top}P\right),
\label{eq:rls}
\end{equation}
followed by symmetrization of $P$ and a trace clip $\operatorname{tr}(P)/F \le \rho$ with $\rho{=}10$. The clip is standard covariance limiting: it bounds the gain and prevents wind-up without introducing a learning rate. This is not only a design argument: the unclipped ablation in Appendix~\ref{app:bank} shows the clip is a small tax on the narrow streams (at most 2.8\% higher MSE) and the difference between working and exploding on the 321- and 862-channel streams. The cost per step is $O(CHF^2)$, which is 56\,$\mu$s on one CPU core for $C{=}7$, $H{=}24$ and $F{=}4$.

\paragraph{The bank, and why a median.}
The best $\lambda$ is regime-dependent; Section~\ref{sec:diagnostics} shows that the oracle forgetting factor shifts across datasets, backbones and delays. We therefore run $K{=}6$ RLS experts with $\lambda \in \{1.0,\, 0.9999,\, 0.9995,\, 0.999,\, 0.995,\, 0.99\}$ plus a frozen passthrough expert, and output the per-coordinate median of the seven predictions. Table~\ref{tab:agg} in Appendix~\ref{app:bank} compares this median with the obvious alternatives on the five DLinear streams at $d = H$. Follow-the-leader (FTL) selection by matured loss collapses to a standardized MSE of 2.75 on ETTh2 and 10.52 on Weather, worse than the frozen model, and discounting the loss ledger barely helps. The problem is not the memory of the ledger but selection lag. Under delay $d$, a hard selector keeps emitting an expert's predictions for up to $d$ steps after that expert has started to fail, and on heavy-tailed streams those episodes dominate the mean. This is the delayed-feedback pathology of \citet{weinberger2002delayed}, and it applies equally to learned mixing rules. We report the one exception with the same standards: on Electricity, where the fastest forgetting factor is best throughout, selection beats the median by 10\% because there is nothing regime-dependent to get wrong. The median needs no feedback. It is bounded by the majority of the experts at every step regardless of delay, and the wide streams supply a second, unplanned argument for it: on Electricity and Traffic, individual mid-range forgetting factors reach standardized MSE in the thousands even with the trace clip, and the per-coordinate median absorbs them, staying within 8\% of the best single $\lambda$ (Appendix~\ref{app:tables}). Its failure modes are the mirror image (six adapting experts outvote one passthrough when the frozen model is already near-optimal; the median follows a mildly poisoned majority); Section~\ref{sec:failures} reports both.

\section{Experiments}
\label{sec:experiments}

\paragraph{Scope.}
The internal grid covers 5 datasets (ETTh2, ETTm1, Weather, Electricity, Traffic) $\times$ 2 backbones (a closed-form DLinear \citep{zeng2023dlinear} and PatchTST \citep{nie2023patchtst}; seeds as in Section~\ref{sec:protocol}) $\times$ 4 delays $\times$ up to 34 methods, for 2{,}144 cached runs.
The 34 methods span the persistence and seasonal-naive baselines; the frozen backbone; simple correctors (bias-EMA, NLMS on the bank's features, single RLS filters over a forgetting-factor sweep, a calibration-only RLS); the bank with its selection-based and unclipped ablations; the ELF-style corrector of \citet{lee2025adapts} in two variants, ELF-linear and ELF-mix, run on every stream with $C \le 21$ because its cost is $O(CHL^2)$ per step; and gradient adapters (SGD, decayed SGD and Adam) on the same features and on the calibration parameterization over learning-rate grids, which serve as same-capacity stand-ins for gradient-trained TTA adapters. Appendix~\ref{app:tables} defines every run identifier and setting. All adapters on a given stream replay the same cached forecast tensors, are linear maps with one unit per (channel, horizon) coordinate except the ELF variants, and start at passthrough, so any difference between them is due to the update rule alone. The external audit runs TAFAS, COSA, PETSA and DynaTTA plus frozen exports on their own checkpoints for all five datasets, with three checkpoint seeds on ETTh2 and ETTm1.

\subsection{Controlled grid: closed-form versus oracle-tuned gradient adapters}
\label{sec:grid}

\begin{table}[t]
\caption{Internal grid at the minimum causal delay $d{=}H$ (standardized MSE; lower is better; best per row in bold). PatchTST cells on ETTh2, ETTm1 and Weather average three seeds. Oracle-GD: the best gradient adapter per cell, learning rate chosen in hindsight. ELF-mix runs on $C\le 21$ streams. Paired bootstrap intervals: Table~\ref{tab:ci_internal}.}
\label{tab:maingrid}
\begin{center}
\small
\setlength{\tabcolsep}{4pt}%
\begin{tabular}{ll ccccccc}
\toprule
Dataset & Backbone & Persist. & S-naive & Frozen & Bias-EMA & Oracle-GD & ELF-mix & RLS-Bank \\
\midrule
ETTh2 & DLinear  & 1.1196 & 1.4837 & 1.2894 & 1.4243 & 1.1343 & \textbf{1.0437} & 1.1715 \\
ETTh2 & PatchTST & 1.1196 & 1.4837 & 6.7959 & 1.3224 & 1.4578 & \textbf{1.1134} & 1.7062 \\
ETTm1 & DLinear  & 1.8007 & 0.9633 & 1.5344 & 2.6371 & 0.6006 & 0.5720 & \textbf{0.5572} \\
ETTm1 & PatchTST & 1.8007 & 0.9633 & 1.1378 & 2.0375 & 0.6656 & \textbf{0.5546} & 0.6435 \\
Weather & DLinear  & 1.0430 & 1.7684 & 1.4118 & 1.8449 & 0.6976 & \textbf{0.6774} & 0.8045 \\
Weather & PatchTST & 1.0430 & 1.7684 & 1.1153 & 0.8160 & 0.6510 & \textbf{0.5929} & 0.7067 \\
Electricity & DLinear  & 71.8317 & 6.4556 & 86.1009 & 7.5702 & 10.4222 & -- & \textbf{6.0358} \\
Electricity & PatchTST & 71.8317 & 6.4556 & 119.5988 & 9.4826 & 7.9078 & -- & \textbf{5.7072} \\
Traffic & DLinear  & 2.5740 & 0.7831 & 1.3955 & 1.3424 & 0.6594 & -- & \textbf{0.5958} \\
Traffic & PatchTST & 2.5740 & 0.7831 & 0.5410 & 0.5213 & 0.4795 & -- & \textbf{0.4707} \\
\bottomrule
\end{tabular}
\end{center}
\end{table}

Table~\ref{tab:maingrid} reports the internal grid at the minimum causal delay, and Figure~\ref{fig:delaycurves} (Appendix~\ref{app:figures}) traces the same cells across delays. Four patterns stand out.

First, one of the two closed-form correctors posts the lowest number in all ten cells, though on ETTh2 the margin over persistence is not significant. The bank matches or beats the oracle gradient adapter in six of ten cells, significantly on both ETTm1 cells, both Electricity cells and both Traffic cells under the paired bootstrap (Table~\ref{tab:ci_internal}); it is statistically tied with the oracle on ETTh2/DLinear and on both Weather cells, and loses significantly only on ETTh2/PatchTST at $d = H$. The oracle is a generous competitor: its learning rate is chosen after the fact per cell, the winning configuration changes from cell to cell, and between three and eleven of its twelve configurations diverge on every stream (6, 6, 4, 3, 6, 4, 11, 11, 7 and 5 across the ten cells, others ending at astronomically large but finite error); on Electricity every plain-SGD configuration diverges, leaving Adam as the only usable gradient optimizer. Figure~\ref{fig:fragility} (Appendix~\ref{app:figures}) contrasts this with the RLS sweep, in which no setting diverges on the 7- and 21-channel streams; on the two widest streams individual mid-range forgetting factors do blow up, and the median absorbs them (Section~\ref{sec:bank}).

Second, on ETTh2 the two families are statistically tied on the linear backbone. On the deep backbone the gradient adapter wins significantly at $d = H$ (a paired differential of $+0.38$ against the bank), and the gap is no longer significant at $d = 4H$. We report this as it is.

Third, the frozen backbones pay for the 25\% pretraining region, and we firewall this from the headline. Frozen PatchTST collapses on ETTh2 under our training recipe (6.80, with seeds between 6.53 and 7.06) while frozen DLinear scores 1.29, and on Electricity both frozen backbones are useless (86.1 and 119.6 against a seasonal-naive of 6.5; Figure~\ref{fig:drift} shows the level shift in one Electricity channel at the start of the online stream). These internal cells measure update rules on equal footing on deliberately imperfect backbones; conclusions about the published methods come only from Section~\ref{sec:headline}, where the backbones are the authors' own, and the gap between the two settings is quantified in Section~\ref{sec:diagnostics}.

Fourth, the leaky reference. At $d = 1$ several adapters appear to beat persistence on ETTh2; at $d = H$ none does, and the ranking reorders (Figures~\ref{fig:leaderboards} and~\ref{fig:delaycurves}). Table~\ref{tab:leaky} in Appendix~\ref{app:tables} quantifies the effect: the causal MSE of bias-EMA exceeds its leaky MSE by 83\% on ETTh2/DLinear and by 110\% on ETTh2/PatchTST, where the leaky number also appears to beat persistence; on ETTm1 the effect is 6 to 8\% for the single filter and the bank but 76\% for a fast-forgetting filter ($\lambda{=}0.99$). The inflation is largest on the drift-heavy data (Figure~\ref{fig:drift}).

\subsection{Published methods on their own checkpoints}
\label{sec:headline}

\begin{table}[t]
\caption{Published methods on their own PatchTST checkpoints (standardized MSE, aligned on forecast origin; best per column in bold): all five datasets at the minimum causal delay $d{=}H$, and the delay axis on the ETT streams. The RLS-Bank and the ELF correctors are replayed on the same frozen forecasts. ETT columns average three checkpoint seeds ($^{\dagger}$three seeds at $d{=}H$, seed 0 at $d{>}H$, so cross-delay comparisons in daggered rows mix seed sets); Weather, Electricity and Traffic use one checkpoint. --: not run (DynaTTA on the 862-channel Traffic stream; ELF for $C > 21$). Intervals: Tables~\ref{tab:ci_external_ettm1}--\ref{tab:ci_external_wet}; full delay tables: Table~\ref{tab:ext_full}.}
\label{tab:external}
\label{tab:extdelay}
\begin{center}
\footnotesize
\setlength{\tabcolsep}{3pt}
\begin{tabular}{l ccc ccc c c c}
\toprule
 & \multicolumn{3}{c}{ETTm1} & \multicolumn{3}{c}{ETTh2} & Weather & Electricity & Traffic \\
\cmidrule(lr){2-4}\cmidrule(lr){5-7}\cmidrule(lr){8-8}\cmidrule(lr){9-9}\cmidrule(lr){10-10}
Method & $d{=}H$ & $2H$ & $4H$ & $d{=}H$ & $2H$ & $4H$ & $d{=}H$ & $d{=}H$ & $d{=}H$ \\
\midrule
Persistence      & 1.8007 & 1.8007 & 1.8007 & 1.1196 & \textbf{1.1196} & \textbf{1.1196} & 1.0430 & 71.8355 & 2.5742 \\
Seasonal-naive   & 0.9633 & 0.9633 & 0.9633 & 1.4837 & 1.4837 & 1.4837 & 1.7683 & 6.4559 & 0.7831 \\
Frozen backbone  & 1.5238 & 1.5238 & 1.5238 & 1.3728 & 1.3728 & 1.3728 & 0.6552 & \textbf{4.7550} & 0.4134 \\
\midrule
TAFAS            & 1.4025 & 1.4025 & 1.4030 & 1.3864 & 1.4050 & 1.3986 & 0.6557 & 4.8071 & 0.4115 \\
COSA             & 1.0239 & 0.9011 & 0.9268 & 1.7343 & 1.8776 & 2.1296 & 0.6480 & 5.5247 & 0.4121 \\
PETSA$^{\dagger}$ & 1.5016 & 1.5761 & 1.5762 & 1.4410 & 1.3340 & 1.3329 & 0.7868 & 6.6216 & 0.4603 \\
DynaTTA$^{\dagger}$ & 0.6168 & 0.6922 & 0.6741 & 1.4875 & 1.4423 & 1.4501 & 0.6965 & 5.4564 & -- \\
\midrule
ELF-mix (run by us) & \textbf{0.5617} & \textbf{0.5677} & \textbf{0.5803} & \textbf{1.0649} & 1.1509 & 1.3329 & \textbf{0.6098} & -- & -- \\
RLS-Bank (ours)  & 0.6606 & 0.6648 & 0.6717 & 1.5076 & 1.5699 & 1.7455 & 0.6235 & 4.7584 & \textbf{0.4096} \\
\bottomrule
\end{tabular}
\end{center}
{\small ELF-linear (no mix): 0.5149 on ETTm1, 1.1877 on ETTh2, 0.7746 on Weather; it does not use the backbone forecast, so it is identical across backbones. Baseline rows can differ from Table~\ref{tab:maingrid} in the last digit (origin inner-join; Appendix~\ref{app:integration}).}
\end{table}

Table~\ref{tab:external} reports the published methods on their own PatchTST checkpoints, at the minimum causal delay on all five datasets and along the delay axis on the ETT streams; Table~\ref{tab:ext_full} gives the full per-dataset delay tables, and paired intervals per checkpoint seed are in Tables~\ref{tab:ci_external_ettm1}--\ref{tab:ci_external_wet}. Appendix~\ref{app:tuned} repeats TAFAS and COSA under the authors' released-script configuration; it is uniformly weaker on ETTm1 and equivalent to the frozen model on ETTh2, so every ETT conclusion below holds against the stronger configuration in each cell.

\paragraph{ETTm1 is the positive control.}
On ETTm1 all four published methods genuinely adapt: TAFAS improves on the frozen model by 0.12 in standardized MSE, COSA by 0.50, DynaTTA by 0.91 and PETSA by 0.02 (significant in every seed for the first three, in one of three seeds for PETSA). The bank improves by 0.86. It beats TAFAS by 53\%, COSA by 26 to 36\% across delays and PETSA by 56\%, with every interval far from zero in every seed. The single significant exception in the entire audit is DynaTTA, which beats the bank by 7\% at the minimum causal delay, significantly in all three seeds, at roughly 2{,}500 times the per-update cost (Section~\ref{sec:cost}); at $d = 2H$ and $d = 4H$ the two are statistically tied. The ELF-style corrector goes one further: ELF-mix is significantly better than all four published methods, including DynaTTA, at every delay and in every seed, and ELF-linear is lower still. Seasonal-naive, at 0.9633, beats TAFAS, PETSA and the frozen model on this stream. Two details deserve comment. TAFAS's MSE is invariant to three decimals across delays, so its adaptation is nearly inert to lag on this stream. COSA is better at $2H$ and $4H$ than at $H$: under the repository-default period-sized batching, its partial-ground-truth updates fire only at $\ell = 0$ and never at $\ell > 0$, because one horizon of label lag removes the entire causal prefix of a period-sized batch, and on ETTm1 those partial updates appear to hurt rather than help (Figure~\ref{fig:cosa}, Appendix~\ref{app:figures}).

\paragraph{ETTh2: one statistical cluster at the minimum delay, harm beyond it.}
On ETTh2 at $d = H$, nothing is significantly better than the frozen model or than persistence in any seed, with one exception on the closed-form side: ELF-mix beats the frozen model in two of three seeds (and is statistically tied with persistence in all three). On the published side the only consistent separation is harmful: PETSA is significantly worse than the frozen model in all three seeds, and DynaTTA in one of three. The bank, TAFAS and COSA are statistically indistinguishable from the frozen model at this delay. As the delay grows the cluster separates, and it separates on the harmful side. The bank is significantly worse than the frozen model from $d = 2H$ ($+0.16$ to $+0.26$ across seeds) and at $d = 4H$ ($+0.33$ to $+0.44$); at $d = 4H$ COSA is worse by $+0.72$ to $+0.80$ in every seed, DynaTTA by $+0.21$ and PETSA by $+0.10$ (checkpoint seed 0), and TAFAS by $+0.02$, significant in one seed and borderline in the other two. The one adapter that does not turn harmful is ELF-mix, whose inverse-error weighting hands control back to the frozen model as its own forecaster degrades; by $d = 4H$ it is statistically indistinguishable from the frozen model.

\paragraph{Weather, Electricity, Traffic: the frozen model and the closed-form correctors set the pace.}
On Weather, TAFAS is inert to four decimals ($|\Delta| \le 0.0006$ against the frozen model at every delay, n.s.), COSA is statistically tied with the frozen model at $H$ and $4H$ and significantly worse at $2H$, and PETSA ($+0.13$, 20\%) and DynaTTA ($+0.04$, 6\%) are significantly worse at every delay. The only methods significantly better than the frozen checkpoint are closed-form: the bank at $2H$ and $4H$ (about 7\%) and ELF-mix at $2H$ and $4H$ (8 to 9\%), with both n.s.\ at $H$. On Electricity the frozen checkpoint is unbeaten: all four published methods are significantly worse than it at $d = H$ (TAFAS $+0.05$, COSA $+0.77$, DynaTTA $+0.70$, PETSA $+1.87$, which is 39\% worse), while the bank is statistically tied with the frozen model at every delay, though its $4H$ interval only just includes zero. On Traffic every method sits within 12\% of the frozen model and most within 1\%: the bank is best at every delay and significantly better than the frozen model at $2H$ and $4H$ (by 1.5 to 1.8\%), TAFAS is significantly better at $H$ by 0.5\% and significantly worse at $2H$ and $4H$, COSA is tied throughout, and PETSA is significantly worse by 11\%. Taken together: outside ETTm1, the largest statistically significant improvement any published method achieves over its own frozen checkpoint, at any delay, is TAFAS's 0.5\% on Traffic, and it reverses sign once labels lag beyond the horizon.

\subsection{Cost}
\label{sec:cost}

Table~\ref{tab:cost} (Appendix~\ref{app:tables}) lists per-update wall time and adapted-parameter counts next to the accuracy results. The bank updates 4{,}032 closed-form weights (168 coordinates $\times$ 4 features $\times$ 6 experts) in 0.056\,ms per stream step on one CPU core on the 7-channel ETT streams, with no tunable hyperparameters; its cost scales as $O(CHF^2)$ and averages 2.9\,ms per step over all ten cells including the 862-channel Traffic stream. The ELF-style corrector costs about 7\,ms per step on the ETT-class streams because it regresses on the full lookback window ($O(CHL^2)$ with $L{=}96$), which is also why we restrict it to $C \le 21$. On the ETT streams, TAFAS and PETSA need roughly 5 to 8\,ms per update with the authors' default flags; COSA performs 20 gradient steps of about 28\,ms each per adaptation event; DynaTTA needs 139 to 158\,ms per update and 438\,s for one pass over the ETTm1 stream. Per update, the bank is therefore two to four orders of magnitude cheaper than every gradient method it is compared with on those streams (amortized per-step gaps are smaller, since update frequencies differ; Table~\ref{tab:cost}), a gap to read together with Table~\ref{tab:external}, where most accuracy differences favor the closed-form correctors or are not significant.

\subsection{Diagnostics: label staleness, plasticity and the training recipe}
\label{sec:diagnostics}

Figure~\ref{fig:staleness} in Appendix~\ref{app:figures} tracks the oracle learning rate of the gradient adapters and the oracle RLS memory $1/(1-\lambda)$ as a function of delay. Optimal plasticity falls as labels get staler: the winning learning rate shrinks with $d$ (on a decade-spaced grid, so sub-decade movement does not register) and the winning forgetting factor rises. The bank follows this shift without being told; its MSE sits within 7\% of the per-cell oracle $\lambda$ in nine of ten cells (from 0.7\% better to 6.6\% worse), with one outlier, Weather/DLinear at $d = H$, where the bank lands 24\% above the best single $\lambda$ before the gap closes by $d = 4H$ (Section~\ref{sec:failures}).

Two findings at the system level are larger than any method-level effect in this paper. The first is recipe sensitivity. The same PatchTST architecture on the same stream, split and scaling scores 6.80 frozen under our training recipe and 1.37 under the authors' recipe on ETTh2 (a factor of 5), and 119.6 against 4.76 on Electricity (a factor of 25); on ETTm1 the sign flips, with our recipe frozen at 1.14 against the authors' 1.52. The training recipe moves frozen online error by more than any adaptation effect we measure, in either direction. The second is a ranking inversion. On ETTm1, frozen PatchTST beats frozen DLinear (1.138 vs.\ 1.534), but DLinear with the bank beats PatchTST with the bank (0.557 vs.\ 0.644): adaptation changes which backbone one should deploy. The full delay curves, per-horizon error profiles, cumulative-error curves and a drift overview of the splits are in Appendix~\ref{app:figures}.

\section{When Closed-Form Adaptation Fails}
\label{sec:failures}

We report our own negative results with the same standards. \textbf{The bank turns harmful on ETTh2 at extended delay.} On ETTh2 with a well-trained frozen backbone the bank is harmful: significantly so at $d = 2H$ ($+0.16$ to $+0.26$ across the three checkpoint seeds) and at $d = 4H$ ($+0.33$ to $+0.44$). The median guards against a minority of exploding experts, not against a majority of mildly harmful ones; with six adapting experts and one passthrough, adaptation wins the vote even when it should not. The failure is not unique to us, since COSA, DynaTTA and PETSA are also significantly worse than the frozen model there at extended delay, but it is ours too. Anchoring the aggregate to the frozen forecast is the obvious repair, and the audit contains direct evidence that it works: ELF-mix, whose online weighter is exactly such an anchor, converges to the frozen model on this stream instead of turning harmful. \textbf{The median follows a poisoned majority on Weather/DLinear.} The Weather/DLinear cell at $d = H$ exhibits the median's other failure mode: most experts are mildly poisoned by the regime, the median follows the majority, and the bank sits 24\% above the best single forgetting factor before recovering at longer delays; the episode is visible in the Weather/DLinear panel of Figure~\ref{fig:cumulative}. \textbf{A hindsight-tuned gradient method can win on heavy tails.} On ETTh2/PatchTST under our training recipe at $d = H$, plain SGD beats RLS significantly ($+0.38$ in favor of SGD). Where the residual process has very heavy tails, per-direction gain control is not enough, and a gradient method with a hindsight-selected learning rate can win. \textbf{The closest published relative beats us on two of the three streams where it runs.} The ELF-style corrector is significantly better than the bank on ETTm1 (by about 0.10 to 0.16) and on ETTh2 (by 0.13 to 0.61 across seeds and delays), and statistically tied with it on Weather at the minimum delay. The bank's remaining advantages are cost (roughly two orders of magnitude, Section~\ref{sec:cost}) and applicability to the wide streams where the ELF-style corrector is impractical and the bank posts its clearest wins. \textbf{The closed-form advantage is robust to delay, not increasing in it.} Our initial hypothesis was that the advantage would grow with delay. It does not (Figure~\ref{fig:delaycurves}).

\section{Limitations, Recommendations and Conclusion}
\label{sec:conclusion}

\paragraph{Limitations.}
\label{sec:limitations}
We evaluate five dataset families but one deep architecture (PatchTST), the backbone all four audited methods release checkpoints for. DynaTTA is not run on the 862-channel Traffic stream (prohibitively slow input-space update); PETSA and DynaTTA use three checkpoint seeds at $d = H$ and one at $d > H$; the released-script re-pass (Appendix~\ref{app:tuned}) covers the ETT streams only; Weather, Electricity and Traffic use a single checkpoint and a reduced external method set, for compute reasons. The ELF-style corrector is restricted to $C \le 21$ by its $O(CHL^2)$ cost, so the wide streams lack the second closed-form reference. External methods are reported under both the repository-default and the released-script configurations where both exist; the release contains no per-dataset tuning. Absolute numbers are not comparable across papers, because our 25/5/70 split is deliberately nonstandard (Section~\ref{sec:protocol}).

\paragraph{Recommendations.}
We distill the audit into five recommendations for evaluating TTA in forecasting. (a) Never update on unmatured windows; report $d = H$ as the primary setting and at least one $d > H$. (b) Report persistence, seasonal-naive and the frozen model in the same table as the adapted models. (c) Include closed-form correctors as mandatory baselines: the bank is about 30 lines of NumPy at 56\,$\mu$s per step on ETT-class streams, and an ELF-style corrector is an even stronger reference where the channel count permits. (d) Report adaptation cost next to accuracy. (e) Report the backbone training recipe, which in our experiments moves online error by up to a factor of 25, more than any adaptation method. Under these rules, on the five benchmarks studied, closed-form correctors set a bar that the audited stack clears exactly once: DynaTTA, on ETTm1, at the minimum causal delay, at roughly 2{,}500 times the per-update cost, and even there an ELF-style corrector remains ahead. Everywhere else, what separates methods is evaluation practice, not adaptation machinery.

\subsection*{AI use statement}
In this work, we used generative AI tools (Claude by Anthropic and DeepSeek) for the following tasks with required disclosure: implementing methods, namely the experimental harness, the analysis code and the integration patches for the baseline repositories. We have not used generative AI tools to generate synthetic datasets, to develop theoretical models or conceptual frameworks, to formulate or prove mathematical claims, to propose or refine hypotheses, to design the experiments, to clean or reformat datasets, or to interpret results; the remaining tasks with required disclosure are not applicable to this work. Additionally, we used generative AI tools for tasks with recommended disclosure: creating and editing software code, creating scientific figures, drafting parts of the manuscript, editing the manuscript to improve readability, and formatting references. We have reviewed all AI-assisted work. All code was read and tested by the authors; the internal pipeline was reproduced across three machines to within $10^{-4}$; every number in the paper was checked against the cached run outputs; and all text was revised by the authors. Research direction, experimental design, execution of all experiments and verification of all results were carried out by the authors. We take responsibility for the final content of this work, including text, claims and artifacts produced with the aid of generative AI.

\subsection*{Ethics statement}
This work re-audits published forecasting methods on public benchmarks (the ETT datasets and the Weather, Electricity and Traffic datasets of the standard long-term-forecasting suite), reuses the authors' released code and checkpoints under their stated licenses (CC BY-NC-SA 4.0 for the COSA reference repository), and introduces no new datasets, human subjects or deployed systems. We report negative findings about existing methods and about our own method with the same standards.

\subsection*{Reproducibility statement}
All internal and external runs are cached and keyed by configuration hash, with the exact run manifest shipped in the release; the pipeline is deterministic given the backbone and was reproduced across three machines to $10^{-4}$. We release the harness notebook, the integration kit (the patcher with its full change log and the runner with alignment asserts), the stream specifications and all cached results.
These artifacts are publicly available at \url{https://github.com/Xodios/TRAINING-ON-THE-FUTURE-A-DELAY-AWARE-AUDIT-OF-TEST-TIME-ADAPTATION-FOR-TIME-SERIES-FORECASTING}. Appendix~\ref{app:integration} logs every deviation from the reference repository, Appendix~\ref{app:bank} gives the pseudocode of the RLS-Bank and the ELF-style corrector's configuration, and Appendix~\ref{app:bootstrap} describes the bootstrap. The COSA reference repository is CC BY-NC-SA 4.0; we distribute patches, not modified code.

\bibliography{iclr2027_conference}
\bibliographystyle{iclr2027_conference}

\appendix

\section{Integration and Deviations Log}
\label{app:integration}
We ran TAFAS, COSA, PETSA and DynaTTA from the COSA reference repository, which bundles all four, on their own PatchTST, their own training recipe, and our split and standardization, on all five datasets. Every deviation from the stock repository is logged here.

\subsection{Dependency incompatibility}
The reference repository's \texttt{requirements.txt} pins \texttt{numpy}~1.23.5 and \texttt{torch}~1.7.1, a snapshot of the authors' 2024 environment. On our Python~3.12 stack with CUDA \texttt{torch}~2.12 these pins do not install: \texttt{numpy}~1.23.5 has no Python~3.12 wheel and its source build fails (the pulled-in \texttt{setuptools} calls \texttt{pkgutil.ImpImporter}, removed in Python~3.12), and the \texttt{torch} pin would replace the container's CUDA runtime. We derived the minimal set of actually imported dependencies by grepping the repository's imports and installed only that set, leaving \texttt{numpy}, \texttt{pandas}, \texttt{matplotlib}, \texttt{scipy} and \texttt{torch} untouched:
\begin{quote}
\raggedright\texttt{pip install yacs wandb scikit-learn einops local-attention reformer-pytorch tqdm}
\end{quote}
All modules used by our runner import cleanly on this stack.

\subsection{Patches}
A single script, \texttt{apply\_patches.py}, applies every code change; it is idempotent and prints one line per patch. Its full output on a fresh clone, and the all-\texttt{[skip]} output of a second invocation that confirms idempotence, are included in the code release.
There are three classes of changes.

\paragraph{(i) \texttt{LABEL\_LAG}: delay-aware maturity and partial-ground-truth semantics (Section~\ref{sec:protocol}).}
An environment variable gives the label lag $\ell = d - H$. In each of the four TTA modules the maturity trigger is shifted by \texttt{+ LABEL\_LAG} and the causally observed partial-ground-truth prefix is reduced by \texttt{LABEL\_LAG} (guarded when it reaches zero). At $\ell = 0$ the patched behavior is bit-identical to the authors' code.

\paragraph{(ii) \texttt{EXPORT\_NPZ}: per-origin prediction export.}
When set, each adapt loop accumulates per-batch predictions and writes an \texttt{.npz} with array \texttt{preds} of shape $[T, C, H]$ in the standardized output space at teardown. This is the sole mechanism by which our scoring pipeline consumes their outputs; we never re-run their models inside our own harness.

\paragraph{(iii) Compatibility fixes.}
Two parts of the reference code fail on current library versions. First, \texttt{datasets/build.py} calls \texttt{Series.apply(lambda row: row.month, 1)} for four calendar features; the deprecated positional argument raises \texttt{TypeError} on pandas$\geq$2. We removed the trailing argument in all call sites (numerically identical). Second, the non-stationary normalization writes \texttt{enc\_window /= stdev} (\texttt{models/forecast.py}) and \texttt{x\_enc /= stdev} (\texttt{models/PatchTST.py}, \texttt{models/normalize.py}) in place. In-place division on a grad-tracked input tensor trips autograd's version counter for any input-space TTA method, and DynaTTA, the one input-space method in the stack, failed on the reference backbone with \texttt{RuntimeError: one of the variables needed for gradient computation has been modified by an inplace operation}. Rewriting the divisions out of place (\texttt{x = x / stdev}; numerically identical on the forward pass) resolved the error. A post-patch re-verification run reproduced the originally reported MSE within 1.1\% on ETTh2 (1.3671 vs.\ 1.3822) and 0.05\% on ETTm1 (0.6191 vs.\ 0.6188), completed all 972\,/\,3{,}900 adaptation events without autograd errors, and confirmed 11{,}196 trainable parameters. Two additional in-place divisions of the same class exist in \texttt{models/iTransformer.py}; we patched them defensively, although the PatchTST code path does not exercise them. We record the incident because any input-space TTA method run on a current PyTorch stack will encounter it.

\subsection{Checkpoint path collision (discarded first external batch)}
The reference repository writes checkpoints to a single shared \texttt{results/checkpoint\_best.pth}. Our first external batch pretrained ETTm1 after ETTh2 against that shared path, so ETTm1's training silently overwrote ETTh2's weights, and the earlier weights could not be reconstructed exactly (cuDNN training is not bit-deterministic). We discarded the entire first batch and modified the runner to use per-(dataset, seed) checkpoint directories with an existence assertion. All external results reported in this paper come from the clean re-run; the three-seed ETT checkpoints and the single-seed Weather, Electricity and Traffic checkpoints all use this layout.

\subsection{Origin alignment}
The repository's ratio arithmetic yields one more online window than our harness (for example 12{,}171 vs.\ 12{,}170 on ETTh2 and 48{,}753 vs.\ 48{,}752 on ETTm1). Exported files carry per-window origin indices, and scoring inner-joins on origin before computing metrics, so every method, theirs and ours, is evaluated on identical tensors; the same inner-join is applied on the Weather, Electricity and Traffic streams.

\subsection{Data layout}
The reference loader expects \texttt{data/<dataset>/<name>.csv}; our harness stores flat \texttt{data/<name>.csv}. The runner detects a flat file and creates the expected layout on first use. The ETT CSVs are fetched from the ETDataset repository; the Weather, Electricity and Traffic CSVs come from the standard Autoformer / Time-Series-Library dataset release \citep{wu2021autoformer} and are auto-fetched or dropped into \texttt{data/} manually.

\subsection{License and attribution}
The COSA reference repository is released under CC BY-NC-SA 4.0. We distribute patches and integration scripts, not modified copies of the repository.

\subsection{Nowcasting connection}
Our label-lag semantics (inputs always current, labels arriving $\ell$ steps late) matches the ragged-edge structure of econometric nowcasting \citep[cf.\ MIDAS regressions]{giannone2008nowcasting}. That literature addresses training-time estimation under mixed frequencies; our setting (test-time adaptation of a pretrained forecaster) and solution (a closed-form corrector) are orthogonal, so we note the shared formal object without a numerical comparison.

\begin{figure}[H]
\begin{center}
\includegraphics[width=\linewidth,height=0.42\textheight,keepaspectratio]{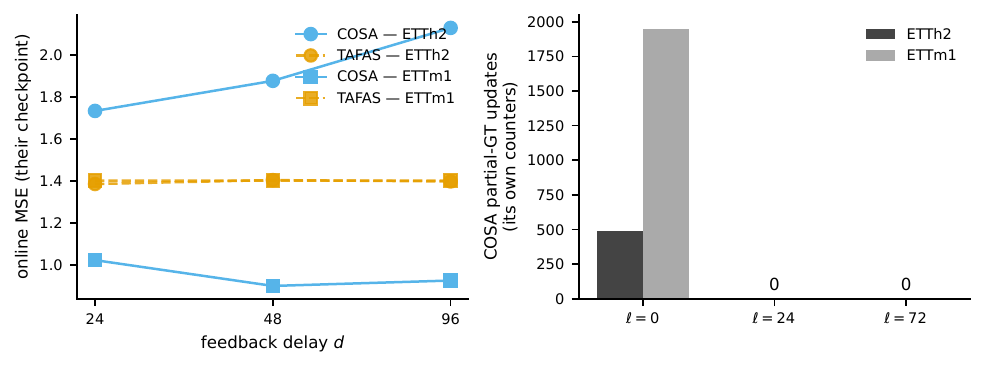}
\end{center}
\caption{COSA and TAFAS under delayed labels on their own ETT checkpoints. \figleft{} Online MSE versus delay. \figright{} Number of partial-ground-truth updates performed by COSA, from its own counters, under the repository-default period-sized batch of about 25 origins: 487 (ETTh2) and 1{,}951 (ETTm1) at $\ell{=}0$, and none at $\ell{>}0$. Counts for the released-script configuration are given in Appendix~\ref{app:tuned}.}
\label{fig:cosa}
\end{figure}

\paragraph{Cross-check of the Figure~\ref{fig:cosa} counters.}
DynaTTA's run report includes \texttt{adaptation\_rate\_history\_length} $=$ 487 (ETTh2) and 1{,}951 (ETTm1), identical to COSA's partial-ground-truth counters, since both derive from the stream's count of partial-window opportunities. This is an independent confirmation, from a different method, of the Figure~\ref{fig:cosa} values.
\clearpage

\section{Full Result Tables}
\label{app:tables}
Method identifiers in this appendix are the names of the cached runs. \texttt{naive} and \texttt{snaive} are persistence and seasonal-naive; \texttt{frozen} is the unadapted backbone. \texttt{bias\_ema} adds an exponential moving average ($\beta{=}0.05$) of the matured residual to the forecast. \texttt{nlms\_feat} is normalized LMS on the four bank features with step size $\mu{=}0.1$ and a small leak ($10^{-3}$) toward passthrough; \texttt{nlms\_feat\_mu$\mu$} varies the step size. \texttt{rls\_feat} is a single RLS filter with $\lambda{=}0.999$ on the same features, \texttt{rls\_feat\_l$\lambda$} its forgetting-factor variants, and \texttt{rls\_calib} the calibration-only variant with features $[\hat y, 1]$, a scale and shift on the backbone forecast. \texttt{rls\_bank\_med} is the RLS-Bank; \texttt{rls\_bank\_ftl} and \texttt{rls\_bank\_dftl} replace the median by follow-the-leader and discounted follow-the-leader ($\gamma{=}0.99$) selection. \texttt{rls\_feat\_noclip} and \texttt{rls\_bank\_med\_noclip} disable the trace clip (Appendix~\ref{app:bank}). \texttt{elf\_linear} and \texttt{elf\_mix} are the ELF-style corrector of \citet{lee2025adapts} re-implemented in the harness: a per-(channel, horizon) linear forecaster fit by growing-window recursive least squares ($\lambda{=}1$) on the raw lookback window, alone (\texttt{elf\_linear}) or combined with the frozen forecast by an online inverse-error weighter with EW-MSE memory $\beta{=}0.99$ (\texttt{elf\_mix}); both update only on released labels, and both are run on streams with $C \le 21$. \texttt{gd\_feat\_*} and \texttt{gd\_calib\_*} are gradient adapters on the feature and calibration parameterizations with plain SGD (\texttt{sgd}), SGD with a decaying step size (\texttt{sgddec}, $\eta_t = \eta_0/\sqrt{1+t/\tau}$ with $\tau{=}1000$) or Adam (\texttt{adam}) at the stated initial learning rate. $\infty$ marks a run that diverged; -- marks a method not run on that stream.

Table~\ref{tab:leaky} isolates the leaky-protocol effect in three cells. Tables~\ref{tab:full_etth2_dl}--\ref{tab:full_traffic_pt} list all methods at all four delays for the ten internal cells, exactly as cached (from \texttt{table\_\{ds\}\_\{backbone\}.csv}). Table~\ref{tab:ext_full} lists the external runs on the published methods' own checkpoints for all five datasets, including our auxiliary adapters replayed on their frozen exports (from \texttt{headline\_*.csv}).

\begin{table}[H]\centering\small
\caption{Leaky-protocol inflation in selected cells: standardized MSE at $d{=}1$ (leaky) and $d{=}H$ (causal), and the relative increase. The full $d{=}1$ columns for every method are in Tables~\ref{tab:full_etth2_dl}--\ref{tab:full_traffic_pt}.}
\label{tab:leaky}
\begin{tabular}{llccc}
\toprule
Cell & Method & $d{=}1$ (leaky) & $d{=}H$ (causal) & Increase \\
\midrule
ETTh2 / DLinear  & bias-EMA & 0.78 & 1.42 & $+83\%$ \\
ETTh2 / PatchTST & bias-EMA & 0.63 & 1.32 & $+110\%$ \\
ETTm1 / PatchTST & RLS ($\lambda{=}0.99$) & 0.46 & 0.81 & $+76\%$ \\
\bottomrule
\end{tabular}
\end{table}

{\scriptsize
\setlength{\tabcolsep}{3pt}
\begin{longtable}[c]{l rrrr rrrr}
\caption{Full internal grid, ETTh2: standardized online MSE at each label-release delay for the DLinear backbone (left) and the PatchTST backbone (right, 3-seed mean). $d{=}1$ is the leaky reference. $\infty$: diverged run; --: not run on this stream.}
\label{tab:full_etth2_dl}
\label{tab:full_etth2_pt}\\
\toprule
 & \multicolumn{4}{c}{DLinear} & \multicolumn{4}{c}{PatchTST (3 seeds)} \\
\cmidrule(lr){2-5}\cmidrule(lr){6-9}
method & $d{=}1$ & $d{=}24$ & $d{=}48$ & $d{=}96$ & $d{=}1$ & $d{=}24$ & $d{=}48$ & $d{=}96$ \\
\midrule
\endfirsthead
\toprule
 & \multicolumn{4}{c}{DLinear} & \multicolumn{4}{c}{PatchTST (3 seeds)} \\
\cmidrule(lr){2-5}\cmidrule(lr){6-9}
method & $d{=}1$ & $d{=}24$ & $d{=}48$ & $d{=}96$ & $d{=}1$ & $d{=}24$ & $d{=}48$ & $d{=}96$ \\
\midrule
\endhead
\midrule
\multicolumn{9}{r}{\scriptsize continues on next page}\\
\endfoot
\bottomrule
\endlastfoot
naive & 1.1196 & 1.1196 & 1.1196 & 1.1196 & 1.1196 & 1.1196 & 1.1196 & 1.1196 \\
snaive & 1.4837 & 1.4837 & 1.4837 & 1.4837 & 1.4837 & 1.4837 & 1.4837 & 1.4837 \\
frozen & 1.2894 & 1.2894 & 1.2894 & 1.2894 & 6.7959 & 6.7959 & 6.7959 & 6.7959 \\
bias\_ema & 0.7805 & 1.4243 & 1.5318 & 1.7152 & 0.6288 & 1.3224 & 1.6028 & 2.1637 \\
\addlinespace
nlms\_feat\_mu0.02 & 5.3040 & 8.6516 & 9.9049 & 12.8339 & 1.6666 & 2.6994 & 2.6393 & 2.8256 \\
nlms\_feat\_mu0.05 & 8.3202 & 21.1996 & 24.6913 & 32.8699 & 2.8070 & 7.7442 & 7.2284 & 6.4395 \\
nlms\_feat & 8.8036 & 37.9675 & 46.6960 & 59.7205 & 3.1443 & 15.2230 & 15.1574 & 12.5984 \\
nlms\_feat\_mu0.3 & 4.5275 & 65.7587 & 85.2619 & 102.8336 & 2.0671 & 26.9198 & 31.5785 & 25.5694 \\
\addlinespace
rls\_feat & 1.0027 & 1.1668 & 1.2681 & 1.4213 & 1.3599 & 1.6975 & 1.8948 & 2.3403 \\
rls\_bank\_med & 1.0054 & 1.1715 & 1.2668 & 1.4168 & 1.3586 & 1.7062 & 1.9069 & 2.3475 \\
rls\_calib & 1.1536 & 1.4909 & 1.7381 & 2.0945 & 1.4451 & 1.9314 & 2.0144 & 2.4220 \\
\addlinespace
rls\_feat\_l1 & 1.0735 & 1.2313 & 1.3216 & 1.4690 & 1.6418 & 1.9805 & 2.1832 & 2.6158 \\
rls\_feat\_l0.9999 & 1.0626 & 1.2201 & 1.3108 & 1.4585 & 1.5833 & 1.9213 & 2.1235 & 2.5576 \\
rls\_feat\_l0.9995 & 1.0283 & 1.1863 & 1.2804 & 1.4298 & 1.4399 & 1.7761 & 1.9762 & 2.4164 \\
rls\_feat\_l0.995 & 0.9188 & 1.5043 & 1.8253 & 2.1069 & 1.1225 & 1.6062 & 1.8373 & 2.3229 \\
rls\_feat\_l0.99 & 0.9240 & 4.2506 & 6.1074 & 7.3959 & 1.1065 & 2.9258 & 3.8350 & 3.9512 \\
\addlinespace
rls\_feat\_noclip & 0.9975 & 1.1627 & 1.2649 & 1.4189 & 1.3104 & 1.6497 & 1.8481 & 2.2970 \\
rls\_bank\_med\_noclip & 1.0026 & 1.1686 & 1.2642 & 1.4146 & 1.3214 & 1.6692 & 1.8696 & 2.3124 \\
rls\_bank\_ftl & 0.9262 & 2.7495 & 4.3654 & 5.5583 & 1.1025 & 2.2355 & 2.9923 & 3.1724 \\
rls\_bank\_dftl & 0.9135 & 2.7070 & 4.2985 & 5.4730 & 1.0557 & 2.1928 & 2.9448 & 3.1552 \\
\addlinespace
elf\_linear & 0.9314 & 1.1877 & 1.2672 & 1.4442 & 0.9314 & 1.1877 & 1.2672 & 1.4442 \\
elf\_mix & 0.9210 & 1.0437 & 1.0748 & 1.1482 & 0.9274 & 1.1134 & 1.1948 & 1.3594 \\
\addlinespace
gd\_feat\_sgd0.0001 & 0.9650 & 1.1690 & 1.2070 & 1.2582 & 1.5500 & 1.8602 & 2.0297 & 2.3595 \\
gd\_feat\_sgd0.001 & 13.5129 & 10.0740 & 9.7203 & 9.5466 & 0.9047 & 1.4578 & 1.6685 & 2.0962 \\
gd\_feat\_sgd0.01 & $\infty$ & $\infty$ & $\infty$ & $\infty$ & $\infty$ & $\infty$ & $\infty$ & $\infty$ \\
gd\_feat\_sgd0.1 & $\infty$ & $\infty$ & $\infty$ & $\infty$ & $\infty$ & $\infty$ & $\infty$ & $\infty$ \\
gd\_feat\_sgddec0.001 & 0.8022 & 1.1343 & 1.1946 & 1.3094 & 1.0297 & 1.5377 & 1.7353 & 2.1329 \\
gd\_feat\_sgddec0.01 & $\infty$ & $\infty$ & $\infty$ & $\infty$ & $\infty$ & $\infty$ & $\infty$ & $\infty$ \\
gd\_feat\_sgddec0.1 & $\infty$ & $\infty$ & $\infty$ & $\infty$ & $\infty$ & $\infty$ & $\infty$ & $\infty$ \\
gd\_feat\_adam0.01 & 0.9359 & 1.3118 & 1.6179 & 2.0543 & 1.1606 & 1.5850 & 1.9620 & 2.6619 \\
gd\_calib\_sgd0.0001 & 1.0200 & 1.2406 & 1.2855 & 1.3579 & 1.6268 & 1.8487 & 2.0215 & 2.3528 \\
gd\_calib\_sgd0.001 & 0.7907 & 1.2512 & 1.4117 & 1.6248 & 0.9673 & 1.5134 & 1.7142 & 2.1366 \\
gd\_calib\_sgd0.01 & $\infty$ & $\infty$ & $\infty$ & $\infty$ & $\infty$ & $\infty$ & $\infty$ & $\infty$ \\
gd\_calib\_sgd0.1 & $\infty$ & $\infty$ & $\infty$ & $\infty$ & $\infty$ & $\infty$ & $\infty$ & $\infty$ \\
\end{longtable}
}

{\scriptsize
\setlength{\tabcolsep}{3pt}
\begin{longtable}[c]{l rrrr rrrr}
\caption{Full internal grid, ETTm1: standardized online MSE at each label-release delay for the DLinear backbone (left) and the PatchTST backbone (right, 3-seed mean). $d{=}1$ is the leaky reference. $\infty$: diverged run; --: not run on this stream.}
\label{tab:full_ettm1_dl}
\label{tab:full_ettm1_pt}\\
\toprule
 & \multicolumn{4}{c}{DLinear} & \multicolumn{4}{c}{PatchTST (3 seeds)} \\
\cmidrule(lr){2-5}\cmidrule(lr){6-9}
method & $d{=}1$ & $d{=}24$ & $d{=}48$ & $d{=}96$ & $d{=}1$ & $d{=}24$ & $d{=}48$ & $d{=}96$ \\
\midrule
\endfirsthead
\toprule
 & \multicolumn{4}{c}{DLinear} & \multicolumn{4}{c}{PatchTST (3 seeds)} \\
\cmidrule(lr){2-5}\cmidrule(lr){6-9}
method & $d{=}1$ & $d{=}24$ & $d{=}48$ & $d{=}96$ & $d{=}1$ & $d{=}24$ & $d{=}48$ & $d{=}96$ \\
\midrule
\endhead
\midrule
\multicolumn{9}{r}{\scriptsize continues on next page}\\
\endfoot
\bottomrule
\endlastfoot
naive & 1.8007 & 1.8007 & 1.8007 & 1.8007 & 1.8007 & 1.8007 & 1.8007 & 1.8007 \\
snaive & 0.9633 & 0.9633 & 0.9633 & 0.9633 & 0.9633 & 0.9633 & 0.9633 & 0.9633 \\
frozen & 1.5344 & 1.5344 & 1.5344 & 1.5344 & 1.1378 & 1.1378 & 1.1378 & 1.1378 \\
bias\_ema & 1.2269 & 2.6371 & 2.2282 & 1.5896 & 0.8316 & 2.0375 & 1.8445 & 1.0218 \\
\addlinespace
nlms\_feat\_mu0.02 & 0.5267 & 0.6525 & 0.6459 & 0.6695 & 0.5662 & 0.7499 & 0.7358 & 0.7490 \\
nlms\_feat\_mu0.05 & 0.4542 & 0.7987 & 0.7859 & 0.8724 & 0.4778 & 0.9447 & 0.9236 & 0.9351 \\
nlms\_feat & 0.3802 & 1.1123 & 1.0274 & 1.2067 & 0.3917 & 1.2836 & 1.2642 & 1.2181 \\
nlms\_feat\_mu0.3 & 0.2511 & 2.1301 & 1.7310 & 1.8884 & 0.2676 & 2.2483 & 2.2861 & 1.8016 \\
\addlinespace
rls\_feat & 0.5281 & 0.5590 & 0.5610 & 0.5659 & 0.6042 & 0.6509 & 0.6552 & 0.6595 \\
rls\_bank\_med & 0.5194 & 0.5572 & 0.5580 & 0.5612 & 0.5947 & 0.6435 & 0.6461 & 0.6497 \\
rls\_calib & 0.5891 & 0.6054 & 0.6035 & 0.6093 & 1.0953 & 1.1267 & 1.1070 & 1.1058 \\
\addlinespace
rls\_feat\_l1 & 0.5806 & 0.5851 & 0.5856 & 0.5871 & 0.6503 & 0.6560 & 0.6564 & 0.6577 \\
rls\_feat\_l0.9999 & 0.5687 & 0.5754 & 0.5760 & 0.5777 & 0.6417 & 0.6506 & 0.6513 & 0.6528 \\
rls\_feat\_l0.9995 & 0.5435 & 0.5609 & 0.5622 & 0.5654 & 0.6209 & 0.6468 & 0.6489 & 0.6519 \\
rls\_feat\_l0.995 & 0.4704 & 0.6090 & 0.6103 & 0.6234 & 0.5253 & 0.7120 & 0.7257 & 0.7425 \\
rls\_feat\_l0.99 & 0.4208 & 0.7084 & 0.7050 & 0.7174 & 0.4581 & 0.8057 & 0.8221 & 0.8494 \\
\addlinespace
rls\_feat\_noclip & 0.5281 & 0.5590 & 0.5610 & 0.5659 & 0.6042 & 0.6509 & 0.6552 & 0.6595 \\
rls\_bank\_med\_noclip & 0.5194 & 0.5572 & 0.5580 & 0.5612 & 0.5947 & 0.6435 & 0.6461 & 0.6497 \\
rls\_bank\_ftl & 0.4208 & 0.7084 & 0.7050 & 0.7174 & 0.4581 & 0.8057 & 0.8221 & 0.8494 \\
rls\_bank\_dftl & 0.4219 & 0.7074 & 0.7039 & 0.7160 & 0.4588 & 0.8049 & 0.8214 & 0.8488 \\
\addlinespace
elf\_linear & 0.5120 & 0.5149 & 0.5159 & 0.5194 & 0.5120 & 0.5149 & 0.5159 & 0.5194 \\
elf\_mix & 0.5625 & 0.5720 & 0.5667 & 0.5751 & 0.5307 & 0.5546 & 0.5590 & 0.5669 \\
\addlinespace
gd\_feat\_sgd0.0001 & 0.6007 & 0.6202 & 0.6182 & 0.6201 & 0.6645 & 0.6782 & 0.6781 & 0.6805 \\
gd\_feat\_sgd0.001 & 0.4893 & 0.6516 & 0.6522 & 0.6617 & 0.5772 & 0.7043 & 0.7037 & 0.7194 \\
gd\_feat\_sgd0.01 & $\infty$ & $\infty$ & $\infty$ & $\infty$ & 0.3802 & 1.0391 & 1.0480 & 1.0678 \\
gd\_feat\_sgd0.1 & $\infty$ & $\infty$ & $\infty$ & $\infty$ & $\infty$ & $\infty$ & $\infty$ & $\infty$ \\
gd\_feat\_sgddec0.001 & 0.5588 & 0.6006 & 0.5973 & 0.6010 & 0.6357 & 0.6656 & 0.6653 & 0.6698 \\
gd\_feat\_sgddec0.01 & 0.4340 & 0.7165 & 0.7251 & 0.7407 & 0.5204 & 0.7626 & 0.7646 & 0.7845 \\
gd\_feat\_sgddec0.1 & $\infty$ & $\infty$ & $\infty$ & $\infty$ & $\infty$ & $\infty$ & $\infty$ & $\infty$ \\
gd\_feat\_adam0.01 & 0.5963 & 0.9036 & 0.9367 & 0.9636 & 0.6467 & 0.9395 & 0.9797 & 1.0833 \\
gd\_calib\_sgd0.0001 & 0.6583 & 0.6717 & 0.6689 & 0.6712 & 1.1120 & 1.1203 & 1.1135 & 1.1132 \\
gd\_calib\_sgd0.001 & 0.5557 & 0.6751 & 0.6600 & 0.6435 & 1.0966 & 1.1895 & 1.1257 & 1.1096 \\
gd\_calib\_sgd0.01 & 1.1374 & 0.9947 & 1.0251 & 1.2121 & 0.9390 & 2.1505 & 1.8529 & 1.1270 \\
gd\_calib\_sgd0.1 & $\infty$ & $\infty$ & $\infty$ & $\infty$ & $\infty$ & $\infty$ & $\infty$ & $\infty$ \\
\end{longtable}
}

{\scriptsize
\setlength{\tabcolsep}{3pt}
\begin{longtable}[c]{l rrrr rrrr}
\caption{Full internal grid, Weather: standardized online MSE at each label-release delay for the DLinear backbone (left) and the PatchTST backbone (right, 3-seed mean). $d{=}1$ is the leaky reference. $\infty$: diverged run; --: not run on this stream.}
\label{tab:full_weather_dl}
\label{tab:full_weather_pt}\\
\toprule
 & \multicolumn{4}{c}{DLinear} & \multicolumn{4}{c}{PatchTST (3 seeds)} \\
\cmidrule(lr){2-5}\cmidrule(lr){6-9}
method & $d{=}1$ & $d{=}24$ & $d{=}48$ & $d{=}96$ & $d{=}1$ & $d{=}24$ & $d{=}48$ & $d{=}96$ \\
\midrule
\endfirsthead
\toprule
 & \multicolumn{4}{c}{DLinear} & \multicolumn{4}{c}{PatchTST (3 seeds)} \\
\cmidrule(lr){2-5}\cmidrule(lr){6-9}
method & $d{=}1$ & $d{=}24$ & $d{=}48$ & $d{=}96$ & $d{=}1$ & $d{=}24$ & $d{=}48$ & $d{=}96$ \\
\midrule
\endhead
\midrule
\multicolumn{9}{r}{\scriptsize continues on next page}\\
\endfoot
\bottomrule
\endlastfoot
naive & 1.0430 & 1.0430 & 1.0430 & 1.0430 & 1.0430 & 1.0430 & 1.0430 & 1.0430 \\
snaive & 1.7684 & 1.7684 & 1.7684 & 1.7684 & 1.7684 & 1.7684 & 1.7684 & 1.7684 \\
frozen & 1.4118 & 1.4118 & 1.4118 & 1.4118 & 1.1153 & 1.1153 & 1.1153 & 1.1153 \\
bias\_ema & 1.0728 & 1.8449 & 2.2388 & 1.6217 & 0.5079 & 0.8160 & 0.9616 & 0.8695 \\
\addlinespace
nlms\_feat\_mu0.02 & 0.7444 & 0.8416 & 0.8319 & 0.8006 & 0.6805 & 0.8029 & 0.8285 & 0.7970 \\
nlms\_feat\_mu0.05 & 1.0292 & 1.3659 & 1.3811 & 1.2596 & 0.9620 & 1.3484 & 1.4591 & 1.3909 \\
nlms\_feat & 1.3391 & 2.2143 & 2.3541 & 2.2231 & 1.2414 & 2.1662 & 2.4281 & 2.5649 \\
nlms\_feat\_mu0.3 & 1.7314 & 4.3133 & 4.5206 & 5.0843 & 1.4455 & 4.0610 & 4.5350 & 5.3694 \\
\addlinespace
rls\_feat & 0.5913 & 1.2378 & 1.3165 & 0.6612 & 0.7645 & 0.8488 & 0.7350 & 0.7184 \\
rls\_bank\_med & 0.5753 & 0.8045 & 0.8316 & 0.6288 & 0.6476 & 0.7067 & 0.6778 & 0.6725 \\
rls\_calib & 0.6620 & 0.7818 & 0.7327 & 0.7114 & 0.6435 & 0.6853 & 0.6667 & 0.6651 \\
\addlinespace
rls\_feat\_l1 & 0.6341 & 0.6579 & 0.6553 & 0.6519 & 0.7233 & 0.7390 & 0.7405 & 0.7397 \\
rls\_feat\_l0.9999 & 0.6115 & 0.6504 & 0.6458 & 0.6331 & 0.6929 & 0.7124 & 0.7136 & 0.7130 \\
rls\_feat\_l0.9995 & 0.5924 & 0.8205 & 0.8354 & 0.6332 & 0.6788 & 0.7216 & 0.6885 & 0.6828 \\
rls\_feat\_l0.995 & 0.7289 & 5.7360 & 6.8722 & 1.0836 & 1.3897 & 1.9251 & 1.3712 & 1.2432 \\
rls\_feat\_l0.99 & 0.7926 & 12.5066 & 15.7063 & 1.5910 & 1.5898 & 2.2126 & 1.5986 & 1.4670 \\
\addlinespace
rls\_feat\_noclip & 0.5913 & 1.2378 & 1.3165 & 0.6612 & 0.7645 & 0.8488 & 0.7350 & 0.7184 \\
rls\_bank\_med\_noclip & 0.5740 & 0.8039 & 0.8309 & 0.6283 & 0.6450 & 0.7052 & 0.6769 & 0.6718 \\
rls\_bank\_ftl & 0.6206 & 10.5191 & 13.6277 & 1.3165 & 1.3800 & 1.8399 & 1.2312 & 1.2323 \\
rls\_bank\_dftl & 0.6395 & 10.6712 & 13.8466 & 1.3738 & 1.4815 & 2.0057 & 1.3498 & 1.3101 \\
\addlinespace
elf\_linear & 0.7296 & 0.7745 & 0.6860 & 0.6252 & 0.7296 & 0.7745 & 0.6860 & 0.6252 \\
elf\_mix & 0.6534 & 0.6774 & 0.6553 & 0.6459 & 0.5631 & 0.5929 & 0.5835 & 0.5758 \\
\addlinespace
gd\_feat\_sgd0.0001 & 0.6641 & 0.6994 & 0.7095 & 0.7153 & 0.6513 & 0.6684 & 0.6724 & 0.6746 \\
gd\_feat\_sgd0.001 & 3.7e7 & 4.71e5 & 2.07e5 & 3.01e5 & 1289.4560 & 15.6188 & 9.9744 & 13.6359 \\
gd\_feat\_sgd0.01 & $\infty$ & $\infty$ & $\infty$ & $\infty$ & $\infty$ & $\infty$ & $\infty$ & $\infty$ \\
gd\_feat\_sgd0.1 & $\infty$ & $\infty$ & $\infty$ & $\infty$ & $\infty$ & $\infty$ & $\infty$ & $\infty$ \\
gd\_feat\_sgddec0.001 & 0.9837 & 0.6976 & 0.7095 & 0.6940 & 0.6452 & 0.6610 & 0.6697 & 0.6710 \\
gd\_feat\_sgddec0.01 & $\infty$ & $\infty$ & $\infty$ & $\infty$ & 7.71e24 & 5.92e21 & 6.64e21 & 1.61e23 \\
gd\_feat\_sgddec0.1 & $\infty$ & $\infty$ & $\infty$ & $\infty$ & $\infty$ & $\infty$ & $\infty$ & $\infty$ \\
gd\_feat\_adam0.01 & 0.6109 & 0.9426 & 1.1101 & 1.2301 & 0.6047 & 0.9808 & 1.1905 & 1.2880 \\
gd\_calib\_sgd0.0001 & 0.8039 & 0.8427 & 0.8530 & 0.8567 & 0.6784 & 0.6877 & 0.6918 & 0.6945 \\
gd\_calib\_sgd0.001 & 40.7481 & 1.6432 & 1.3732 & 0.7740 & 0.5948 & 0.6510 & 0.6705 & 0.6650 \\
gd\_calib\_sgd0.01 & $\infty$ & $\infty$ & $\infty$ & $\infty$ & 2.11e12 & 7.64e10 & 8.99e9 & 1.04e8 \\
gd\_calib\_sgd0.1 & $\infty$ & $\infty$ & $\infty$ & $\infty$ & $\infty$ & $\infty$ & $\infty$ & $\infty$ \\
\end{longtable}
}

{\scriptsize
\setlength{\tabcolsep}{3pt}
\begin{longtable}[c]{l rrrr rrrr}
\caption{Full internal grid, Electricity: standardized online MSE at each label-release delay for the DLinear backbone (left) and the PatchTST backbone (right, one seed). $d{=}1$ is the leaky reference. $\infty$: diverged run; --: not run on this stream.}
\label{tab:full_elec_dl}
\label{tab:full_elec_pt}\\
\toprule
 & \multicolumn{4}{c}{DLinear} & \multicolumn{4}{c}{PatchTST (1 seed)} \\
\cmidrule(lr){2-5}\cmidrule(lr){6-9}
method & $d{=}1$ & $d{=}24$ & $d{=}48$ & $d{=}96$ & $d{=}1$ & $d{=}24$ & $d{=}48$ & $d{=}96$ \\
\midrule
\endfirsthead
\toprule
 & \multicolumn{4}{c}{DLinear} & \multicolumn{4}{c}{PatchTST (1 seed)} \\
\cmidrule(lr){2-5}\cmidrule(lr){6-9}
method & $d{=}1$ & $d{=}24$ & $d{=}48$ & $d{=}96$ & $d{=}1$ & $d{=}24$ & $d{=}48$ & $d{=}96$ \\
\midrule
\endhead
\midrule
\multicolumn{9}{r}{\scriptsize continues on next page}\\
\endfoot
\bottomrule
\endlastfoot
naive & 71.8317 & 71.8317 & 71.8317 & 71.8317 & 71.8317 & 71.8317 & 71.8317 & 71.8317 \\
snaive & 6.4556 & 6.4556 & 6.4556 & 6.4556 & 6.4556 & 6.4556 & 6.4556 & 6.4556 \\
frozen & 86.1009 & 86.1009 & 86.1009 & 86.1009 & 119.5988 & 119.5988 & 119.5988 & 119.5988 \\
bias\_ema & 6.2748 & 7.5702 & 7.4239 & 7.6333 & 8.5912 & 9.4826 & 9.5285 & 10.0969 \\
\addlinespace
nlms\_feat\_mu0.02 & 8.8390 & 9.5226 & 9.6619 & 9.7356 & 6.5679 & 7.1724 & 7.2415 & 7.4223 \\
nlms\_feat\_mu0.05 & 7.9004 & 9.5393 & 9.9116 & 10.1309 & 5.7370 & 7.1672 & 7.2657 & 7.7448 \\
nlms\_feat & 7.4015 & 10.2071 & 11.0460 & 11.3678 & 5.2895 & 7.8419 & 7.9533 & 8.8367 \\
nlms\_feat\_mu0.3 & 6.5703 & 11.2002 & 13.3258 & 14.0758 & 4.5325 & 9.2938 & 9.5596 & 11.2124 \\
\addlinespace
rls\_feat & 5.6938 & 5.9561 & 6.0605 & 6.3266 & 5.2936 & 5.5879 & 5.6621 & 6.0852 \\
rls\_bank\_med & 5.7419 & 6.0358 & 6.1558 & 6.4026 & 5.4150 & 5.7072 & 5.7880 & 6.1912 \\
rls\_calib & 6.3054 & 6.5259 & 6.6133 & 6.8929 & 5.8049 & 6.0210 & 6.0808 & 6.4166 \\
\addlinespace
rls\_feat\_l1 & 6.6857 & 6.8692 & 6.9498 & 7.1528 & 5.7059 & 5.9093 & 5.9692 & 6.3481 \\
rls\_feat\_l0.9999 & 6.4567 & 6.6487 & 6.7353 & 6.9413 & 5.6556 & 5.8633 & 5.9204 & 6.2991 \\
rls\_feat\_l0.9995 & 1.03e4 & 6705.9750 & 6970.0410 & 7432.3540 & 28.8400 & 26.2438 & 23.8053 & 20.5027 \\
rls\_feat\_l0.995 & 2.16e4 & 1.05e4 & 1.1e4 & 9037.2320 & 6.23e5 & 6.39e5 & 6.77e5 & 5.89e5 \\
rls\_feat\_l0.99 & 4.8598 & 5.6599 & 5.8081 & 6.2218 & 1.29e4 & 9813.0800 & 9304.1750 & 1.03e4 \\
\addlinespace
rls\_feat\_noclip & 5.22e7 & 5.32e7 & 6.36e7 & 4.78e7 & 1.1e7 & 7.46e6 & 5.25e6 & 5.78e6 \\
rls\_bank\_med\_noclip & 5.7220 & 6.0198 & 6.1363 & 6.3868 & 5.3116 & 5.6099 & 5.6931 & 6.0968 \\
rls\_bank\_ftl & 4.6500 & 5.4607 & 5.6226 & 6.0334 & 4.6916 & 5.5493 & 5.6630 & 6.3127 \\
rls\_bank\_dftl & 4.6457 & 5.4580 & 5.6246 & 6.0288 & 4.6975 & 5.5396 & 5.6464 & 6.2917 \\
\addlinespace
gd\_feat\_sgd0.0001 & $\infty$ & $\infty$ & $\infty$ & $\infty$ & $\infty$ & $\infty$ & $\infty$ & $\infty$ \\
gd\_feat\_sgd0.001 & $\infty$ & $\infty$ & $\infty$ & $\infty$ & $\infty$ & $\infty$ & $\infty$ & $\infty$ \\
gd\_feat\_sgd0.01 & $\infty$ & $\infty$ & $\infty$ & $\infty$ & $\infty$ & $\infty$ & $\infty$ & $\infty$ \\
gd\_feat\_sgd0.1 & $\infty$ & $\infty$ & $\infty$ & $\infty$ & $\infty$ & $\infty$ & $\infty$ & $\infty$ \\
gd\_feat\_sgddec0.001 & $\infty$ & $\infty$ & $\infty$ & $\infty$ & $\infty$ & $\infty$ & $\infty$ & $\infty$ \\
gd\_feat\_sgddec0.01 & $\infty$ & $\infty$ & $\infty$ & $\infty$ & $\infty$ & $\infty$ & $\infty$ & $\infty$ \\
gd\_feat\_sgddec0.1 & $\infty$ & $\infty$ & $\infty$ & $\infty$ & $\infty$ & $\infty$ & $\infty$ & $\infty$ \\
gd\_feat\_adam0.01 & 8.9249 & 10.4222 & 10.7196 & 10.3810 & 7.0412 & 7.9078 & 8.2557 & 8.3098 \\
gd\_calib\_sgd0.0001 & $\infty$ & $\infty$ & $\infty$ & $\infty$ & $\infty$ & $\infty$ & $\infty$ & $\infty$ \\
gd\_calib\_sgd0.001 & $\infty$ & $\infty$ & $\infty$ & $\infty$ & $\infty$ & $\infty$ & $\infty$ & $\infty$ \\
gd\_calib\_sgd0.01 & $\infty$ & $\infty$ & $\infty$ & $\infty$ & $\infty$ & $\infty$ & $\infty$ & $\infty$ \\
gd\_calib\_sgd0.1 & $\infty$ & $\infty$ & $\infty$ & $\infty$ & $\infty$ & $\infty$ & $\infty$ & $\infty$ \\
\end{longtable}
}

{\scriptsize
\setlength{\tabcolsep}{3pt}
\begin{longtable}[c]{l rrrr rrrr}
\caption{Full internal grid, Traffic: standardized online MSE at each label-release delay for the DLinear backbone (left) and the PatchTST backbone (right, one seed). $d{=}1$ is the leaky reference. $\infty$: diverged run; --: not run on this stream.}
\label{tab:full_traffic_dl}
\label{tab:full_traffic_pt}\\
\toprule
 & \multicolumn{4}{c}{DLinear} & \multicolumn{4}{c}{PatchTST (1 seed)} \\
\cmidrule(lr){2-5}\cmidrule(lr){6-9}
method & $d{=}1$ & $d{=}24$ & $d{=}48$ & $d{=}96$ & $d{=}1$ & $d{=}24$ & $d{=}48$ & $d{=}96$ \\
\midrule
\endfirsthead
\toprule
 & \multicolumn{4}{c}{DLinear} & \multicolumn{4}{c}{PatchTST (1 seed)} \\
\cmidrule(lr){2-5}\cmidrule(lr){6-9}
method & $d{=}1$ & $d{=}24$ & $d{=}48$ & $d{=}96$ & $d{=}1$ & $d{=}24$ & $d{=}48$ & $d{=}96$ \\
\midrule
\endhead
\midrule
\multicolumn{9}{r}{\scriptsize continues on next page}\\
\endfoot
\bottomrule
\endlastfoot
naive & 2.5740 & 2.5740 & 2.5740 & 2.5740 & 2.5740 & 2.5740 & 2.5740 & 2.5740 \\
snaive & 0.7831 & 0.7831 & 0.7831 & 0.7831 & 0.7831 & 0.7831 & 0.7831 & 0.7831 \\
frozen & 1.3955 & 1.3955 & 1.3955 & 1.3955 & 0.5410 & 0.5410 & 0.5410 & 0.5410 \\
bias\_ema & 1.2961 & 1.3424 & 1.3257 & 1.3198 & 0.4726 & 0.5213 & 0.5296 & 0.5338 \\
\addlinespace
nlms\_feat\_mu0.02 & 0.6395 & 0.6766 & 0.6681 & 0.6529 & 0.5018 & 0.5201 & 0.5107 & 0.4987 \\
nlms\_feat\_mu0.05 & 0.6463 & 0.7483 & 0.7155 & 0.6944 & 0.5378 & 0.5916 & 0.5597 & 0.5339 \\
nlms\_feat & 0.6660 & 0.9146 & 0.8102 & 0.8032 & 0.5969 & 0.7432 & 0.6597 & 0.6229 \\
nlms\_feat\_mu0.3 & 0.6919 & 1.7576 & 1.2111 & 1.2583 & 0.6988 & 1.4879 & 1.1214 & 1.0500 \\
\addlinespace
rls\_feat & 0.5784 & 0.5922 & 0.5911 & 0.5893 & 0.4607 & 0.4687 & 0.4666 & 0.4666 \\
rls\_bank\_med & 0.5806 & 0.5958 & 0.5951 & 0.5914 & 0.4624 & 0.4707 & 0.4691 & 0.4686 \\
rls\_calib & 0.6030 & 0.6127 & 0.6135 & 0.6141 & 0.4714 & 0.4758 & 0.4759 & 0.4763 \\
\addlinespace
rls\_feat\_l1 & 0.6059 & 0.6134 & 0.6127 & 0.6133 & 0.4801 & 0.4842 & 0.4829 & 0.4833 \\
rls\_feat\_l0.9999 & 0.6005 & 0.6086 & 0.6079 & 0.6083 & 0.4761 & 0.4806 & 0.4792 & 0.4796 \\
rls\_feat\_l0.9995 & 21.3265 & 57.7726 & 39.4705 & 182.4051 & 30.3868 & 37.5891 & 39.3364 & 30.1723 \\
rls\_feat\_l0.995 & 0.5688 & 0.6108 & 0.6061 & 0.5891 & 0.4543 & 0.4790 & 0.4729 & 0.4682 \\
rls\_feat\_l0.99 & 0.5709 & 0.6619 & 0.6540 & 0.6087 & 0.4567 & 0.5182 & 0.5223 & 0.4826 \\
\addlinespace
rls\_feat\_noclip & 2279.2231 & 4584.5747 & 3405.1182 & 4435.9038 & 7.77e6 & 2.36e7 & 3.57e7 & 8.5e7 \\
rls\_bank\_med\_noclip & 0.5801 & 0.5955 & 0.5948 & 0.5911 & 0.4621 & 0.4705 & 0.4688 & 0.4684 \\
rls\_bank\_ftl & 0.5655 & 0.6284 & 0.6193 & 0.6000 & 0.4525 & 0.4910 & 0.4815 & 0.4761 \\
rls\_bank\_dftl & 0.5660 & 0.6412 & 0.6293 & 0.6002 & 0.4512 & 0.5042 & 0.5097 & 0.4761 \\
\addlinespace
gd\_feat\_sgd0.0001 & 0.7023 & 0.7246 & 0.7282 & 0.7266 & 0.4869 & 0.4960 & 0.4967 & 0.4938 \\
gd\_feat\_sgd0.001 & $\infty$ & $\infty$ & $\infty$ & $\infty$ & 0.9206 & 1.0752 & 0.5805 & 0.6243 \\
gd\_feat\_sgd0.01 & $\infty$ & $\infty$ & $\infty$ & $\infty$ & $\infty$ & $\infty$ & $\infty$ & $\infty$ \\
gd\_feat\_sgd0.1 & $\infty$ & $\infty$ & $\infty$ & $\infty$ & $\infty$ & $\infty$ & $\infty$ & $\infty$ \\
gd\_feat\_sgddec0.001 & 0.6052 & 0.6594 & 0.6616 & 0.6658 & 0.4656 & 0.4934 & 0.4937 & 0.4968 \\
gd\_feat\_sgddec0.01 & $\infty$ & $\infty$ & $\infty$ & $\infty$ & $\infty$ & $\infty$ & $\infty$ & $\infty$ \\
gd\_feat\_sgddec0.1 & $\infty$ & $\infty$ & $\infty$ & $\infty$ & $\infty$ & $\infty$ & $\infty$ & $\infty$ \\
gd\_feat\_adam0.01 & 0.6432 & 0.6910 & 0.7065 & 0.6420 & 0.4996 & 0.5060 & 0.5069 & 0.4942 \\
gd\_calib\_sgd0.0001 & 0.7333 & 0.7463 & 0.7521 & 0.7500 & 0.4995 & 0.5004 & 0.5007 & 0.5005 \\
gd\_calib\_sgd0.001 & 1.64e4 & 2.02e4 & 1.31e4 & 4.44e4 & 0.4718 & 0.4795 & 0.4814 & 0.4783 \\
gd\_calib\_sgd0.01 & $\infty$ & $\infty$ & $\infty$ & $\infty$ & 70.2172 & 176.2186 & 49.5892 & 2.4399 \\
gd\_calib\_sgd0.1 & $\infty$ & $\infty$ & $\infty$ & $\infty$ & $\infty$ & $\infty$ & $\infty$ & $\infty$ \\
\end{longtable}
}

{\scriptsize
\begin{longtable}[c]{llrrr}
\caption{External evaluation on the published methods' own PatchTST checkpoints, all five datasets (standardized MSE, aligned on forecast origin). Auxiliary rows (\texttt{rls\_feat}, the no-clip bank, the gradient proxies, \texttt{bias\_ema}) and the ELF-style correctors are our adapters replayed on the same frozen exports. ETT rows are means over three checkpoint seeds where all seeds were run (PETSA and DynaTTA: three seeds at $d{=}H$, seed 0 at $d{>}H$; script configurations: seed 0); Weather, Electricity and Traffic use one checkpoint. --: not run.}
\label{tab:ext_full}\\
\toprule
dataset & method & $d{=}H$ & $d{=}2H$ & $d{=}4H$ \\
\midrule
\endfirsthead
\toprule
dataset & method & $d{=}H$ & $d{=}2H$ & $d{=}4H$ \\
\midrule
\endhead
\midrule
\multicolumn{5}{r}{\scriptsize continues on next page}\\
\endfoot
\bottomrule
\endlastfoot
ETTm1 & frozen (their ckpt) & 1.5238 & 1.5238 & 1.5238 \\
 & TAFAS & 1.4025 & 1.4025 & 1.4030 \\
 & TAFAS (script cfg.) & 1.4656 & 1.4655 & 1.4656 \\
 & COSA & 1.0239 & 0.9011 & 0.9268 \\
 & COSA (script cfg.) & 1.1839 & 1.2116 & 1.2563 \\
 & PETSA & 1.5016 & 1.5761 & 1.5762 \\
 & DynaTTA & 0.6168 & 0.6922 & 0.6741 \\
 & RLS-Bank (ours) & 0.6606 & 0.6648 & 0.6717 \\
 & ELF-linear (run by us) & 0.5149 & 0.5159 & 0.5194 \\
 & ELF-mix (run by us) & 0.5617 & 0.5677 & 0.5803 \\
 & rls\_feat & 0.6666 & 0.6714 & 0.6812 \\
 & rls\_bank\_med\_noclip & 0.6606 & 0.6648 & 0.6717 \\
 & gd\_feat\_sgddec0.001 & 0.6980 & 0.6985 & 0.7019 \\
 & gd\_feat\_sgd0.001 & 0.7334 & 0.7332 & 0.7429 \\
 & bias\_ema & 2.8392 & 2.5796 & 1.2772 \\
 & persistence & 1.8007 & 1.8007 & 1.8007 \\
 & seasonal-naive & 0.9633 & 0.9633 & 0.9633 \\
\addlinespace
ETTh2 & frozen (their ckpt) & 1.3728 & 1.3728 & 1.3728 \\
 & TAFAS & 1.3864 & 1.4050 & 1.3986 \\
 & TAFAS (script cfg.) & 1.2428 & 1.2410 & 1.2388 \\
 & COSA & 1.7343 & 1.8776 & 2.1296 \\
 & COSA (script cfg.) & 1.2357 & 1.2360 & 1.2353 \\
 & PETSA & 1.4410 & 1.3340 & 1.3329 \\
 & DynaTTA & 1.4875 & 1.4423 & 1.4501 \\
 & RLS-Bank (ours) & 1.5076 & 1.5699 & 1.7455 \\
 & ELF-linear (run by us) & 1.1877 & 1.2672 & 1.4442 \\
 & ELF-mix (run by us) & 1.0649 & 1.1509 & 1.3329 \\
 & rls\_feat & 1.5466 & 1.5835 & 1.8154 \\
 & rls\_bank\_med\_noclip & 1.5051 & 1.5664 & 1.7445 \\
 & gd\_feat\_sgddec0.001 & 1.9630 & 2.2956 & 2.4972 \\
 & gd\_feat\_sgd0.001 & 2.0588 & 2.5670 & 2.8717 \\
 & bias\_ema & 1.5027 & 1.6463 & 2.1224 \\
 & persistence & 1.1196 & 1.1196 & 1.1196 \\
 & seasonal-naive & 1.4837 & 1.4837 & 1.4837 \\
\addlinespace
Weather & frozen (their ckpt) & 0.6552 & 0.6552 & 0.6552 \\
 & TAFAS & 0.6557 & 0.6553 & 0.6550 \\
 & COSA & 0.6480 & 0.6652 & 0.6597 \\
 & PETSA & 0.7868 & 0.7870 & 0.7864 \\
 & DynaTTA & 0.6965 & 0.6702 & 0.6756 \\
 & RLS-Bank (ours) & 0.6235 & 0.6091 & 0.6066 \\
 & ELF-linear (run by us) & 0.7746 & 0.6858 & 0.6253 \\
 & ELF-mix (run by us) & 0.6098 & 0.6015 & 0.5934 \\
 & rls\_feat & 0.6951 & 0.6475 & 0.6455 \\
 & rls\_bank\_med\_noclip & 0.6230 & 0.6081 & 0.6059 \\
 & gd\_feat\_sgddec0.001 & 0.6272 & 0.6316 & 0.6332 \\
 & gd\_feat\_sgd0.001 & 4.96e4 & 4.85e4 & 1.07e5 \\
 & bias\_ema & 0.7715 & 0.7881 & 0.7718 \\
 & persistence & 1.0430 & 1.0430 & 1.0430 \\
 & seasonal-naive & 1.7683 & 1.7683 & 1.7683 \\
\addlinespace
Electricity & frozen (their ckpt) & 4.7550 & 4.7550 & 4.7550 \\
 & TAFAS & 4.8071 & 4.7469 & 4.7401 \\
 & COSA & 5.5247 & 5.2068 & 5.2832 \\
 & PETSA & 6.6216 & 6.5833 & 6.5949 \\
 & DynaTTA & 5.4564 & 5.2331 & 5.2348 \\
 & RLS-Bank (ours) & 4.7584 & 4.7616 & 5.2326 \\
 & rls\_feat & 4.7768 & 4.7799 & 5.2601 \\
 & rls\_bank\_med\_noclip & 4.7661 & 4.7647 & 5.2355 \\
 & gd\_feat\_sgddec0.001 & $\infty$ & $\infty$ & $\infty$ \\
 & gd\_feat\_sgd0.001 & $\infty$ & $\infty$ & $\infty$ \\
 & bias\_ema & 5.5800 & 5.2507 & 5.7192 \\
 & persistence & 71.8355 & 71.8355 & 71.8355 \\
 & seasonal-naive & 6.4559 & 6.4559 & 6.4559 \\
\addlinespace
Traffic & frozen (their ckpt) & 0.4134 & 0.4134 & 0.4134 \\
 & TAFAS & 0.4115 & 0.4167 & 0.4162 \\
 & COSA & 0.4121 & 0.4136 & 0.4125 \\
 & PETSA & 0.4603 & 0.4557 & 0.4553 \\
 & RLS-Bank (ours) & 0.4096 & 0.4071 & 0.4060 \\
 & rls\_feat & 0.4112 & 0.4082 & 0.4070 \\
 & rls\_bank\_med\_noclip & 0.4096 & 0.4071 & 0.4060 \\
 & gd\_feat\_sgddec0.001 & 0.4257 & 0.4214 & 0.4199 \\
 & gd\_feat\_sgd0.001 & 4.34e29 & 7.17e28 & 3.26e29 \\
 & bias\_ema & 0.4528 & 0.4605 & 0.4607 \\
 & persistence & 2.5742 & 2.5742 & 2.5742 \\
 & seasonal-naive & 0.7831 & 0.7831 & 0.7831 \\
\end{longtable}
}

\begin{table}[H]
\caption{Adaptation cost: per-update wall time and number of adapted parameters. Wall times for the published methods are measured on the 7-channel ETT streams with the authors' default flags; parameter counts are taken from the runs' own logs. The bank adapts $4{,}032$ closed-form weights ($168$ coordinates $\times F{=}4 \times K{=}6$) on those streams. Wall times are per update, and update frequency differs across methods (the bank updates once per released label; the published methods update per scheduled batch or adaptation event), so per-update ratios and amortized per-step costs differ.
}
\label{tab:cost}
\begin{center}
\small
\setlength{\tabcolsep}{4pt}%
\begin{tabular}{llrp{0.28\linewidth}}
\toprule
Method & Per-update wall time & Adapted params & Notes \\
\midrule
RLS-Bank (ours) & 0.056 ms / step & 4{,}032 & one CPU core; $O(CHF^2)$; 2.9 ms/step mean over all cells \\
ELF-style corrector & $\approx$7 ms / step & 16{,}296 & $O(CHL^2)$, $L{=}96$; run on $C\le 21$ streams \\
TAFAS           & $\approx$5--7 ms & 9{,}914 & authors' code, default flags \\
PETSA           & $\approx$8 ms & 3{,}962 & authors' code, default flags \\
COSA            & $\approx$28 ms $\times$ 20 steps & 5{,}047$^{\dagger}$ & per adaptation event \\
DynaTTA         & ${\sim}$139--158 ms & 11{,}196 & 438 s for one ETTm1 pass \\
\bottomrule
\end{tabular}

{\scriptsize $^{\dagger}$Released-script configuration: 5{,}887 parameters (larger context buffer; Appendix~\ref{app:tuned}).}
\end{center}
\end{table}

\clearpage
\section{Additional Figures}
\label{app:figures}

\begin{figure}[H]
\begin{center}
\includegraphics[width=\linewidth,height=0.8\textheight,keepaspectratio]{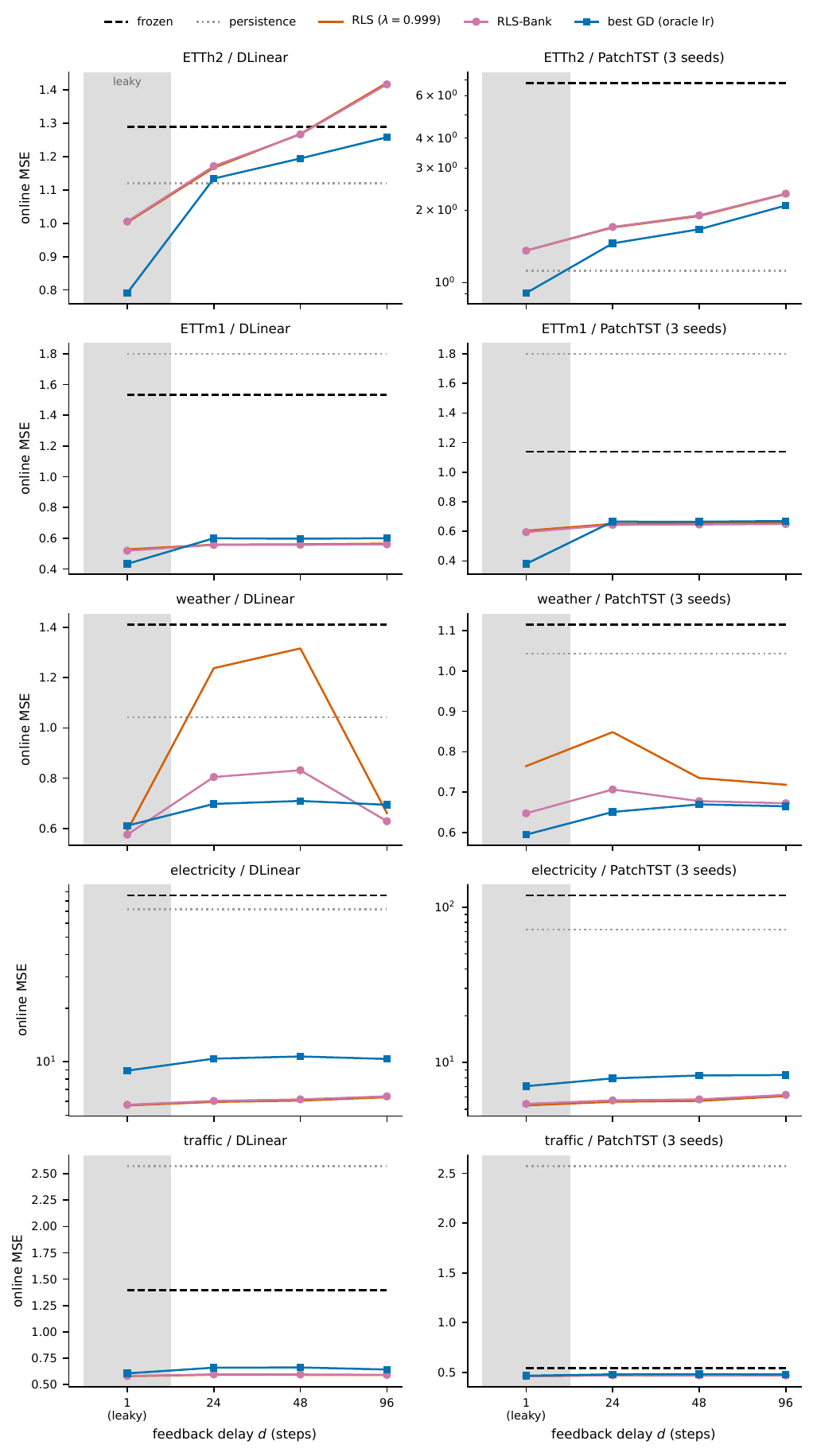}%
\end{center}
\caption{Standardized MSE versus label-release delay $d \in \{1, H, 2H, 4H\}$ ($d{=}1$ is the leaky reference) in the ten internal cells, for the frozen backbone, persistence, a single RLS filter ($\lambda{=}0.999$), the RLS-Bank, and the best gradient adapter (oracle learning rate per delay). The ETTh2/PatchTST panel and both Electricity panels use a logarithmic $y$-axis.}
\label{fig:delaycurves}
\end{figure}

\begin{figure}[H]
\begin{center}
\includegraphics[width=\linewidth,height=0.42\textheight,keepaspectratio]{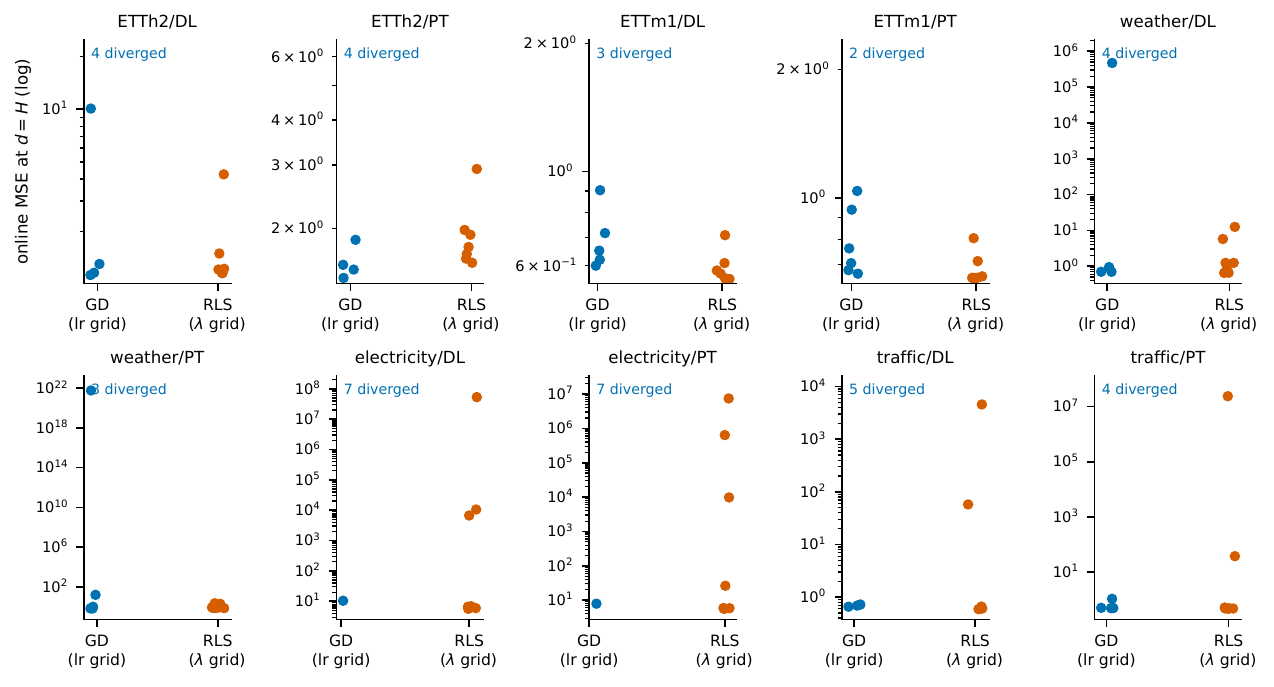}
\end{center}
\caption{Hyperparameter fragility at $d{=}H$: online MSE of every gradient-adapter learning rate (left in each panel) and of every RLS forgetting factor (right) per cell. Panel annotations count divergences among the eight feature-space configurations; the four calibration-space configurations bring the totals to the twelve-configuration counts quoted in Section~\ref{sec:grid} (Appendix~\ref{app:tables}). On the 321-channel Electricity stream every plain-SGD configuration diverges. No RLS setting diverges on the 7- and 21-channel streams; on the two widest streams individual mid-range forgetting factors reach very large but finite error, and the median bank absorbs them (Section~\ref{sec:bank}).}
\label{fig:fragility}
\end{figure}

\begin{figure}[H]
\begin{center}
\includegraphics[width=\linewidth,height=0.42\textheight,keepaspectratio]{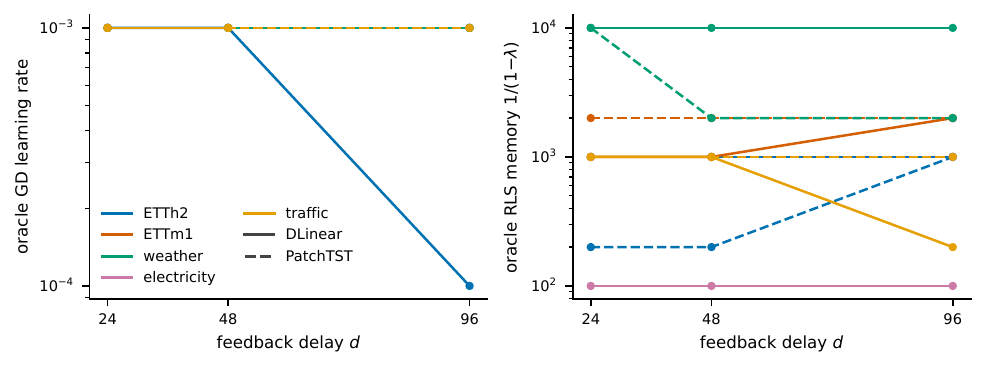}
\end{center}
\caption{Label staleness and plasticity: oracle gradient learning rate (left) and oracle RLS memory $1/(1-\lambda)$ (right) as a function of delay, per cell. The learning-rate grid is decade-spaced, so sub-decade movement does not register in the left panel.}
\label{fig:staleness}
\end{figure}

\begin{figure}[H]
\begin{center}
\includegraphics[width=\linewidth,height=0.42\textheight,keepaspectratio]{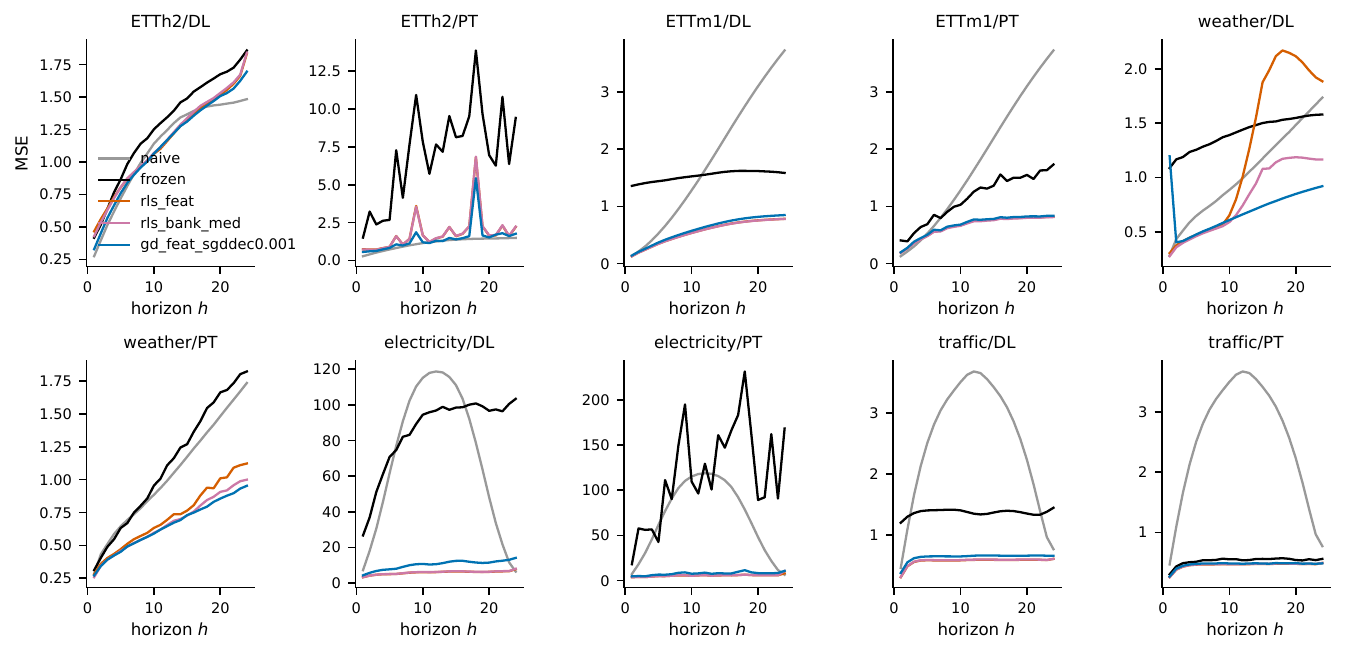}
\end{center}
\caption{Per-horizon MSE at $d{=}H$ across the internal cells.}
\label{fig:perhorizon}
\end{figure}

\begin{figure}[H]
\begin{center}
\includegraphics[width=\linewidth,height=0.6\textheight,keepaspectratio]{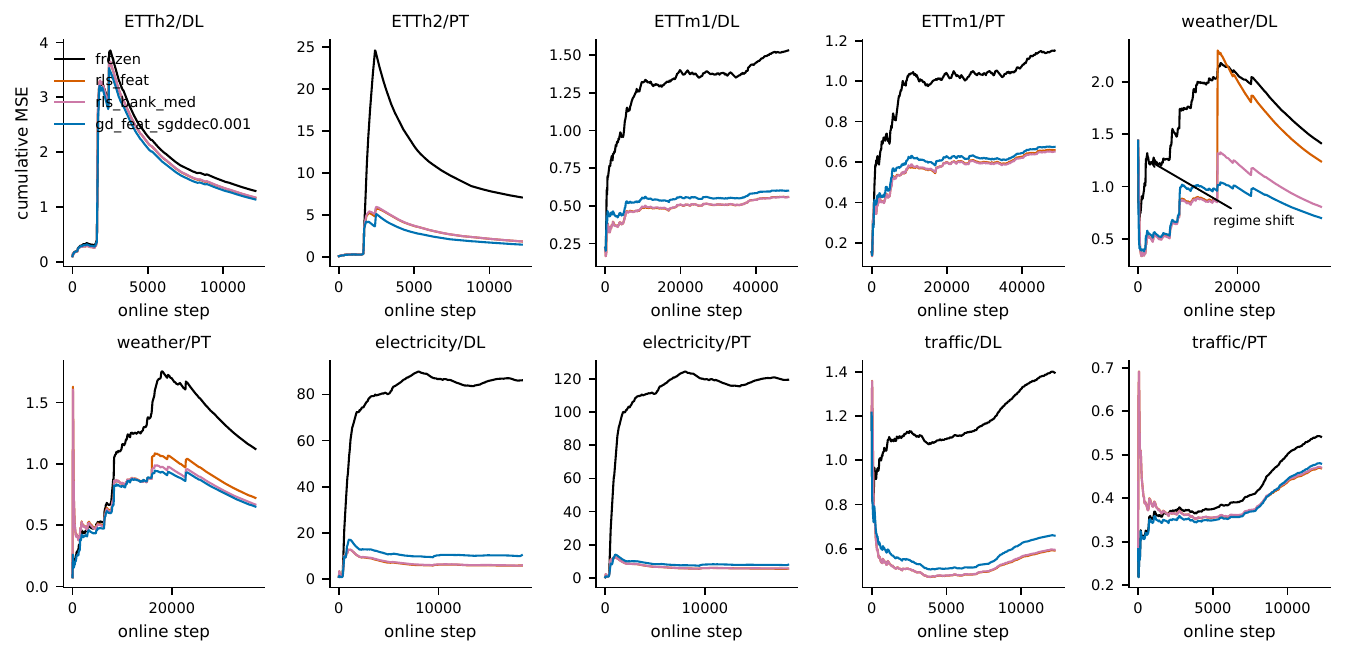}%
\end{center}
\caption{Cumulative-error curves over the online stream, per internal cell.}
\label{fig:cumulative}
\end{figure}

\begin{figure}[H]
\begin{center}
\includegraphics[width=\linewidth]{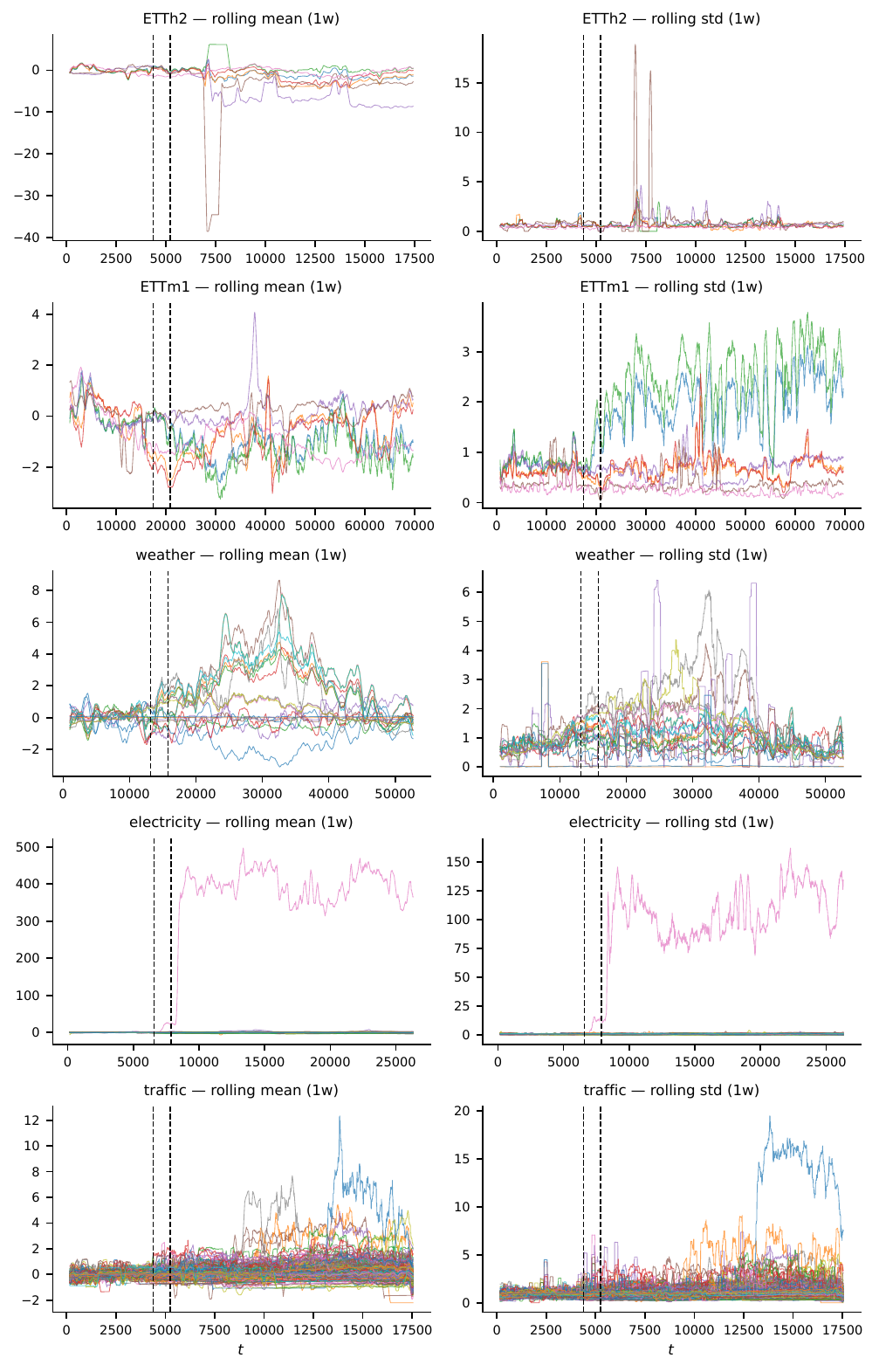}
\end{center}
\caption{Drift overview of the pretraining/validation/online splits: rolling one-week mean and standard deviation per dataset.}
\label{fig:drift}
\end{figure}

\clearpage
\section{RLS-Bank Details}
\label{app:bank}
\paragraph{Pseudocode.}
{\small
\begin{verbatim}
Algorithm 1: RLS-Bank (per channel c, horizon h; K forgetting factors
             + passthrough)
Init: w_k = [1,0,0,0], P_k = rho * I  for k = 1..K
      # P starts at the trace-clip bound
      w_0 = passthrough (never updated)
At stream step i:
  if a label (x_j, y_j) with j = i - d is released:
      for k = 1..K:            # a-priori RLS update, forgetting lambda_k
          g   = P_k x_j / (lambda_k + x_j^T P_k x_j)
          w_k += g (y_j - w_k^T x_j)
          P_k  = (P_k - g x_j^T P_k) / lambda_k
          symmetrize;  clip tr(P_k)/F <= rho
  emit  median_{k in {0..K}} ( w_k^T x_i )
\end{verbatim}
}

\paragraph{Aggregation ablation.}
Table~\ref{tab:agg} compares the median with feedback-driven selection on the five DLinear streams at $d = H$; Section~\ref{sec:bank} discusses the result, including the one stream (Electricity) where selection wins because a single fast forgetting factor is best throughout.

\begin{table}[H]\centering\small
\caption{Aggregation ablation on the five DLinear streams at $d{=}H$ (standardized MSE): the best single forgetting factor chosen in hindsight, follow-the-leader (FTL) selection by matured loss, discounted FTL, and the median bank. Bold marks the best non-oracle aggregation per column.}
\label{tab:agg}
\begin{tabular}{lccccc}
\toprule
Aggregation & ETTh2 & ETTm1 & Weather & Electricity & Traffic \\
\midrule
Best single $\lambda$ (oracle) & 1.1668 & 0.5590 & 0.6504 & 5.6599 & 0.5922 \\
FTL selection & 2.7495 & 0.7084 & 10.5191 & 5.4607 & 0.6284 \\
Discounted FTL ($\gamma{=}0.99$) & 2.7070 & 0.7074 & 10.6712 & \textbf{5.4580} & 0.6412 \\
Median bank (ours) & \textbf{1.1715} & \textbf{0.5572} & \textbf{0.8045} & 6.0358 & \textbf{0.5958} \\
\bottomrule
\end{tabular}
\end{table}

\paragraph{Feature ablation: calibration versus deviation features.}
On ETTm1/PatchTST at the minimum causal delay, the calibration-only variant \texttt{rls\_calib} (a per-coordinate scale and shift on the backbone forecast) scores 1.127, essentially the frozen model's 1.138, while \texttt{rls\_feat} with the two deviation features reaches 0.651 (Table~\ref{tab:full_ettm1_pt}). Pure recalibration leaves signal unused; the deviation features are what let a linear corrector reach the bank's regime.

\paragraph{NLMS structural failure.}
Normalized LMS on the same features fails across its entire step-size grid on ETTh2/DLinear: every setting lands between 8.7 and 103 standardized MSE at the causal and extended delays, against the frozen model's 1.29 (Table~\ref{tab:full_etth2_dl}). The failure is not a tuning problem. NLMS lacks the per-direction gain control that the covariance matrix $P$ provides in RLS, and no scalar step size recovers it. The sweep isolates $P$ as the component doing the work, a class-of-methods answer from adaptive filtering that adding a learning rate does not reach.

\paragraph{Trace clip: measured ablation.}
We ship the unclipped variants (\texttt{rls\_feat\_noclip}, \texttt{rls\_bank\_med\_noclip}) alongside the clipped ones in every cell, which turns the clip from a design argument into a measured result. On the 7- and 21-channel streams the clip is a small tax: disabling it changes the single filter's MSE by at most 0.4\% on most cells and by up to 2.8\% in its favor on ETTh2/PatchTST (1.6497 unclipped vs.\ 1.6975 clipped at $d{=}H$), and it leaves the ETTm1 numbers identical to four decimals. On the wide streams the clip is the difference between working and exploding: at $d{=}H$ the unclipped single filter reaches $5.3\times 10^{7}$ on Electricity/DLinear, $7.5\times 10^{6}$ on Electricity/PatchTST, $4.6\times 10^{3}$ on Traffic/DLinear and $2.4\times 10^{7}$ on Traffic/PatchTST, against 5.96, 5.59, 0.59 and 0.47 clipped. The unclipped \emph{bank} survives everywhere (6.02 vs.\ 6.04 on Electricity/DLinear), because the per-coordinate median discards the exploding expert; this is the same robustness mechanism as in Section~\ref{sec:bank}, observed here from the opposite direction.

\paragraph{ELF-style corrector configuration.}
For completeness: our re-implementation of the released ELF recipe \citep{lee2025adapts} in the harness's own terms fits a per-(channel, horizon) linear forecaster by recursive least squares on the raw lookback window (growing-window OLS at $\lambda{=}1$, the form the ELF paper calls OLS-equivalent) and combines it with the frozen forecast by an online inverse-error weighter (EW-MSE memory $\beta{=}0.99$). Both the fit and the weighter update only on released labels, so the corrector lives on the same delay axis as everything else. It is restricted to streams with $C\le 21$ because its per-step cost is $O(CHL^2)$.

\section{Bootstrap Details}
\label{app:bootstrap}
All intervals in the paper come from a circular moving-block bootstrap on paired per-step loss differentials. For two methods scored on the same stream, we form the per-origin difference of losses, resample it in contiguous blocks of length twice the seasonal period (wrapping circularly), and report the 2.5th and 97.5th percentiles of the resampled mean over $B{=}500$ resamples. The block length is chosen to cover the dominant daily dependence; the main caveat is that correlation at lags beyond the block length is not captured, which would make the intervals anti-conservative. Internal PatchTST cells with several seeds pool the paired differentials of the seeds, which is why a pooled differential can differ slightly from the difference of the seed means in Table~\ref{tab:maingrid}; external comparisons are computed per checkpoint seed and listed with their seed. On the two widest streams the internal comparisons are restricted to the essential pairs at $d{=}H$ and $d{=}4H$ (compute budget). Tables~\ref{tab:ci_internal} and~\ref{tab:ci_external_ettm1}--\ref{tab:ci_external_wet} list the cached intervals behind every significance claim in the main text.

{\tiny
\setlength{\tabcolsep}{4pt}
\begin{longtable}[c]{llllrrr}
\caption{Internal cells: paired per-step moving-block bootstrap (block $=2\times$ period, $B{=}500$), pooled over seeds where a cell has several. Negative mean $=$ first method better; $^{*}$ $=$ 95\% CI excludes zero.}
\label{tab:ci_internal}\\
\toprule
dataset & backbone & delay & pair & mean\_diff & ci\_lo & ci\_hi \\
\midrule
\endfirsthead
\toprule
dataset & backbone & delay & pair & mean\_diff & ci\_lo & ci\_hi \\
\midrule
\endhead
\midrule
\multicolumn{7}{r}{\scriptsize continues on next page}\\
\endfoot
\bottomrule
\endlastfoot
ETTh2 & dlinear & 24 & elf\_linear - naive & 0.0681 & -0.1415 & 0.3442 \\
ETTh2 & dlinear & 24 & elf\_linear - rls\_bank\_med & 0.0162 & -0.1343 & 0.2041 \\
ETTh2 & dlinear & 24 & elf\_mix - gd\_feat\_sgddec0.001$^{*}$ & -0.0906 & -0.1302 & -0.0542 \\
ETTh2 & dlinear & 24 & elf\_mix - naive & -0.0759 & -0.2114 & 0.1292 \\
ETTh2 & dlinear & 24 & elf\_mix - rls\_bank\_med$^{*}$ & -0.1278 & -0.2249 & -0.0562 \\
ETTh2 & dlinear & 24 & rls\_bank\_med - frozen$^{*}$ & -0.1179 & -0.1944 & -0.0369 \\
ETTh2 & dlinear & 24 & rls\_bank\_med - gd\_feat\_sgddec0.001 & 0.0372 & -0.0274 & 0.1131 \\
ETTh2 & dlinear & 24 & rls\_bank\_med - naive & 0.0519 & -0.1321 & 0.3311 \\
ETTh2 & dlinear & 24 & rls\_feat - gd\_feat\_sgddec0.001 & 0.0325 & -0.0330 & 0.1092 \\
ETTh2 & dlinear & 96 & elf\_linear - naive & 0.3246 & -0.1004 & 0.8572 \\
ETTh2 & dlinear & 96 & elf\_linear - rls\_bank\_med & 0.0273 & -0.3383 & 0.4899 \\
ETTh2 & dlinear & 96 & elf\_mix - gd\_feat\_sgd0.0001$^{*}$ & -0.1100 & -0.2026 & -0.0062 \\
ETTh2 & dlinear & 96 & elf\_mix - naive & 0.0286 & -0.1842 & 0.3216 \\
ETTh2 & dlinear & 96 & elf\_mix - rls\_bank\_med$^{*}$ & -0.2686 & -0.5620 & -0.0412 \\
ETTh2 & dlinear & 96 & rls\_bank\_med - frozen & 0.1274 & -0.1063 & 0.4416 \\
ETTh2 & dlinear & 96 & rls\_bank\_med - gd\_feat\_sgd0.0001 & 0.1586 & -0.0450 & 0.4887 \\
ETTh2 & dlinear & 96 & rls\_bank\_med - naive & 0.2973 & -0.0798 & 0.8873 \\
ETTh2 & dlinear & 96 & rls\_feat - gd\_feat\_sgd0.0001 & 0.1631 & -0.0469 & 0.4424 \\
ETTh2 & patchtst & 24 & elf\_linear - naive & 0.0681 & -0.1527 & 0.3918 \\
ETTh2 & patchtst & 24 & elf\_linear - rls\_bank\_med$^{*}$ & -0.6374 & -1.0480 & -0.3309 \\
ETTh2 & patchtst & 24 & elf\_mix - gd\_feat\_sgd0.001$^{*}$ & -0.3314 & -0.6877 & -0.0733 \\
ETTh2 & patchtst & 24 & elf\_mix - naive & -0.0091 & -0.1928 & 0.2513 \\
ETTh2 & patchtst & 24 & elf\_mix - rls\_bank\_med$^{*}$ & -0.7146 & -1.0681 & -0.4438 \\
ETTh2 & patchtst & 24 & rls\_bank\_med - frozen$^{*}$ & -5.2368 & -7.1403 & -3.4713 \\
ETTh2 & patchtst & 24 & rls\_bank\_med - gd\_feat\_sgd0.001$^{*}$ & 0.3832 & 0.1146 & 0.6132 \\
ETTh2 & patchtst & 24 & rls\_bank\_med - naive$^{*}$ & 0.7055 & 0.2784 & 1.2331 \\
ETTh2 & patchtst & 24 & rls\_feat - gd\_feat\_sgd0.001$^{*}$ & 0.3755 & 0.1422 & 0.5781 \\
ETTh2 & patchtst & 96 & elf\_linear - naive & 0.3246 & -0.0720 & 0.8268 \\
ETTh2 & patchtst & 96 & elf\_linear - rls\_bank\_med$^{*}$ & -1.0606 & -1.9194 & -0.4124 \\
ETTh2 & patchtst & 96 & elf\_mix - gd\_feat\_sgd0.001$^{*}$ & -0.8028 & -1.4862 & -0.3048 \\
ETTh2 & patchtst & 96 & elf\_mix - naive & 0.2479 & -0.1642 & 0.7945 \\
ETTh2 & patchtst & 96 & elf\_mix - rls\_bank\_med$^{*}$ & -1.1373 & -1.8034 & -0.5894 \\
ETTh2 & patchtst & 96 & rls\_bank\_med - frozen$^{*}$ & -4.5571 & -6.3396 & -2.9971 \\
ETTh2 & patchtst & 96 & rls\_bank\_med - gd\_feat\_sgd0.001 & 0.3346 & -0.1365 & 0.6896 \\
ETTh2 & patchtst & 96 & rls\_bank\_med - naive$^{*}$ & 1.3852 & 0.4350 & 2.7215 \\
ETTh2 & patchtst & 96 & rls\_feat - gd\_feat\_sgd0.001 & 0.3332 & -0.1491 & 0.6737 \\
ETTm1 & dlinear & 24 & elf\_linear - naive$^{*}$ & -1.2858 & -1.3971 & -1.1685 \\
ETTm1 & dlinear & 24 & elf\_linear - rls\_bank\_med$^{*}$ & -0.0423 & -0.0503 & -0.0347 \\
ETTm1 & dlinear & 24 & elf\_mix - gd\_feat\_sgddec0.001$^{*}$ & -0.0286 & -0.0428 & -0.0142 \\
ETTm1 & dlinear & 24 & elf\_mix - naive$^{*}$ & -1.2287 & -1.3467 & -1.1094 \\
ETTm1 & dlinear & 24 & elf\_mix - rls\_bank\_med$^{*}$ & 0.0148 & 0.0055 & 0.0242 \\
ETTm1 & dlinear & 24 & rls\_bank\_med - frozen$^{*}$ & -0.9772 & -1.0360 & -0.9181 \\
ETTm1 & dlinear & 24 & rls\_bank\_med - gd\_feat\_sgddec0.001$^{*}$ & -0.0434 & -0.0518 & -0.0357 \\
ETTm1 & dlinear & 24 & rls\_bank\_med - naive$^{*}$ & -1.2435 & -1.3648 & -1.1321 \\
ETTm1 & dlinear & 24 & rls\_feat - gd\_feat\_sgddec0.001$^{*}$ & -0.0416 & -0.0499 & -0.0335 \\
ETTm1 & dlinear & 96 & elf\_linear - naive$^{*}$ & -1.2813 & -1.3972 & -1.1533 \\
ETTm1 & dlinear & 96 & elf\_linear - rls\_bank\_med$^{*}$ & -0.0418 & -0.0523 & -0.0310 \\
ETTm1 & dlinear & 96 & elf\_mix - gd\_feat\_sgddec0.001$^{*}$ & -0.0258 & -0.0424 & -0.0111 \\
ETTm1 & dlinear & 96 & elf\_mix - naive$^{*}$ & -1.2255 & -1.3357 & -1.1000 \\
ETTm1 & dlinear & 96 & elf\_mix - rls\_bank\_med$^{*}$ & 0.0139 & 0.0025 & 0.0252 \\
ETTm1 & dlinear & 96 & rls\_bank\_med - frozen$^{*}$ & -0.9732 & -1.0350 & -0.9106 \\
ETTm1 & dlinear & 96 & rls\_bank\_med - gd\_feat\_sgddec0.001$^{*}$ & -0.0398 & -0.0492 & -0.0315 \\
ETTm1 & dlinear & 96 & rls\_bank\_med - naive$^{*}$ & -1.2395 & -1.3747 & -1.1301 \\
ETTm1 & dlinear & 96 & rls\_feat - gd\_feat\_sgddec0.001$^{*}$ & -0.0350 & -0.0431 & -0.0271 \\
ETTm1 & patchtst & 24 & elf\_linear - naive$^{*}$ & -1.2858 & -1.4127 & -1.1597 \\
ETTm1 & patchtst & 24 & elf\_linear - rls\_bank\_med$^{*}$ & -0.1363 & -0.1547 & -0.1197 \\
ETTm1 & patchtst & 24 & elf\_mix - gd\_feat\_sgddec0.001$^{*}$ & -0.1191 & -0.1351 & -0.1053 \\
ETTm1 & patchtst & 24 & elf\_mix - naive$^{*}$ & -1.2448 & -1.3637 & -1.1319 \\
ETTm1 & patchtst & 24 & elf\_mix - rls\_bank\_med$^{*}$ & -0.0953 & -0.1080 & -0.0825 \\
ETTm1 & patchtst & 24 & rls\_bank\_med - frozen$^{*}$ & -0.5012 & -0.5765 & -0.4290 \\
ETTm1 & patchtst & 24 & rls\_bank\_med - gd\_feat\_sgddec0.001$^{*}$ & -0.0238 & -0.0296 & -0.0178 \\
ETTm1 & patchtst & 24 & rls\_bank\_med - naive$^{*}$ & -1.1495 & -1.2735 & -1.0239 \\
ETTm1 & patchtst & 24 & rls\_feat - gd\_feat\_sgddec0.001$^{*}$ & -0.0163 & -0.0247 & -0.0065 \\
ETTm1 & patchtst & 96 & elf\_linear - naive$^{*}$ & -1.2813 & -1.4046 & -1.1585 \\
ETTm1 & patchtst & 96 & elf\_linear - rls\_bank\_med$^{*}$ & -0.1385 & -0.1608 & -0.1171 \\
ETTm1 & patchtst & 96 & elf\_mix - gd\_feat\_sgddec0.001$^{*}$ & -0.1105 & -0.1265 & -0.0936 \\
ETTm1 & patchtst & 96 & elf\_mix - naive$^{*}$ & -1.2319 & -1.3575 & -1.1254 \\
ETTm1 & patchtst & 96 & elf\_mix - rls\_bank\_med$^{*}$ & -0.0891 & -0.1069 & -0.0749 \\
ETTm1 & patchtst & 96 & rls\_bank\_med - frozen$^{*}$ & -0.4945 & -0.5675 & -0.4292 \\
ETTm1 & patchtst & 96 & rls\_bank\_med - gd\_feat\_sgddec0.001$^{*}$ & -0.0214 & -0.0274 & -0.0143 \\
ETTm1 & patchtst & 96 & rls\_bank\_med - naive$^{*}$ & -1.1428 & -1.2577 & -1.0359 \\
ETTm1 & patchtst & 96 & rls\_feat - gd\_feat\_sgddec0.001$^{*}$ & -0.0114 & -0.0198 & -0.0029 \\
electricity & dlinear & 24 & rls\_bank\_med - frozen$^{*}$ & -80.0645 & -82.2347 & -77.6082 \\
electricity & dlinear & 24 & rls\_bank\_med - gd\_feat\_adam0.01$^{*}$ & -4.3863 & -4.8753 & -3.8806 \\
electricity & dlinear & 24 & rls\_bank\_med - naive$^{*}$ & -65.7958 & -68.8694 & -62.8963 \\
electricity & dlinear & 24 & rls\_feat - gd\_feat\_adam0.01$^{*}$ & -4.4660 & -5.0449 & -3.9084 \\
electricity & dlinear & 96 & rls\_bank\_med - frozen$^{*}$ & -79.6977 & -82.0649 & -77.4113 \\
electricity & dlinear & 96 & rls\_bank\_med - gd\_feat\_adam0.01$^{*}$ & -3.9784 & -4.5440 & -3.4567 \\
electricity & dlinear & 96 & rls\_bank\_med - naive$^{*}$ & -65.4290 & -68.6888 & -62.4810 \\
electricity & dlinear & 96 & rls\_feat - gd\_feat\_adam0.01$^{*}$ & -4.0544 & -4.6444 & -3.5210 \\
electricity & patchtst & 24 & rls\_bank\_med - frozen$^{*}$ & -113.8916 & -117.0955 & -111.1290 \\
electricity & patchtst & 24 & rls\_bank\_med - gd\_feat\_adam0.01$^{*}$ & -2.2006 & -2.6649 & -1.8204 \\
electricity & patchtst & 24 & rls\_bank\_med - naive$^{*}$ & -66.1245 & -69.0728 & -63.1645 \\
electricity & patchtst & 24 & rls\_feat - gd\_feat\_adam0.01$^{*}$ & -2.3199 & -2.7505 & -1.9507 \\
electricity & patchtst & 96 & rls\_bank\_med - frozen$^{*}$ & -113.4076 & -116.7331 & -110.4390 \\
electricity & patchtst & 96 & rls\_bank\_med - gd\_feat\_adam0.01$^{*}$ & -2.1187 & -2.5989 & -1.6044 \\
electricity & patchtst & 96 & rls\_bank\_med - naive$^{*}$ & -65.6404 & -68.9198 & -62.7178 \\
electricity & patchtst & 96 & rls\_feat - gd\_feat\_adam0.01$^{*}$ & -2.2247 & -2.7476 & -1.7093 \\
traffic & dlinear & 24 & rls\_bank\_med - frozen$^{*}$ & -0.7997 & -0.8401 & -0.7640 \\
traffic & dlinear & 24 & rls\_bank\_med - gd\_feat\_sgddec0.001$^{*}$ & -0.0636 & -0.0740 & -0.0547 \\
traffic & dlinear & 24 & rls\_bank\_med - naive$^{*}$ & -1.9783 & -2.0780 & -1.8775 \\
traffic & dlinear & 24 & rls\_feat - gd\_feat\_sgddec0.001$^{*}$ & -0.0672 & -0.0784 & -0.0579 \\
traffic & dlinear & 96 & rls\_bank\_med - frozen$^{*}$ & -0.8041 & -0.8435 & -0.7686 \\
traffic & dlinear & 96 & rls\_bank\_med - gd\_feat\_adam0.01$^{*}$ & -0.0506 & -0.0612 & -0.0407 \\
traffic & dlinear & 96 & rls\_bank\_med - naive$^{*}$ & -1.9826 & -2.0798 & -1.8709 \\
traffic & dlinear & 96 & rls\_feat - gd\_feat\_adam0.01$^{*}$ & -0.0527 & -0.0633 & -0.0423 \\
traffic & patchtst & 24 & rls\_bank\_med - frozen$^{*}$ & -0.0703 & -0.0805 & -0.0592 \\
traffic & patchtst & 24 & rls\_bank\_med - gd\_calib\_sgd0.001$^{*}$ & -0.0088 & -0.0132 & -0.0031 \\
traffic & patchtst & 24 & rls\_bank\_med - naive$^{*}$ & -2.1033 & -2.2032 & -2.0052 \\
traffic & patchtst & 24 & rls\_feat - gd\_calib\_sgd0.001$^{*}$ & -0.0108 & -0.0152 & -0.0051 \\
traffic & patchtst & 96 & rls\_bank\_med - frozen$^{*}$ & -0.0724 & -0.0830 & -0.0612 \\
traffic & patchtst & 96 & rls\_bank\_med - gd\_calib\_sgd0.001$^{*}$ & -0.0098 & -0.0133 & -0.0056 \\
traffic & patchtst & 96 & rls\_bank\_med - naive$^{*}$ & -2.1054 & -2.2054 & -2.0156 \\
traffic & patchtst & 96 & rls\_feat - gd\_calib\_sgd0.001$^{*}$ & -0.0118 & -0.0155 & -0.0072 \\
weather & dlinear & 24 & elf\_linear - naive$^{*}$ & -0.2685 & -0.4474 & -0.0315 \\
weather & dlinear & 24 & elf\_linear - rls\_bank\_med & -0.0301 & -0.2993 & 0.1424 \\
weather & dlinear & 24 & elf\_mix - gd\_feat\_sgddec0.001 & -0.0202 & -0.0739 & 0.0265 \\
weather & dlinear & 24 & elf\_mix - naive$^{*}$ & -0.3656 & -0.4748 & -0.2660 \\
weather & dlinear & 24 & elf\_mix - rls\_bank\_med & -0.1271 & -0.5947 & 0.0509 \\
weather & dlinear & 24 & rls\_bank\_med - frozen$^{*}$ & -0.6072 & -0.8513 & -0.2674 \\
weather & dlinear & 24 & rls\_bank\_med - gd\_feat\_sgddec0.001 & 0.1069 & -0.1168 & 0.4915 \\
weather & dlinear & 24 & rls\_bank\_med - naive & -0.2385 & -0.5185 & 0.1577 \\
weather & dlinear & 24 & rls\_feat - gd\_feat\_sgddec0.001 & 0.5402 & -0.0964 & 1.9894 \\
weather & dlinear & 96 & elf\_linear - naive$^{*}$ & -0.4178 & -0.5453 & -0.3233 \\
weather & dlinear & 96 & elf\_linear - rls\_bank\_med & -0.0036 & -0.0225 & 0.0131 \\
weather & dlinear & 96 & elf\_mix - gd\_feat\_sgddec0.001$^{*}$ & -0.0481 & -0.1170 & -0.0066 \\
weather & dlinear & 96 & elf\_mix - naive$^{*}$ & -0.3971 & -0.5008 & -0.3083 \\
weather & dlinear & 96 & elf\_mix - rls\_bank\_med & 0.0172 & -0.0100 & 0.0412 \\
weather & dlinear & 96 & rls\_bank\_med - frozen$^{*}$ & -0.7830 & -0.9165 & -0.6507 \\
weather & dlinear & 96 & rls\_bank\_med - gd\_feat\_sgddec0.001$^{*}$ & -0.0653 & -0.1215 & -0.0304 \\
weather & dlinear & 96 & rls\_bank\_med - naive$^{*}$ & -0.4142 & -0.5159 & -0.3188 \\
weather & dlinear & 96 & rls\_feat - gd\_feat\_sgddec0.001 & -0.0328 & -0.0704 & 0.0127 \\
weather & patchtst & 24 & elf\_linear - naive$^{*}$ & -0.2685 & -0.4515 & -0.0024 \\
weather & patchtst & 24 & elf\_linear - rls\_bank\_med & 0.1061 & -0.0180 & 0.2717 \\
weather & patchtst & 24 & elf\_mix - gd\_calib\_sgd0.001$^{*}$ & -0.0585 & -0.0860 & -0.0161 \\
weather & patchtst & 24 & elf\_mix - naive$^{*}$ & -0.4503 & -0.5723 & -0.3499 \\
weather & patchtst & 24 & elf\_mix - rls\_bank\_med$^{*}$ & -0.0757 & -0.1059 & -0.0511 \\
weather & patchtst & 24 & rls\_bank\_med - frozen$^{*}$ & -0.4520 & -0.5801 & -0.3274 \\
weather & patchtst & 24 & rls\_bank\_med - gd\_calib\_sgd0.001 & 0.0172 & -0.0153 & 0.0672 \\
weather & patchtst & 24 & rls\_bank\_med - naive$^{*}$ & -0.3746 & -0.4782 & -0.2691 \\
weather & patchtst & 24 & rls\_feat - gd\_calib\_sgd0.001 & 0.0709 & -0.0070 & 0.2098 \\
weather & patchtst & 96 & elf\_linear - naive$^{*}$ & -0.4178 & -0.5311 & -0.3270 \\
weather & patchtst & 96 & elf\_linear - rls\_bank\_med$^{*}$ & -0.0422 & -0.0816 & -0.0075 \\
weather & patchtst & 96 & elf\_mix - gd\_calib\_sgd0.001$^{*}$ & -0.0910 & -0.1145 & -0.0717 \\
weather & patchtst & 96 & elf\_mix - naive$^{*}$ & -0.4698 & -0.5871 & -0.3785 \\
weather & patchtst & 96 & elf\_mix - rls\_bank\_med$^{*}$ & -0.0941 & -0.1369 & -0.0600 \\
weather & patchtst & 96 & rls\_bank\_med - frozen$^{*}$ & -0.4530 & -0.5943 & -0.3442 \\
weather & patchtst & 96 & rls\_bank\_med - gd\_calib\_sgd0.001 & 0.0032 & -0.0219 & 0.0447 \\
weather & patchtst & 96 & rls\_bank\_med - naive$^{*}$ & -0.3757 & -0.4998 & -0.2854 \\
weather & patchtst & 96 & rls\_feat - gd\_calib\_sgd0.001 & 0.0355 & -0.0157 & 0.1177 \\
\end{longtable}
}

{\tiny
\setlength{\tabcolsep}{4pt}
\begin{longtable}[c]{llllrrr}
\caption{External comparisons on the authors' own ETTm1 checkpoints (origin-aligned streams; same bootstrap), per checkpoint seed, including the released-script (\texttt{\_tuned}) configurations. Negative mean $=$ first method better; $^{*}$ $=$ 95\% CI excludes zero.}
\label{tab:ci_external_ettm1}\\
\toprule
dataset & seed & delay & pair & mean\_diff & ci\_lo & ci\_hi \\
\midrule
\endfirsthead
\toprule
dataset & seed & delay & pair & mean\_diff & ci\_lo & ci\_hi \\
\midrule
\endhead
\midrule
\multicolumn{7}{r}{\scriptsize continues on next page}\\
\endfoot
\bottomrule
\endlastfoot
ETTm1 & 0 & 24 & cosa - frozen$^{*}$ & -0.6349 & -0.7144 & -0.5521 \\
ETTm1 & 0 & 24 & cosa\_tuned - cosa$^{*}$ & 0.1511 & 0.1152 & 0.1876 \\
ETTm1 & 0 & 24 & cosa\_tuned - frozen$^{*}$ & -0.4838 & -0.5441 & -0.4282 \\
ETTm1 & 0 & 24 & dynatta - frozen$^{*}$ & -1.0485 & -1.1691 & -0.9337 \\
ETTm1 & 0 & 24 & elf\_linear - rls\_bank\_med$^{*}$ & -0.1559 & -0.1760 & -0.1363 \\
ETTm1 & 0 & 24 & elf\_mix - cosa$^{*}$ & -0.4698 & -0.5285 & -0.4169 \\
ETTm1 & 0 & 24 & elf\_mix - cosa\_tuned$^{*}$ & -0.6209 & -0.6973 & -0.5560 \\
ETTm1 & 0 & 24 & elf\_mix - dynatta$^{*}$ & -0.0562 & -0.0719 & -0.0395 \\
ETTm1 & 0 & 24 & elf\_mix - frozen$^{*}$ & -1.1047 & -1.2294 & -0.9892 \\
ETTm1 & 0 & 24 & elf\_mix - naive$^{*}$ & -1.2378 & -1.3476 & -1.1150 \\
ETTm1 & 0 & 24 & elf\_mix - petsa$^{*}$ & -1.0345 & -1.1444 & -0.9324 \\
ETTm1 & 0 & 24 & elf\_mix - rls\_bank\_med$^{*}$ & -0.1079 & -0.1257 & -0.0896 \\
ETTm1 & 0 & 24 & elf\_mix - tafas$^{*}$ & -0.9019 & -1.0081 & -0.8093 \\
ETTm1 & 0 & 24 & elf\_mix - tafas\_tuned$^{*}$ & -0.9027 & -1.0020 & -0.8097 \\
ETTm1 & 0 & 24 & petsa - frozen$^{*}$ & -0.0702 & -0.0873 & -0.0534 \\
ETTm1 & 0 & 24 & rls\_bank\_med - cosa$^{*}$ & -0.3619 & -0.4196 & -0.3054 \\
ETTm1 & 0 & 24 & rls\_bank\_med - cosa\_tuned$^{*}$ & -0.5131 & -0.5861 & -0.4469 \\
ETTm1 & 0 & 24 & rls\_bank\_med - dynatta$^{*}$ & 0.0520 & 0.0283 & 0.0760 \\
ETTm1 & 0 & 24 & rls\_bank\_med - frozen$^{*}$ & -0.9968 & -1.1151 & -0.8767 \\
ETTm1 & 0 & 24 & rls\_bank\_med - naive$^{*}$ & -1.1299 & -1.2498 & -1.0129 \\
ETTm1 & 0 & 24 & rls\_bank\_med - petsa$^{*}$ & -0.9266 & -1.0428 & -0.8169 \\
ETTm1 & 0 & 24 & rls\_bank\_med - tafas$^{*}$ & -0.7941 & -0.8850 & -0.7084 \\
ETTm1 & 0 & 24 & rls\_bank\_med - tafas\_tuned$^{*}$ & -0.7948 & -0.8908 & -0.7042 \\
ETTm1 & 0 & 24 & tafas - frozen$^{*}$ & -0.2028 & -0.2387 & -0.1691 \\
ETTm1 & 0 & 24 & tafas\_tuned - frozen$^{*}$ & -0.2020 & -0.2322 & -0.1740 \\
ETTm1 & 0 & 24 & tafas\_tuned - tafas & 0.0007 & -0.0133 & 0.0132 \\
ETTm1 & 0 & 48 & cosa - frozen$^{*}$ & -0.7521 & -0.8433 & -0.6650 \\
ETTm1 & 0 & 48 & cosa\_tuned - cosa$^{*}$ & 0.2961 & 0.2553 & 0.3376 \\
ETTm1 & 0 & 48 & cosa\_tuned - frozen$^{*}$ & -0.4560 & -0.5095 & -0.4032 \\
ETTm1 & 0 & 48 & dynatta - frozen$^{*}$ & -0.9754 & -1.0779 & -0.8719 \\
ETTm1 & 0 & 48 & elf\_linear - rls\_bank\_med$^{*}$ & -0.1597 & -0.1828 & -0.1366 \\
ETTm1 & 0 & 48 & elf\_mix - cosa$^{*}$ & -0.3465 & -0.3886 & -0.3041 \\
ETTm1 & 0 & 48 & elf\_mix - cosa\_tuned$^{*}$ & -0.6426 & -0.7117 & -0.5824 \\
ETTm1 & 0 & 48 & elf\_mix - dynatta$^{*}$ & -0.1232 & -0.1466 & -0.1023 \\
ETTm1 & 0 & 48 & elf\_mix - frozen$^{*}$ & -1.0986 & -1.2188 & -0.9942 \\
ETTm1 & 0 & 48 & elf\_mix - naive$^{*}$ & -1.2316 & -1.3473 & -1.1222 \\
ETTm1 & 0 & 48 & elf\_mix - petsa$^{*}$ & -1.0071 & -1.1050 & -0.9059 \\
ETTm1 & 0 & 48 & elf\_mix - rls\_bank\_med$^{*}$ & -0.1065 & -0.1291 & -0.0876 \\
ETTm1 & 0 & 48 & elf\_mix - tafas$^{*}$ & -0.8958 & -0.9794 & -0.8046 \\
ETTm1 & 0 & 48 & elf\_mix - tafas\_tuned$^{*}$ & -0.8964 & -0.9931 & -0.8110 \\
ETTm1 & 0 & 48 & petsa - frozen$^{*}$ & -0.0915 & -0.1129 & -0.0728 \\
ETTm1 & 0 & 48 & rls\_bank\_med - cosa$^{*}$ & -0.2400 & -0.2865 & -0.1998 \\
ETTm1 & 0 & 48 & rls\_bank\_med - cosa\_tuned$^{*}$ & -0.5361 & -0.6115 & -0.4705 \\
ETTm1 & 0 & 48 & rls\_bank\_med - dynatta & -0.0167 & -0.0476 & 0.0153 \\
ETTm1 & 0 & 48 & rls\_bank\_med - frozen$^{*}$ & -0.9921 & -1.1132 & -0.8663 \\
ETTm1 & 0 & 48 & rls\_bank\_med - naive$^{*}$ & -1.1252 & -1.2487 & -1.0214 \\
ETTm1 & 0 & 48 & rls\_bank\_med - petsa$^{*}$ & -0.9006 & -1.0116 & -0.7949 \\
ETTm1 & 0 & 48 & rls\_bank\_med - tafas$^{*}$ & -0.7893 & -0.8851 & -0.7007 \\
ETTm1 & 0 & 48 & rls\_bank\_med - tafas\_tuned$^{*}$ & -0.7900 & -0.8853 & -0.6998 \\
ETTm1 & 0 & 48 & tafas - frozen$^{*}$ & -0.2028 & -0.2384 & -0.1695 \\
ETTm1 & 0 & 48 & tafas\_tuned - frozen$^{*}$ & -0.2021 & -0.2322 & -0.1713 \\
ETTm1 & 0 & 48 & tafas\_tuned - tafas & 0.0006 & -0.0122 & 0.0123 \\
ETTm1 & 0 & 96 & cosa - frozen$^{*}$ & -0.7294 & -0.8234 & -0.6513 \\
ETTm1 & 0 & 96 & cosa\_tuned - cosa$^{*}$ & 0.3182 & 0.2730 & 0.3675 \\
ETTm1 & 0 & 96 & cosa\_tuned - frozen$^{*}$ & -0.4113 & -0.4646 & -0.3608 \\
ETTm1 & 0 & 96 & dynatta - frozen$^{*}$ & -0.9935 & -1.1176 & -0.8857 \\
ETTm1 & 0 & 96 & elf\_linear - rls\_bank\_med$^{*}$ & -0.1643 & -0.1924 & -0.1406 \\
ETTm1 & 0 & 96 & elf\_mix - cosa$^{*}$ & -0.3550 & -0.3976 & -0.3143 \\
ETTm1 & 0 & 96 & elf\_mix - cosa\_tuned$^{*}$ & -0.6731 & -0.7469 & -0.6044 \\
ETTm1 & 0 & 96 & elf\_mix - dynatta$^{*}$ & -0.0909 & -0.1098 & -0.0719 \\
ETTm1 & 0 & 96 & elf\_mix - frozen$^{*}$ & -1.0844 & -1.1973 & -0.9754 \\
ETTm1 & 0 & 96 & elf\_mix - naive$^{*}$ & -1.2175 & -1.3356 & -1.1019 \\
ETTm1 & 0 & 96 & elf\_mix - petsa$^{*}$ & -0.9930 & -1.1072 & -0.8894 \\
ETTm1 & 0 & 96 & elf\_mix - rls\_bank\_med$^{*}$ & -0.1005 & -0.1248 & -0.0780 \\
ETTm1 & 0 & 96 & elf\_mix - tafas$^{*}$ & -0.8816 & -0.9804 & -0.7841 \\
ETTm1 & 0 & 96 & elf\_mix - tafas\_tuned$^{*}$ & -0.8824 & -0.9779 & -0.7960 \\
ETTm1 & 0 & 96 & petsa - frozen$^{*}$ & -0.0914 & -0.1124 & -0.0693 \\
ETTm1 & 0 & 96 & rls\_bank\_med - cosa$^{*}$ & -0.2545 & -0.3022 & -0.2099 \\
ETTm1 & 0 & 96 & rls\_bank\_med - cosa\_tuned$^{*}$ & -0.5726 & -0.6481 & -0.4986 \\
ETTm1 & 0 & 96 & rls\_bank\_med - dynatta & 0.0096 & -0.0154 & 0.0348 \\
ETTm1 & 0 & 96 & rls\_bank\_med - frozen$^{*}$ & -0.9839 & -1.1036 & -0.8624 \\
ETTm1 & 0 & 96 & rls\_bank\_med - naive$^{*}$ & -1.1170 & -1.2326 & -0.9972 \\
ETTm1 & 0 & 96 & rls\_bank\_med - petsa$^{*}$ & -0.8925 & -0.9956 & -0.7811 \\
ETTm1 & 0 & 96 & rls\_bank\_med - tafas$^{*}$ & -0.7810 & -0.8720 & -0.6858 \\
ETTm1 & 0 & 96 & rls\_bank\_med - tafas\_tuned$^{*}$ & -0.7818 & -0.8744 & -0.6906 \\
ETTm1 & 0 & 96 & tafas - frozen$^{*}$ & -0.2028 & -0.2386 & -0.1678 \\
ETTm1 & 0 & 96 & tafas\_tuned - frozen$^{*}$ & -0.2020 & -0.2340 & -0.1717 \\
ETTm1 & 0 & 96 & tafas\_tuned - tafas & 0.0008 & -0.0123 & 0.0136 \\
ETTm1 & 1 & 24 & cosa - frozen$^{*}$ & -0.4586 & -0.5269 & -0.3893 \\
ETTm1 & 1 & 24 & dynatta - frozen$^{*}$ & -0.8714 & -0.9815 & -0.7763 \\
ETTm1 & 1 & 24 & elf\_linear - rls\_bank\_med$^{*}$ & -0.1394 & -0.1595 & -0.1220 \\
ETTm1 & 1 & 24 & elf\_mix - cosa$^{*}$ & -0.4761 & -0.5324 & -0.4195 \\
ETTm1 & 1 & 24 & elf\_mix - dynatta$^{*}$ & -0.0633 & -0.0846 & -0.0431 \\
ETTm1 & 1 & 24 & elf\_mix - frozen$^{*}$ & -0.9348 & -1.0375 & -0.8465 \\
ETTm1 & 1 & 24 & elf\_mix - naive$^{*}$ & -1.2385 & -1.3568 & -1.1202 \\
ETTm1 & 1 & 24 & elf\_mix - petsa$^{*}$ & -0.9348 & -1.0222 & -0.8377 \\
ETTm1 & 1 & 24 & elf\_mix - rls\_bank\_med$^{*}$ & -0.0921 & -0.1079 & -0.0763 \\
ETTm1 & 1 & 24 & elf\_mix - tafas$^{*}$ & -0.8547 & -0.9480 & -0.7591 \\
ETTm1 & 1 & 24 & petsa - frozen & 0.0000 & -0.0078 & 0.0079 \\
ETTm1 & 1 & 24 & rls\_bank\_med - cosa$^{*}$ & -0.3841 & -0.4533 & -0.3241 \\
ETTm1 & 1 & 24 & rls\_bank\_med - dynatta$^{*}$ & 0.0287 & 0.0035 & 0.0543 \\
ETTm1 & 1 & 24 & rls\_bank\_med - frozen$^{*}$ & -0.8427 & -0.9453 & -0.7477 \\
ETTm1 & 1 & 24 & rls\_bank\_med - naive$^{*}$ & -1.1464 & -1.2622 & -1.0374 \\
ETTm1 & 1 & 24 & rls\_bank\_med - petsa$^{*}$ & -0.8427 & -0.9410 & -0.7368 \\
ETTm1 & 1 & 24 & rls\_bank\_med - tafas$^{*}$ & -0.7626 & -0.8513 & -0.6743 \\
ETTm1 & 1 & 24 & tafas - frozen$^{*}$ & -0.0801 & -0.0952 & -0.0644 \\
ETTm1 & 1 & 48 & cosa - frozen$^{*}$ & -0.5908 & -0.6653 & -0.5120 \\
ETTm1 & 1 & 48 & elf\_linear - rls\_bank\_med$^{*}$ & -0.1422 & -0.1623 & -0.1242 \\
ETTm1 & 1 & 48 & elf\_mix - cosa$^{*}$ & -0.3381 & -0.3790 & -0.2982 \\
ETTm1 & 1 & 48 & elf\_mix - frozen$^{*}$ & -0.9289 & -1.0259 & -0.8361 \\
ETTm1 & 1 & 48 & elf\_mix - naive$^{*}$ & -1.2326 & -1.3482 & -1.1230 \\
ETTm1 & 1 & 48 & elf\_mix - rls\_bank\_med$^{*}$ & -0.0900 & -0.1076 & -0.0725 \\
ETTm1 & 1 & 48 & elf\_mix - tafas$^{*}$ & -0.8489 & -0.9384 & -0.7627 \\
ETTm1 & 1 & 48 & rls\_bank\_med - cosa$^{*}$ & -0.2481 & -0.2879 & -0.2059 \\
ETTm1 & 1 & 48 & rls\_bank\_med - frozen$^{*}$ & -0.8389 & -0.9526 & -0.7429 \\
ETTm1 & 1 & 48 & rls\_bank\_med - naive$^{*}$ & -1.1426 & -1.2623 & -1.0241 \\
ETTm1 & 1 & 48 & rls\_bank\_med - tafas$^{*}$ & -0.7589 & -0.8624 & -0.6613 \\
ETTm1 & 1 & 48 & tafas - frozen$^{*}$ & -0.0800 & -0.0940 & -0.0655 \\
ETTm1 & 1 & 96 & cosa - frozen$^{*}$ & -0.5654 & -0.6343 & -0.5029 \\
ETTm1 & 1 & 96 & elf\_linear - rls\_bank\_med$^{*}$ & -0.1451 & -0.1673 & -0.1241 \\
ETTm1 & 1 & 96 & elf\_mix - cosa$^{*}$ & -0.3514 & -0.3940 & -0.3083 \\
ETTm1 & 1 & 96 & elf\_mix - frozen$^{*}$ & -0.9168 & -1.0128 & -0.8210 \\
ETTm1 & 1 & 96 & elf\_mix - naive$^{*}$ & -1.2205 & -1.3300 & -1.1143 \\
ETTm1 & 1 & 96 & elf\_mix - rls\_bank\_med$^{*}$ & -0.0844 & -0.1043 & -0.0659 \\
ETTm1 & 1 & 96 & elf\_mix - tafas$^{*}$ & -0.8380 & -0.9316 & -0.7513 \\
ETTm1 & 1 & 96 & rls\_bank\_med - cosa$^{*}$ & -0.2670 & -0.3133 & -0.2146 \\
ETTm1 & 1 & 96 & rls\_bank\_med - frozen$^{*}$ & -0.8324 & -0.9314 & -0.7352 \\
ETTm1 & 1 & 96 & rls\_bank\_med - naive$^{*}$ & -1.1362 & -1.2445 & -1.0314 \\
ETTm1 & 1 & 96 & rls\_bank\_med - tafas$^{*}$ & -0.7536 & -0.8500 & -0.6591 \\
ETTm1 & 1 & 96 & tafas - frozen$^{*}$ & -0.0788 & -0.0931 & -0.0648 \\
ETTm1 & 2 & 24 & cosa - frozen$^{*}$ & -0.4062 & -0.4617 & -0.3478 \\
ETTm1 & 2 & 24 & dynatta - frozen$^{*}$ & -0.8007 & -0.8964 & -0.7044 \\
ETTm1 & 2 & 24 & elf\_linear - rls\_bank\_med$^{*}$ & -0.1420 & -0.1617 & -0.1243 \\
ETTm1 & 2 & 24 & elf\_mix - cosa$^{*}$ & -0.4407 & -0.4969 & -0.3842 \\
ETTm1 & 2 & 24 & elf\_mix - dynatta$^{*}$ & -0.0462 & -0.0609 & -0.0321 \\
ETTm1 & 2 & 24 & elf\_mix - frozen$^{*}$ & -0.8469 & -0.9291 & -0.7596 \\
ETTm1 & 2 & 24 & elf\_mix - naive$^{*}$ & -1.2408 & -1.3563 & -1.1310 \\
ETTm1 & 2 & 24 & elf\_mix - petsa$^{*}$ & -0.8504 & -0.9482 & -0.7593 \\
ETTm1 & 2 & 24 & elf\_mix - rls\_bank\_med$^{*}$ & -0.0970 & -0.1155 & -0.0819 \\
ETTm1 & 2 & 24 & elf\_mix - tafas$^{*}$ & -0.7657 & -0.8419 & -0.6877 \\
ETTm1 & 2 & 24 & petsa - frozen & 0.0035 & -0.0042 & 0.0117 \\
ETTm1 & 2 & 24 & rls\_bank\_med - cosa$^{*}$ & -0.3437 & -0.4047 & -0.2883 \\
ETTm1 & 2 & 24 & rls\_bank\_med - dynatta$^{*}$ & 0.0507 & 0.0310 & 0.0708 \\
ETTm1 & 2 & 24 & rls\_bank\_med - frozen$^{*}$ & -0.7499 & -0.8549 & -0.6544 \\
ETTm1 & 2 & 24 & rls\_bank\_med - naive$^{*}$ & -1.1438 & -1.2594 & -1.0259 \\
ETTm1 & 2 & 24 & rls\_bank\_med - petsa$^{*}$ & -0.7535 & -0.8472 & -0.6662 \\
ETTm1 & 2 & 24 & rls\_bank\_med - tafas$^{*}$ & -0.6688 & -0.7535 & -0.5871 \\
ETTm1 & 2 & 24 & tafas - frozen$^{*}$ & -0.0812 & -0.0979 & -0.0658 \\
ETTm1 & 2 & 48 & cosa - frozen$^{*}$ & -0.5252 & -0.5913 & -0.4646 \\
ETTm1 & 2 & 48 & elf\_linear - rls\_bank\_med$^{*}$ & -0.1449 & -0.1686 & -0.1255 \\
ETTm1 & 2 & 48 & elf\_mix - cosa$^{*}$ & -0.3156 & -0.3521 & -0.2747 \\
ETTm1 & 2 & 48 & elf\_mix - frozen$^{*}$ & -0.8409 & -0.9300 & -0.7550 \\
ETTm1 & 2 & 48 & elf\_mix - naive$^{*}$ & -1.2347 & -1.3448 & -1.1271 \\
ETTm1 & 2 & 48 & elf\_mix - rls\_bank\_med$^{*}$ & -0.0948 & -0.1120 & -0.0773 \\
ETTm1 & 2 & 48 & elf\_mix - tafas$^{*}$ & -0.7598 & -0.8430 & -0.6870 \\
ETTm1 & 2 & 48 & rls\_bank\_med - cosa$^{*}$ & -0.2208 & -0.2619 & -0.1805 \\
ETTm1 & 2 & 48 & rls\_bank\_med - frozen$^{*}$ & -0.7460 & -0.8362 & -0.6557 \\
ETTm1 & 2 & 48 & rls\_bank\_med - naive$^{*}$ & -1.1399 & -1.2562 & -1.0290 \\
ETTm1 & 2 & 48 & rls\_bank\_med - tafas$^{*}$ & -0.6649 & -0.7464 & -0.5912 \\
ETTm1 & 2 & 48 & tafas - frozen$^{*}$ & -0.0811 & -0.0983 & -0.0663 \\
ETTm1 & 2 & 96 & cosa - frozen$^{*}$ & -0.4962 & -0.5643 & -0.4199 \\
ETTm1 & 2 & 96 & elf\_linear - rls\_bank\_med$^{*}$ & -0.1476 & -0.1685 & -0.1268 \\
ETTm1 & 2 & 96 & elf\_mix - cosa$^{*}$ & -0.3332 & -0.3749 & -0.2899 \\
ETTm1 & 2 & 96 & elf\_mix - frozen$^{*}$ & -0.8294 & -0.9266 & -0.7370 \\
ETTm1 & 2 & 96 & elf\_mix - naive$^{*}$ & -1.2233 & -1.3340 & -1.1232 \\
ETTm1 & 2 & 96 & elf\_mix - rls\_bank\_med$^{*}$ & -0.0895 & -0.1091 & -0.0711 \\
ETTm1 & 2 & 96 & elf\_mix - tafas$^{*}$ & -0.7485 & -0.8369 & -0.6756 \\
ETTm1 & 2 & 96 & rls\_bank\_med - cosa$^{*}$ & -0.2437 & -0.2893 & -0.2009 \\
ETTm1 & 2 & 96 & rls\_bank\_med - frozen$^{*}$ & -0.7398 & -0.8248 & -0.6490 \\
ETTm1 & 2 & 96 & rls\_bank\_med - naive$^{*}$ & -1.1337 & -1.2456 & -1.0085 \\
ETTm1 & 2 & 96 & rls\_bank\_med - tafas$^{*}$ & -0.6590 & -0.7406 & -0.5791 \\
ETTm1 & 2 & 96 & tafas - frozen$^{*}$ & -0.0808 & -0.0977 & -0.0633 \\
\end{longtable}
}

{\tiny
\setlength{\tabcolsep}{4pt}
\begin{longtable}[c]{llllrrr}
\caption{External comparisons on the authors' own ETTh2 checkpoints, per checkpoint seed (same conventions as Table~\ref{tab:ci_external_ettm1}).}
\label{tab:ci_external_etth2}\\
\toprule
dataset & seed & delay & pair & mean\_diff & ci\_lo & ci\_hi \\
\midrule
\endfirsthead
\toprule
dataset & seed & delay & pair & mean\_diff & ci\_lo & ci\_hi \\
\midrule
\endhead
\midrule
\multicolumn{7}{r}{\scriptsize continues on next page}\\
\endfoot
\bottomrule
\endlastfoot
ETTh2 & 0 & 24 & cosa - frozen & 0.2939 & -0.0715 & 0.8575 \\
ETTh2 & 0 & 24 & cosa\_tuned - cosa & -0.2937 & -0.8590 & 0.0494 \\
ETTh2 & 0 & 24 & cosa\_tuned - frozen & 0.0002 & -0.0101 & 0.0090 \\
ETTh2 & 0 & 24 & dynatta - frozen$^{*}$ & 0.1317 & 0.0300 & 0.2502 \\
ETTh2 & 0 & 24 & elf\_linear - rls\_bank\_med & -0.1259 & -0.3579 & 0.0411 \\
ETTh2 & 0 & 24 & elf\_mix - cosa$^{*}$ & -0.4937 & -1.2319 & -0.0534 \\
ETTh2 & 0 & 24 & elf\_mix - cosa\_tuned & -0.2001 & -0.4264 & 0.0029 \\
ETTh2 & 0 & 24 & elf\_mix - dynatta$^{*}$ & -0.3315 & -0.5685 & -0.1449 \\
ETTh2 & 0 & 24 & elf\_mix - frozen & -0.1999 & -0.4535 & 0.0110 \\
ETTh2 & 0 & 24 & elf\_mix - naive & -0.0840 & -0.2372 & 0.1392 \\
ETTh2 & 0 & 24 & elf\_mix - petsa$^{*}$ & -0.2868 & -0.5794 & -0.0648 \\
ETTh2 & 0 & 24 & elf\_mix - rls\_bank\_med$^{*}$ & -0.2780 & -0.5516 & -0.0850 \\
ETTh2 & 0 & 24 & elf\_mix - tafas$^{*}$ & -0.2107 & -0.5127 & -0.0095 \\
ETTh2 & 0 & 24 & elf\_mix - tafas\_tuned & -0.2071 & -0.5230 & 0.0006 \\
ETTh2 & 0 & 24 & petsa - frozen$^{*}$ & 0.0869 & 0.0320 & 0.1674 \\
ETTh2 & 0 & 24 & rls\_bank\_med - cosa & -0.2157 & -0.6926 & 0.0582 \\
ETTh2 & 0 & 24 & rls\_bank\_med - cosa\_tuned & 0.0780 & -0.0561 & 0.2421 \\
ETTh2 & 0 & 24 & rls\_bank\_med - dynatta & -0.0686 & -0.1924 & 0.0517 \\
ETTh2 & 0 & 24 & rls\_bank\_med - frozen & 0.0782 & -0.0559 & 0.2657 \\
ETTh2 & 0 & 24 & rls\_bank\_med - naive & 0.1941 & -0.1444 & 0.7012 \\
ETTh2 & 0 & 24 & rls\_bank\_med - petsa & -0.0088 & -0.1659 & 0.1795 \\
ETTh2 & 0 & 24 & rls\_bank\_med - tafas & 0.0673 & -0.0543 & 0.2255 \\
ETTh2 & 0 & 24 & rls\_bank\_med - tafas\_tuned & 0.0709 & -0.0710 & 0.2561 \\
ETTh2 & 0 & 24 & tafas - frozen & 0.0108 & -0.0031 & 0.0247 \\
ETTh2 & 0 & 24 & tafas\_tuned - frozen & 0.0073 & -0.0047 & 0.0244 \\
ETTh2 & 0 & 24 & tafas\_tuned - tafas & -0.0036 & -0.0130 & 0.0071 \\
ETTh2 & 0 & 48 & cosa - frozen & 0.5269 & -0.0724 & 1.3992 \\
ETTh2 & 0 & 48 & cosa\_tuned - cosa & -0.5264 & -1.3067 & 0.0409 \\
ETTh2 & 0 & 48 & cosa\_tuned - frozen & 0.0005 & -0.0030 & 0.0039 \\
ETTh2 & 0 & 48 & dynatta - frozen$^{*}$ & 0.2068 & 0.0947 & 0.3519 \\
ETTh2 & 0 & 48 & elf\_linear - rls\_bank\_med & -0.1236 & -0.4504 & 0.0862 \\
ETTh2 & 0 & 48 & elf\_mix - cosa$^{*}$ & -0.6756 & -1.4685 & -0.1373 \\
ETTh2 & 0 & 48 & elf\_mix - cosa\_tuned & -0.1493 & -0.4233 & 0.0490 \\
ETTh2 & 0 & 48 & elf\_mix - dynatta$^{*}$ & -0.3556 & -0.7491 & -0.0861 \\
ETTh2 & 0 & 48 & elf\_mix - frozen & -0.1488 & -0.4727 & 0.0582 \\
ETTh2 & 0 & 48 & elf\_mix - naive & -0.0329 & -0.2346 & 0.2101 \\
ETTh2 & 0 & 48 & elf\_mix - petsa$^{*}$ & -0.2473 & -0.5524 & -0.0050 \\
ETTh2 & 0 & 48 & elf\_mix - rls\_bank\_med$^{*}$ & -0.3041 & -0.6171 & -0.0805 \\
ETTh2 & 0 & 48 & elf\_mix - tafas & -0.1747 & -0.5168 & 0.0691 \\
ETTh2 & 0 & 48 & elf\_mix - tafas\_tuned & -0.1543 & -0.4894 & 0.0664 \\
ETTh2 & 0 & 48 & petsa - frozen$^{*}$ & 0.0986 & 0.0364 & 0.1757 \\
ETTh2 & 0 & 48 & rls\_bank\_med - cosa & -0.3716 & -1.0775 & 0.1319 \\
ETTh2 & 0 & 48 & rls\_bank\_med - cosa\_tuned$^{*}$ & 0.1548 & 0.0001 & 0.3650 \\
ETTh2 & 0 & 48 & rls\_bank\_med - dynatta & -0.0515 & -0.2149 & 0.1428 \\
ETTh2 & 0 & 48 & rls\_bank\_med - frozen$^{*}$ & 0.1553 & 0.0005 & 0.3896 \\
ETTh2 & 0 & 48 & rls\_bank\_med - naive & 0.2712 & -0.1334 & 0.8176 \\
ETTh2 & 0 & 48 & rls\_bank\_med - petsa & 0.0568 & -0.1254 & 0.2573 \\
ETTh2 & 0 & 48 & rls\_bank\_med - tafas & 0.1294 & -0.0015 & 0.3550 \\
ETTh2 & 0 & 48 & rls\_bank\_med - tafas\_tuned & 0.1498 & -0.0187 & 0.3655 \\
ETTh2 & 0 & 48 & tafas - frozen & 0.0259 & -0.0001 & 0.0605 \\
ETTh2 & 0 & 48 & tafas\_tuned - frozen & 0.0056 & -0.0045 & 0.0191 \\
ETTh2 & 0 & 48 & tafas\_tuned - tafas$^{*}$ & -0.0203 & -0.0476 & -0.0018 \\
ETTh2 & 0 & 96 & cosa - frozen$^{*}$ & 0.7597 & 0.1082 & 1.6112 \\
ETTh2 & 0 & 96 & cosa\_tuned - cosa$^{*}$ & -0.7599 & -1.6141 & -0.0736 \\
ETTh2 & 0 & 96 & cosa\_tuned - frozen & -0.0001 & -0.0022 & 0.0018 \\
ETTh2 & 0 & 96 & dynatta - frozen$^{*}$ & 0.2146 & 0.0858 & 0.3874 \\
ETTh2 & 0 & 96 & elf\_linear - rls\_bank\_med & -0.1455 & -0.5625 & 0.2159 \\
ETTh2 & 0 & 96 & elf\_mix - cosa$^{*}$ & -0.7523 & -1.5237 & -0.1936 \\
ETTh2 & 0 & 96 & elf\_mix - cosa\_tuned & 0.0076 & -0.1558 & 0.1906 \\
ETTh2 & 0 & 96 & elf\_mix - dynatta & -0.2072 & -0.4843 & 0.0237 \\
ETTh2 & 0 & 96 & elf\_mix - frozen & 0.0075 & -0.1625 & 0.2046 \\
ETTh2 & 0 & 96 & elf\_mix - naive & 0.1234 & -0.1751 & 0.5152 \\
ETTh2 & 0 & 96 & elf\_mix - petsa & -0.0899 & -0.3114 & 0.1072 \\
ETTh2 & 0 & 96 & elf\_mix - rls\_bank\_med$^{*}$ & -0.3467 & -0.6594 & -0.1166 \\
ETTh2 & 0 & 96 & elf\_mix - tafas & -0.0129 & -0.1889 & 0.1788 \\
ETTh2 & 0 & 96 & elf\_mix - tafas\_tuned & 0.0042 & -0.1576 & 0.1826 \\
ETTh2 & 0 & 96 & petsa - frozen$^{*}$ & 0.0974 & 0.0380 & 0.1886 \\
ETTh2 & 0 & 96 & rls\_bank\_med - cosa & -0.4055 & -1.1203 & 0.1566 \\
ETTh2 & 0 & 96 & rls\_bank\_med - cosa\_tuned$^{*}$ & 0.3544 & 0.0769 & 0.7154 \\
ETTh2 & 0 & 96 & rls\_bank\_med - dynatta & 0.1396 & -0.0738 & 0.3953 \\
ETTh2 & 0 & 96 & rls\_bank\_med - frozen$^{*}$ & 0.3542 & 0.0723 & 0.7038 \\
ETTh2 & 0 & 96 & rls\_bank\_med - naive & 0.4701 & -0.0861 & 1.1040 \\
ETTh2 & 0 & 96 & rls\_bank\_med - petsa & 0.2568 & -0.0095 & 0.5969 \\
ETTh2 & 0 & 96 & rls\_bank\_med - tafas$^{*}$ & 0.3339 & 0.0726 & 0.6872 \\
ETTh2 & 0 & 96 & rls\_bank\_med - tafas\_tuned$^{*}$ & 0.3509 & 0.0644 & 0.6962 \\
ETTh2 & 0 & 96 & tafas - frozen$^{*}$ & 0.0204 & 0.0004 & 0.0510 \\
ETTh2 & 0 & 96 & tafas\_tuned - frozen & 0.0033 & -0.0031 & 0.0111 \\
ETTh2 & 0 & 96 & tafas\_tuned - tafas$^{*}$ & -0.0171 & -0.0415 & -0.0014 \\
ETTh2 & 1 & 24 & cosa - frozen & 0.3958 & -0.0483 & 1.2160 \\
ETTh2 & 1 & 24 & dynatta - frozen & 0.1146 & -0.0476 & 0.2987 \\
ETTh2 & 1 & 24 & elf\_linear - rls\_bank\_med & -0.3248 & -0.8866 & 0.0154 \\
ETTh2 & 1 & 24 & elf\_mix - cosa$^{*}$ & -0.7167 & -1.7815 & -0.0790 \\
ETTh2 & 1 & 24 & elf\_mix - dynatta$^{*}$ & -0.4355 & -0.8025 & -0.1587 \\
ETTh2 & 1 & 24 & elf\_mix - frozen$^{*}$ & -0.3209 & -0.7779 & -0.0265 \\
ETTh2 & 1 & 24 & elf\_mix - naive & -0.0504 & -0.2336 & 0.2050 \\
ETTh2 & 1 & 24 & elf\_mix - petsa$^{*}$ & -0.3915 & -0.9244 & -0.0651 \\
ETTh2 & 1 & 24 & elf\_mix - rls\_bank\_med$^{*}$ & -0.4433 & -0.9519 & -0.0881 \\
ETTh2 & 1 & 24 & elf\_mix - tafas$^{*}$ & -0.3340 & -0.7834 & -0.0275 \\
ETTh2 & 1 & 24 & petsa - frozen$^{*}$ & 0.0706 & 0.0319 & 0.1175 \\
ETTh2 & 1 & 24 & rls\_bank\_med - cosa & -0.2734 & -0.8668 & 0.0982 \\
ETTh2 & 1 & 24 & rls\_bank\_med - dynatta & 0.0078 & -0.1436 & 0.1940 \\
ETTh2 & 1 & 24 & rls\_bank\_med - frozen & 0.1224 & -0.0058 & 0.3214 \\
ETTh2 & 1 & 24 & rls\_bank\_med - naive & 0.3929 & -0.1255 & 1.1721 \\
ETTh2 & 1 & 24 & rls\_bank\_med - petsa & 0.0518 & -0.0710 & 0.2054 \\
ETTh2 & 1 & 24 & rls\_bank\_med - tafas$^{*}$ & 0.1093 & 0.0012 & 0.2775 \\
ETTh2 & 1 & 24 & tafas - frozen & 0.0131 & -0.0044 & 0.0318 \\
ETTh2 & 1 & 48 & cosa - frozen & 0.5452 & -0.0798 & 1.3995 \\
ETTh2 & 1 & 48 & elf\_linear - rls\_bank\_med & -0.2944 & -0.8788 & 0.0921 \\
ETTh2 & 1 & 48 & elf\_mix - cosa$^{*}$ & -0.7730 & -1.6687 & -0.2176 \\
ETTh2 & 1 & 48 & elf\_mix - frozen & -0.2278 & -0.6602 & 0.0494 \\
ETTh2 & 1 & 48 & elf\_mix - naive & 0.0428 & -0.2181 & 0.4018 \\
ETTh2 & 1 & 48 & elf\_mix - rls\_bank\_med$^{*}$ & -0.3993 & -0.8948 & -0.1020 \\
ETTh2 & 1 & 48 & elf\_mix - tafas & -0.2627 & -0.7261 & 0.0401 \\
ETTh2 & 1 & 48 & rls\_bank\_med - cosa & -0.3737 & -1.0736 & 0.2004 \\
ETTh2 & 1 & 48 & rls\_bank\_med - frozen$^{*}$ & 0.1715 & 0.0222 & 0.3721 \\
ETTh2 & 1 & 48 & rls\_bank\_med - naive & 0.4421 & -0.0784 & 1.1595 \\
ETTh2 & 1 & 48 & rls\_bank\_med - tafas$^{*}$ & 0.1366 & 0.0137 & 0.3255 \\
ETTh2 & 1 & 48 & tafas - frozen & 0.0350 & -0.0001 & 0.0759 \\
ETTh2 & 1 & 96 & cosa - frozen$^{*}$ & 0.7957 & 0.0936 & 1.7247 \\
ETTh2 & 1 & 96 & elf\_linear - rls\_bank\_med & -0.2747 & -0.9025 & 0.1536 \\
ETTh2 & 1 & 96 & elf\_mix - cosa$^{*}$ & -0.8330 & -1.6660 & -0.2243 \\
ETTh2 & 1 & 96 & elf\_mix - frozen & -0.0374 & -0.2457 & 0.1803 \\
ETTh2 & 1 & 96 & elf\_mix - naive & 0.2332 & -0.1250 & 0.7579 \\
ETTh2 & 1 & 96 & elf\_mix - rls\_bank\_med$^{*}$ & -0.3661 & -0.7026 & -0.1386 \\
ETTh2 & 1 & 96 & elf\_mix - tafas & -0.0640 & -0.2970 & 0.1709 \\
ETTh2 & 1 & 96 & rls\_bank\_med - cosa & -0.4669 & -1.1976 & 0.0499 \\
ETTh2 & 1 & 96 & rls\_bank\_med - frozen$^{*}$ & 0.3287 & 0.0747 & 0.7044 \\
ETTh2 & 1 & 96 & rls\_bank\_med - naive & 0.5993 & -0.0090 & 1.4156 \\
ETTh2 & 1 & 96 & rls\_bank\_med - tafas$^{*}$ & 0.3021 & 0.0701 & 0.6072 \\
ETTh2 & 1 & 96 & tafas - frozen & 0.0266 & -0.0016 & 0.0644 \\
ETTh2 & 2 & 24 & cosa - frozen & 0.3949 & -0.0496 & 1.0263 \\
ETTh2 & 2 & 24 & dynatta - frozen & 0.0828 & -0.1599 & 0.2896 \\
ETTh2 & 2 & 24 & elf\_linear - rls\_bank\_med$^{*}$ & -0.5089 & -1.2433 & -0.0260 \\
ETTh2 & 2 & 24 & elf\_mix - cosa$^{*}$ & -0.7977 & -1.8412 & -0.1100 \\
ETTh2 & 2 & 24 & elf\_mix - dynatta$^{*}$ & -0.4855 & -0.8557 & -0.2064 \\
ETTh2 & 2 & 24 & elf\_mix - frozen$^{*}$ & -0.4028 & -0.9323 & -0.0544 \\
ETTh2 & 2 & 24 & elf\_mix - naive & -0.0297 & -0.2246 & 0.2645 \\
ETTh2 & 2 & 24 & elf\_mix - petsa$^{*}$ & -0.4501 & -0.9930 & -0.0960 \\
ETTh2 & 2 & 24 & elf\_mix - rls\_bank\_med$^{*}$ & -0.6067 & -1.3121 & -0.1534 \\
ETTh2 & 2 & 24 & elf\_mix - tafas$^{*}$ & -0.4199 & -1.0155 & -0.0461 \\
ETTh2 & 2 & 24 & petsa - frozen$^{*}$ & 0.0473 & 0.0004 & 0.1066 \\
ETTh2 & 2 & 24 & rls\_bank\_med - cosa & -0.1910 & -0.6263 & 0.1334 \\
ETTh2 & 2 & 24 & rls\_bank\_med - dynatta & 0.1212 & -0.1229 & 0.5470 \\
ETTh2 & 2 & 24 & rls\_bank\_med - frozen$^{*}$ & 0.2039 & 0.0158 & 0.4719 \\
ETTh2 & 2 & 24 & rls\_bank\_med - naive & 0.5770 & -0.0761 & 1.3701 \\
ETTh2 & 2 & 24 & rls\_bank\_med - petsa & 0.1566 & -0.0254 & 0.4492 \\
ETTh2 & 2 & 24 & rls\_bank\_med - tafas$^{*}$ & 0.1868 & 0.0096 & 0.4582 \\
ETTh2 & 2 & 24 & tafas - frozen & 0.0171 & -0.0012 & 0.0402 \\
ETTh2 & 2 & 48 & cosa - frozen & 0.4424 & -0.0765 & 1.2968 \\
ETTh2 & 2 & 48 & elf\_linear - rls\_bank\_med$^{*}$ & -0.4899 & -1.1512 & -0.0096 \\
ETTh2 & 2 & 48 & elf\_mix - cosa$^{*}$ & -0.7315 & -1.4982 & -0.2314 \\
ETTh2 & 2 & 48 & elf\_mix - frozen & -0.2891 & -0.7523 & 0.0092 \\
ETTh2 & 2 & 48 & elf\_mix - naive & 0.0840 & -0.2186 & 0.5615 \\
ETTh2 & 2 & 48 & elf\_mix - rls\_bank\_med$^{*}$ & -0.5535 & -1.2054 & -0.1260 \\
ETTh2 & 2 & 48 & elf\_mix - tafas & -0.3250 & -0.7516 & 0.0009 \\
ETTh2 & 2 & 48 & rls\_bank\_med - cosa & -0.1780 & -0.7479 & 0.3456 \\
ETTh2 & 2 & 48 & rls\_bank\_med - frozen$^{*}$ & 0.2644 & 0.0368 & 0.5783 \\
ETTh2 & 2 & 48 & rls\_bank\_med - naive & 0.6375 & -0.0239 & 1.6291 \\
ETTh2 & 2 & 48 & rls\_bank\_med - tafas$^{*}$ & 0.2285 & 0.0380 & 0.4781 \\
ETTh2 & 2 & 48 & tafas - frozen$^{*}$ & 0.0360 & 0.0002 & 0.0920 \\
ETTh2 & 2 & 96 & cosa - frozen$^{*}$ & 0.7153 & 0.0966 & 1.6422 \\
ETTh2 & 2 & 96 & elf\_linear - rls\_bank\_med & -0.4839 & -1.2168 & 0.0116 \\
ETTh2 & 2 & 96 & elf\_mix - cosa$^{*}$ & -0.8049 & -1.5961 & -0.2998 \\
ETTh2 & 2 & 96 & elf\_mix - frozen & -0.0896 & -0.3456 & 0.1934 \\
ETTh2 & 2 & 96 & elf\_mix - naive & 0.2835 & -0.1805 & 0.9131 \\
ETTh2 & 2 & 96 & elf\_mix - rls\_bank\_med$^{*}$ & -0.5250 & -0.9476 & -0.1978 \\
ETTh2 & 2 & 96 & elf\_mix - tafas & -0.1202 & -0.4044 & 0.1463 \\
ETTh2 & 2 & 96 & rls\_bank\_med - cosa & -0.2799 & -0.9651 & 0.2307 \\
ETTh2 & 2 & 96 & rls\_bank\_med - frozen$^{*}$ & 0.4354 & 0.0835 & 0.8974 \\
ETTh2 & 2 & 96 & rls\_bank\_med - naive$^{*}$ & 0.8085 & 0.0797 & 1.8412 \\
ETTh2 & 2 & 96 & rls\_bank\_med - tafas$^{*}$ & 0.4048 & 0.1004 & 0.8461 \\
ETTh2 & 2 & 96 & tafas - frozen & 0.0305 & -0.0001 & 0.0784 \\
\end{longtable}
}

{\tiny
\setlength{\tabcolsep}{4pt}
\begin{longtable}[c]{llllrrr}
\caption{External comparisons on the authors' own Weather, Electricity and Traffic checkpoints (one checkpoint each; same conventions as Table~\ref{tab:ci_external_ettm1}).}
\label{tab:ci_external_wet}\\
\toprule
dataset & seed & delay & pair & mean\_diff & ci\_lo & ci\_hi \\
\midrule
\endfirsthead
\toprule
dataset & seed & delay & pair & mean\_diff & ci\_lo & ci\_hi \\
\midrule
\endhead
\midrule
\multicolumn{7}{r}{\scriptsize continues on next page}\\
\endfoot
\bottomrule
\endlastfoot
electricity & 0 & 24 & cosa - frozen$^{*}$ & 0.7697 & 0.3705 & 1.2404 \\
electricity & 0 & 24 & dynatta - frozen$^{*}$ & 0.7014 & 0.5194 & 0.9116 \\
electricity & 0 & 24 & petsa - frozen$^{*}$ & 1.8665 & 1.7743 & 1.9607 \\
electricity & 0 & 24 & rls\_bank\_med - cosa$^{*}$ & -0.7663 & -1.1181 & -0.4957 \\
electricity & 0 & 24 & rls\_bank\_med - dynatta$^{*}$ & -0.6980 & -0.9423 & -0.4030 \\
electricity & 0 & 24 & rls\_bank\_med - frozen & 0.0034 & -0.1782 & 0.2374 \\
electricity & 0 & 24 & rls\_bank\_med - naive$^{*}$ & -67.0770 & -70.4049 & -64.2900 \\
electricity & 0 & 24 & rls\_bank\_med - petsa$^{*}$ & -1.8631 & -2.0952 & -1.6049 \\
electricity & 0 & 24 & rls\_bank\_med - tafas & -0.0487 & -0.2200 & 0.2159 \\
electricity & 0 & 24 & tafas - frozen$^{*}$ & 0.0521 & 0.0263 & 0.0808 \\
electricity & 0 & 48 & cosa - frozen$^{*}$ & 0.4518 & 0.3275 & 0.5862 \\
electricity & 0 & 48 & dynatta - frozen$^{*}$ & 0.4781 & 0.3365 & 0.6139 \\
electricity & 0 & 48 & petsa - frozen$^{*}$ & 1.8283 & 1.7353 & 1.9216 \\
electricity & 0 & 48 & rls\_bank\_med - cosa$^{*}$ & -0.4452 & -0.7216 & -0.1128 \\
electricity & 0 & 48 & rls\_bank\_med - dynatta$^{*}$ & -0.4715 & -0.7287 & -0.0983 \\
electricity & 0 & 48 & rls\_bank\_med - frozen & 0.0066 & -0.2449 & 0.3629 \\
electricity & 0 & 48 & rls\_bank\_med - naive$^{*}$ & -67.0738 & -70.2155 & -63.6950 \\
electricity & 0 & 48 & rls\_bank\_med - petsa$^{*}$ & -1.8217 & -2.1024 & -1.4562 \\
electricity & 0 & 48 & rls\_bank\_med - tafas & 0.0147 & -0.2495 & 0.3900 \\
electricity & 0 & 48 & tafas - frozen & -0.0081 & -0.0450 & 0.0277 \\
electricity & 0 & 96 & cosa - frozen$^{*}$ & 0.5282 & 0.4094 & 0.6752 \\
electricity & 0 & 96 & dynatta - frozen$^{*}$ & 0.4798 & 0.3271 & 0.6268 \\
electricity & 0 & 96 & petsa - frozen$^{*}$ & 1.8398 & 1.7469 & 1.9269 \\
electricity & 0 & 96 & rls\_bank\_med - cosa & -0.0505 & -0.5880 & 0.5725 \\
electricity & 0 & 96 & rls\_bank\_med - dynatta & -0.0022 & -0.5280 & 0.7276 \\
electricity & 0 & 96 & rls\_bank\_med - frozen & 0.4776 & -0.0012 & 1.1002 \\
electricity & 0 & 96 & rls\_bank\_med - naive$^{*}$ & -66.6027 & -70.0701 & -63.4074 \\
electricity & 0 & 96 & rls\_bank\_med - petsa$^{*}$ & -1.3622 & -1.8662 & -0.6644 \\
electricity & 0 & 96 & rls\_bank\_med - tafas & 0.4926 & -0.0397 & 1.1082 \\
electricity & 0 & 96 & tafas - frozen & -0.0149 & -0.0460 & 0.0167 \\
traffic & 0 & 24 & cosa - frozen & -0.0013 & -0.0032 & 0.0013 \\
traffic & 0 & 24 & petsa - frozen$^{*}$ & 0.0469 & 0.0447 & 0.0492 \\
traffic & 0 & 24 & rls\_bank\_med - cosa & -0.0025 & -0.0059 & 0.0023 \\
traffic & 0 & 24 & rls\_bank\_med - frozen & -0.0038 & -0.0073 & 0.0002 \\
traffic & 0 & 24 & rls\_bank\_med - naive$^{*}$ & -2.1646 & -2.2614 & -2.0714 \\
traffic & 0 & 24 & rls\_bank\_med - petsa$^{*}$ & -0.0507 & -0.0549 & -0.0462 \\
traffic & 0 & 24 & rls\_bank\_med - tafas & -0.0020 & -0.0051 & 0.0025 \\
traffic & 0 & 24 & tafas - frozen$^{*}$ & -0.0019 & -0.0030 & -0.0005 \\
traffic & 0 & 48 & cosa - frozen & 0.0001 & -0.0014 & 0.0021 \\
traffic & 0 & 48 & petsa - frozen$^{*}$ & 0.0423 & 0.0403 & 0.0443 \\
traffic & 0 & 48 & rls\_bank\_med - cosa$^{*}$ & -0.0065 & -0.0089 & -0.0036 \\
traffic & 0 & 48 & rls\_bank\_med - frozen$^{*}$ & -0.0063 & -0.0093 & -0.0031 \\
traffic & 0 & 48 & rls\_bank\_med - naive$^{*}$ & -2.1671 & -2.2729 & -2.0699 \\
traffic & 0 & 48 & rls\_bank\_med - petsa$^{*}$ & -0.0486 & -0.0524 & -0.0449 \\
traffic & 0 & 48 & rls\_bank\_med - tafas$^{*}$ & -0.0097 & -0.0121 & -0.0066 \\
traffic & 0 & 48 & tafas - frozen$^{*}$ & 0.0033 & 0.0017 & 0.0049 \\
traffic & 0 & 96 & cosa - frozen & -0.0009 & -0.0021 & 0.0002 \\
traffic & 0 & 96 & petsa - frozen$^{*}$ & 0.0419 & 0.0398 & 0.0440 \\
traffic & 0 & 96 & rls\_bank\_med - cosa$^{*}$ & -0.0065 & -0.0089 & -0.0039 \\
traffic & 0 & 96 & rls\_bank\_med - frozen$^{*}$ & -0.0075 & -0.0100 & -0.0044 \\
traffic & 0 & 96 & rls\_bank\_med - naive$^{*}$ & -2.1682 & -2.2777 & -2.0705 \\
traffic & 0 & 96 & rls\_bank\_med - petsa$^{*}$ & -0.0493 & -0.0528 & -0.0457 \\
traffic & 0 & 96 & rls\_bank\_med - tafas$^{*}$ & -0.0102 & -0.0129 & -0.0074 \\
traffic & 0 & 96 & tafas - frozen$^{*}$ & 0.0028 & 0.0013 & 0.0043 \\
weather & 0 & 24 & cosa - frozen & -0.0072 & -0.0206 & 0.0039 \\
weather & 0 & 24 & dynatta - frozen$^{*}$ & 0.0413 & 0.0241 & 0.0600 \\
weather & 0 & 24 & elf\_linear - rls\_bank\_med$^{*}$ & 0.1511 & 0.0269 & 0.3361 \\
weather & 0 & 24 & elf\_mix - cosa & -0.0382 & -0.0697 & 0.0155 \\
weather & 0 & 24 & elf\_mix - dynatta$^{*}$ & -0.0867 & -0.1180 & -0.0454 \\
weather & 0 & 24 & elf\_mix - frozen & -0.0454 & -0.0785 & 0.0073 \\
weather & 0 & 24 & elf\_mix - naive$^{*}$ & -0.4332 & -0.5431 & -0.3349 \\
weather & 0 & 24 & elf\_mix - petsa$^{*}$ & -0.1770 & -0.2131 & -0.1287 \\
weather & 0 & 24 & elf\_mix - rls\_bank\_med & -0.0137 & -0.0360 & 0.0170 \\
weather & 0 & 24 & elf\_mix - tafas$^{*}$ & -0.0459 & -0.0782 & -0.0002 \\
weather & 0 & 24 & petsa - frozen$^{*}$ & 0.1316 & 0.1112 & 0.1529 \\
weather & 0 & 24 & rls\_bank\_med - cosa & -0.0245 & -0.0738 & 0.0337 \\
weather & 0 & 24 & rls\_bank\_med - dynatta$^{*}$ & -0.0730 & -0.1147 & -0.0167 \\
weather & 0 & 24 & rls\_bank\_med - frozen & -0.0317 & -0.0837 & 0.0291 \\
weather & 0 & 24 & rls\_bank\_med - naive$^{*}$ & -0.4195 & -0.5490 & -0.3190 \\
weather & 0 & 24 & rls\_bank\_med - petsa$^{*}$ & -0.1633 & -0.2093 & -0.1088 \\
weather & 0 & 24 & rls\_bank\_med - tafas & -0.0322 & -0.0846 & 0.0281 \\
weather & 0 & 24 & tafas - frozen & 0.0005 & -0.0006 & 0.0018 \\
weather & 0 & 48 & cosa - frozen$^{*}$ & 0.0100 & 0.0010 & 0.0204 \\
weather & 0 & 48 & dynatta - frozen$^{*}$ & 0.0150 & 0.0003 & 0.0271 \\
weather & 0 & 48 & elf\_linear - rls\_bank\_med$^{*}$ & 0.0768 & 0.0101 & 0.1817 \\
weather & 0 & 48 & elf\_mix - cosa$^{*}$ & -0.0637 & -0.0848 & -0.0381 \\
weather & 0 & 48 & elf\_mix - dynatta$^{*}$ & -0.0687 & -0.0838 & -0.0502 \\
weather & 0 & 48 & elf\_mix - frozen$^{*}$ & -0.0537 & -0.0718 & -0.0350 \\
weather & 0 & 48 & elf\_mix - naive$^{*}$ & -0.4415 & -0.5347 & -0.3602 \\
weather & 0 & 48 & elf\_mix - petsa$^{*}$ & -0.1854 & -0.2122 & -0.1582 \\
weather & 0 & 48 & elf\_mix - rls\_bank\_med & -0.0075 & -0.0307 & 0.0262 \\
weather & 0 & 48 & elf\_mix - tafas$^{*}$ & -0.0538 & -0.0693 & -0.0301 \\
weather & 0 & 48 & petsa - frozen$^{*}$ & 0.1318 & 0.1096 & 0.1544 \\
weather & 0 & 48 & rls\_bank\_med - cosa$^{*}$ & -0.0562 & -0.0995 & -0.0225 \\
weather & 0 & 48 & rls\_bank\_med - dynatta$^{*}$ & -0.0612 & -0.0959 & -0.0351 \\
weather & 0 & 48 & rls\_bank\_med - frozen$^{*}$ & -0.0461 & -0.0882 & -0.0134 \\
weather & 0 & 48 & rls\_bank\_med - naive$^{*}$ & -0.4339 & -0.5418 & -0.3391 \\
weather & 0 & 48 & rls\_bank\_med - petsa$^{*}$ & -0.1779 & -0.2188 & -0.1481 \\
weather & 0 & 48 & rls\_bank\_med - tafas$^{*}$ & -0.0462 & -0.0869 & -0.0141 \\
weather & 0 & 48 & tafas - frozen & 0.0001 & -0.0009 & 0.0014 \\
weather & 0 & 96 & cosa - frozen & 0.0045 & -0.0071 & 0.0187 \\
weather & 0 & 96 & dynatta - frozen$^{*}$ & 0.0204 & 0.0104 & 0.0331 \\
weather & 0 & 96 & elf\_linear - rls\_bank\_med$^{*}$ & 0.0187 & 0.0026 & 0.0379 \\
weather & 0 & 96 & elf\_mix - cosa$^{*}$ & -0.0663 & -0.0835 & -0.0515 \\
weather & 0 & 96 & elf\_mix - dynatta$^{*}$ & -0.0822 & -0.0994 & -0.0674 \\
weather & 0 & 96 & elf\_mix - frozen$^{*}$ & -0.0618 & -0.0720 & -0.0510 \\
weather & 0 & 96 & elf\_mix - naive$^{*}$ & -0.4496 & -0.5541 & -0.3700 \\
weather & 0 & 96 & elf\_mix - petsa$^{*}$ & -0.1929 & -0.2225 & -0.1678 \\
weather & 0 & 96 & elf\_mix - rls\_bank\_med & -0.0131 & -0.0405 & 0.0280 \\
weather & 0 & 96 & elf\_mix - tafas$^{*}$ & -0.0616 & -0.0749 & -0.0509 \\
weather & 0 & 96 & petsa - frozen$^{*}$ & 0.1312 & 0.1085 & 0.1542 \\
weather & 0 & 96 & rls\_bank\_med - cosa$^{*}$ & -0.0532 & -0.0946 & -0.0214 \\
weather & 0 & 96 & rls\_bank\_med - dynatta$^{*}$ & -0.0690 & -0.1135 & -0.0380 \\
weather & 0 & 96 & rls\_bank\_med - frozen$^{*}$ & -0.0486 & -0.0904 & -0.0169 \\
weather & 0 & 96 & rls\_bank\_med - naive$^{*}$ & -0.4364 & -0.5462 & -0.3451 \\
weather & 0 & 96 & rls\_bank\_med - petsa$^{*}$ & -0.1798 & -0.2146 & -0.1448 \\
weather & 0 & 96 & rls\_bank\_med - tafas$^{*}$ & -0.0485 & -0.0860 & -0.0180 \\
weather & 0 & 96 & tafas - frozen & -0.0002 & -0.0015 & 0.0014 \\
\end{longtable}
}

\section{Released-Script Configuration Re-Pass (ETT streams)}
\label{app:tuned}
The reference release contains a single global configuration in its \texttt{scripts/} directory; no per-dataset tuning exists in the release. Relative to the repository defaults used in Section~\ref{sec:headline}, the script settings change the following flags: for TAFAS, learning rate $0.005 \to 0.001$; for COSA, learning rate $\to 0.001$, adaptation steps $20 \to 3$, PAAS batch $25 \to 48$, context buffer $5 \to 10$ (growing the adapter from 5{,}047 to 5{,}887 parameters), and the \texttt{PAAS} flag off $\to$ on. We re-ran both methods under these settings (\texttt{tafas\_tuned}, \texttt{cosa\_tuned}) on the same ETT checkpoints (seed 0), streams and origin-aligned scoring as the main text; the default-configuration values in Table~\ref{tab:tuned} are therefore also the seed-0 stream, so the two configurations are compared like for like, and they differ from the three-seed means of Table~\ref{tab:external}. The \texttt{\_tuned} bootstrap pairs appear in Tables~\ref{tab:ci_external_ettm1} and~\ref{tab:ci_external_etth2}. Under the script configuration COSA's larger batches keep causal prefixes for their older origins even under lag, so its partial-ground-truth updates no longer vanish at $\ell > 0$ (79 and 60 partial updates on ETTh2 at $\ell = 24$ and $\ell = 72$); it remains delay-degraded on ETTm1.

No single released configuration is defensible on both streams: the differences between the authors' default and script settings are themselves statistically significant, in opposite directions on the two datasets (COSA script vs.\ default: $+0.15^{*}$ to $+0.32^{*}$ worse on ETTm1, $-0.76^{*}$ better on ETTh2 at $d{=}4H$), while the bank runs one configuration everywhere.

\begin{table}[H]\centering\scriptsize
\caption{Repository-default vs.\ released-script configurations on the authors' own
ETT checkpoints, seed 0 (standardized MSE, origin-aligned). The script configuration is significantly
worse on ETTm1 at every delay and equivalent to the frozen model on ETTh2 ($|\Delta|\le 0.0005$, n.s.).
Script-configuration COSA uses a larger context buffer (5{,}887 parameters vs.\ 5{,}047).}
\label{tab:tuned}
\begin{tabular}{llrrr}\toprule
 & configuration & $d{=}H$ & $d{=}2H$ & $d{=}4H$ \\\midrule
ETTh2 & TAFAS default & 1.2463 & 1.2614 & 1.2558 \\
 & TAFAS script & 1.2428 & 1.2410 & 1.2388 \\
 & COSA default & 1.5293 & 1.7624 & 1.9952 \\
 & COSA script & 1.2357 & 1.2360 & 1.2353 \\
 & RLS-Bank (ref.) & 1.3136 & 1.3908 & 1.5897 \\
 & \emph{frozen 1.2355; persistence 1.1196} & & & \\
\addlinespace
ETTm1 & TAFAS default & 1.4649 & 1.4649 & 1.4648 \\
 & TAFAS script & 1.4656 & 1.4655 & 1.4656 \\
 & COSA default & 1.0327 & 0.9155 & 0.9382 \\
 & COSA script & 1.1839 & 1.2116 & 1.2563 \\
 & RLS-Bank (ref.) & 0.6708 & 0.6755 & 0.6837 \\
 & \emph{frozen 1.6676; persistence 1.8007} & & & \\
\addlinespace
\bottomrule\end{tabular}
\end{table}

\end{document}